\documentclass[twoside,11pt]{article}
\usepackage{blindtext}
\usepackage{mathtools}
\usepackage[preprint]{jmlr2e}

\usepackage{hyperref}
\usepackage{url}

\usepackage{rotating}
\usepackage{booktabs}
\usepackage{array}
\usepackage{amsmath, amsfonts}
\usepackage{color}
\usepackage{hyperref}
\usepackage{bbm}
\usepackage[noend]{algpseudocode}
\usepackage{enumerate}
\usepackage{stmaryrd,scalerel}
\usepackage{enumitem}
\usepackage{mathtools}
\mathtoolsset{showonlyrefs}
\usepackage[ruled,vlined]{algorithm2e}
\usepackage{caption}
\usepackage{xr}
\usepackage{multirow}
\usepackage{makecell}
\hypersetup{ hidelinks }
\usepackage{subfig}

\usepackage[normalem]{ulem}
\usepackage{soul}
\usepackage[toc,page]{appendix}

\newtheorem{prop}{Proposition}
\newtheorem{cor}{Corollary}
\newtheorem{recipe}{Recipe}

\usepackage{chngcntr}
\usepackage{apptools}
\AtAppendix{\counterwithin{theorem}{section}}
\AtAppendix{\counterwithin{prop}{section}}

\usepackage{mathtools}

\def\R{\mathbb{R}}
\def\E{\mathbb{E}}

\newcommand{\Y}{\mathcal{Y}}
\newcommand{\X}{\mathcal{X}}

\newcommand{\ind}[1]{\mathbbm{1}\left\{#1\right\}}

\renewcommand{\P}{\mathbb{P}}

\newcommand{\Chat}{\widehat{\mathcal{C}}}

\newcommand{\qhat}{\hat{q}}
\newcommand{\qhatclean}{\hat{q}_{\rm clean}}
\newcommand{\qhatnoisy}{\hat{q}_{\rm noisy}}
\newcommand{\Chatnoisy}{\Chat_{\rm noisy}}
\newcommand{\Chatnoisyt}{\Chat_{\rm noisy}^{t}}

\newcommand{\Xtest}{X_{\rm test}}
\newcommand{\Ytest}{Y_{\rm test}}

\newcommand{\tilY}{\tilde{Y}}
\newcommand{\ts}{\tilde{s}}

\newcommand{\indep}{\perp \!\!\! \perp}

\makeatletter
\def\blfootnote{\xdef\@thefnmark{}\@footnotetext}
\makeatother

\usepackage{lastpage}
\jmlrheading{25}{2024}{1-\pageref{LastPage}}{11/23; Revised 10/24}{10/24}{23-1549}{Bat-Sheva Einbinder, Shai Feldman, Stephen Bates, Anastasios N. Angelopoulos, Asaf Gendler, Yaniv Romano} 
\ShortHeadings{Label Noise Robustness of Conformal Prediction}{Einbinder, Feldman, Bates, Angelopoulos, Gendler, Romano}

\firstpageno{1}

\begin{document}
\title{Label Noise Robustness of Conformal Prediction}


\author{\name Bat-Sheva Einbinder$^*$ \email bat-shevab@campus.technion.ac.il \\
       \addr Department of Electrical and Computer Engineering \\
              Technion - Israel Institute of Technology
       \AND
       \name Shai Feldman$^*$ \email shai.feldman@cs.technion.ac.il \\
       \addr Department of Computer Science\\
       Technion - Israel Institute of Technology
      \AND
      \name Stephen Bates
      \email stephenbates@mit.edu \\
       \addr Department of Electrical Engineering and Computer Science \\
            Massachusetts Institute of Technology
        \AND
      \name Anastasios N. Angelopoulos
      \email angelopoulos@berkeley.edu \\
        \addr Department of Electrical Engineering and Computer Science\\
        University of California, Berkeley
        \AND
      \name Asaf Gendler
      \email asafgendler@campus.technion.ac.il \\
        \addr Department of Electrical and Computer Engineering \\
        Technion---Israel Institute of Technology
       \AND
         \name Yaniv Romano
      \email yromano@technion.ac.il\\
              \addr Departments of Electrical and Computer Engineering and of Computer Science\\
        Technion---Israel Institute of Technology\\
       }
\def\thefootnote{*}\footnotetext{Equal contribution. These authors are listed in alphabetical order.}

\editor{Zaid Harchaoui}

\maketitle











\begin{abstract}%
We study the robustness of conformal prediction, a powerful tool for uncertainty quantification, to label noise. Our analysis tackles both regression and classification problems, characterizing when and how it is possible to construct uncertainty sets that correctly cover the unobserved noiseless ground truth labels. We further extend our theory and formulate the requirements for correctly controlling a general loss function, such as the false negative proportion, with noisy labels. 
Our theory and experiments suggest that conformal prediction and risk-controlling techniques with noisy labels attain conservative risk over the clean ground truth labels 
whenever the noise is dispersive and increases variability.
In other adversarial cases, we can also correct for noise of bounded size in the conformal prediction algorithm in order to ensure achieving the correct risk of the ground truth labels without score or data regularity. 
\end{abstract}


\section{Introduction}

In most supervised classification and regression tasks, one would assume the provided labels reflect the ground truth.
In reality, this assumption is often violated \citep{cheng2022many,xu2019l_dmi,yuan2018iterative, lee2022binary, cauchois2022weak}. 
For example, doctors labeling the same medical image may have different subjective opinions about the diagnosis, leading to variability in the ground truth label itself. To quote \citet{abdalla2023hurdles}: ``Noise in labeling schemas and gold label annotations are pervasive in medical imaging classification and affect downstream clinical deployment.''
In other settings, such variability may arise due to sensor noise, data entry mistakes, the subjectivity of a human annotator, or many other sources.
In other words, the labels we use to train machine learning (ML) models may often be noisy in the sense that these are not necessarily the ground truth. Consequently, this can result in the formulation of invalid data-driven conclusions. 

The above discussion emphasizes the critical need for making reliable predictions in real-world scenarios, especially when dealing with imperfect training data. An effective way to enhance the reliability of ML models is to quantify their prediction uncertainty. Conformal prediction~\citep{vovk2005algorithmic,vovk1999machine,angelopoulos2021gentle} is a generic uncertainty quantification tool that transforms the output of any ML model into prediction sets that are guaranteed to cover the future, unknown test labels with high probability. This guarantee holds for any data distribution and sample size, under the sole assumption that the training and test data are i.i.d. However, obtaining valid uncertainty quantification via conformal prediction in the presence of noisy labels remains unclear since the noise breaks the i.i.d. assumption. In this paper, we aim to address this specific challenge and precisely characterize under what conditions conformal methods, applied to noisy data, would yield prediction sets guaranteed to cover the unseen clean, ground truth label.
Additionally, we analyze the effect of label noise on risk-controlling techniques, which extend the conformal prediction approach to construct uncertainty sets with a guaranteed control over a general risk function~\citep{bates2021distributionfree, angelopoulos2021learn, angelopoulos2022conformal}.
We also show that our theory can be applied to online settings, e.g., to quantify prediction uncertainty for time series data with noisy labels. Overall, we analyze the behavior of conformal prediction and risk-controlling methods for several common loss functions and noise models, highlighting their built-in robustness to dispersive, variability-increasing noise and vulnerability to adversarial noise. Adversarial noise might reduce the uncertainty of the response variable, potentially causing the prediction sets to be too small and achieve a low coverage rate.
We note that a summary of this paper and its key contributions can be found in~\citep{feldman2023conformal}.


\section{Conformal Prediction Under Label Noise}

\subsection{Problem Setup}

Consider a \emph{calibration data set} of i.i.d. observations $\{ (X_i,Y_i) \}_{i=1}^n$ sampled from an arbitrary unknown distribution $P_{XY}$. Here, $X_i \in \mathbb{R}^p$ is the feature vector that contains $p$ features for the $i$-th sample, and $Y_i$ denotes its response, which can be discrete for classification tasks or continuous for regression tasks.
Given the calibration data set, an i.i.d. test data point $(\Xtest,\Ytest)$, and a pre-trained model $\hat{f}$, conformal prediction constructs a set $\widehat{\mathcal{C}}_{\rm clean}(\Xtest)$ that contains the unknown test response, $Y_{\rm test}$, with high probability, e.g., 90\%. We refer to $\widehat{\mathcal{C}}_{\rm clean}(\Xtest)$ as `clean'  to underscore that the prediction set is formed by utilizing samples from the clean data distribution.
That is, for a user-specified level $\alpha \in (0,1)$,
\begin{align} 
    \label{eq:marginal-coverage}
    \mathbb{P}\left(Y_{\rm test} \in \widehat{\mathcal{C}}_{\rm clean} (X_{\rm test}) \right) \geq 1-\alpha.
\end{align}
This property is called {\it marginal coverage}, where the probability is defined over the calibration and test data.

In the setting of label noise, we only observe the corrupted labels $\tilY_i = g(Y_i)$ for some \emph{corruption function} $g : \Y \times [0,1] \to \Y$, so the i.i.d. assumption and marginal coverage guarantee are invalidated.
The corruption is random; we will always take the second argument of $g$ to be a random seed $U$ uniformly distributed on $[0,1]$. To ease notation, we leave the second argument implicit henceforth.
Nonetheless, using the noisy calibration data, we seek to form a prediction set $\widehat{\mathcal{C}}_{\rm noisy}(\Xtest)$ that covers the clean, uncorrupted test label, $\Ytest$.
More precisely, our goal is to delineate when it is possible to provide guarantees of the form
\begin{equation}
    \label{eq:goal}
    \mathbb{P}\left(Y_{\rm test} \in \widehat{\mathcal{C}}_{\rm noisy} (X_{\rm test}) \right) \geq 1-\alpha,
\end{equation}
where the probability is taken jointly over the calibration data, test data, and corruption function (this will be the case for the remainder of the paper).

More formally, we use the model $\hat{f}$ to construct a score function, $s : \X \times \Y \to \R$, which is engineered to be large when the model is uncertain and small otherwise. We will introduce different score functions for both classification and regression as needed in the following subsections. Abbreviate the scores on each calibration data point as $s_i = s(X_i,Y_i)$ for each $i=1,...,n$.
Conformal prediction tells us that we can achieve a marginal coverage guarantee by picking $\qhatclean = s_{(\lceil (n+1) (1-\alpha) \rceil)}$ as the $\lceil (n+1) (1-\alpha) \rceil$-smallest of the calibration scores and constructing the prediction sets as
\begin{equation}
    \label{eq:conformal_pred_set}
    \widehat{\mathcal{C}}_{\rm clean}\left(X_{\rm test}\right)=\left\{y\in\mathcal{Y}:s\left(X_{\rm test},y\right)\leq\qhatclean\right\}.
\end{equation}
In this paper, we do not allow ourselves access to the calibration labels, only their noisy versions, $\tilde{Y}_1, \ldots, \tilde{Y}_n$, so we cannot calculate $\hat{q}_{\rm clean}$.
Instead, we can calculate the noisy quantile $\hat{q}_{\rm noisy}$ as the $\lceil (n+1)(1-\alpha) \rceil$-smallest of the \emph{noisy} score functions, $\ts_i = s(X_i,\tilY_i)$.
The main formal question of our work is whether the resulting prediction set, $\Chat_{\rm noisy}(X_{\rm test}) = \{y : s(X_{\rm test},y) \leq \hat{q}_{\rm noisy}\}$, covers the clean label as in~\eqref{eq:goal}.
We state this general recipe algorithmically for future reference:
\begin{recipe}[Conformal prediction with noisy labels]
    \label{recipe}
    \begin{enumerate}
        \item[]
        \item Consider i.i.d. data points $(X_1,Y_1), \ldots, (X_n,Y_n), (X_{\rm test}, Y_{\rm test})$, a \emph{corruption model} $g: \Y \to \Y$, and a score function $s: \X \times \Y \to \R$.
        \item Compute the conformal quantile with the corrupted labels, \begin{equation}\label{eq:qnoisy} \qhatnoisy = \mathrm{Quantile}\left( \frac{(n+1)(1-\alpha)}{n}, \{s(X_i, \tilY_i)\}_{i=1}^n \right),\end{equation} where $\tilY_i = g(Y_i)$.
        \item Construct the prediction set using the noisy conformal quantile, \begin{equation}\label{eq:Cnoisy} \Chatnoisy = \{ y : s(\Xtest,y) \leq \qhatnoisy \}.\end{equation}
    \end{enumerate}
\end{recipe}
This recipe produces prediction sets that cover the noisy label at the desired coverage rate \citep{vovk2005algorithmic, angelopoulos2021gentle}:
\begin{equation}
\P\left(\tilde{Y}_{\rm test} \in \Chatnoisy(X_{\rm test})\right) \geq 1-\alpha.
\end{equation}

Note that achieving~\eqref{eq:goal} is impossible in general---see Proposition~\ref{prop:general-case}.
However, we present some realistic noise models under which this goal is satisfied.
In the following sections, we provide real-data experiments 
indicating that conformal prediction and risk controlling methods achieve valid risk/coverage even with access only to noisy labels.
In Section~\ref{sec:conformal_analysis}, we outline the precise conditions for these guarantees.
There is a particular type of anti-dispersive noise that causes failure, and we see this both in theory and practice.
If such noise can be avoided, a user should feel safe deploying uncertainty quantification techniques even with noisy labels. Finally, we note that there exist several works that study the robustness of conformal prediction under ambiguous labels, and we discuss them in more detail in Section~\ref{sec:related_work}.


\subsection{Empirical Evidence}

\subsubsection*{Classification}\label{sec:real_class}

As a real-world example of label noise, we conduct an image classification experiment where we only observe one annotator's label but seek to cover the majority vote of many annotators.
For this purpose, we use the CIFAR-10H data set, used by~\cite{peterson2019human,battleday2020capturing,singh2020end}, which contains 10,000 images labeled by approximately 50 annotators.
We calibrate using only a single annotator's label and seek to cover the majority vote of the 50 annotators.
The single annotator differs from the ground truth labels in approximately 5\% of the images.

\begin{figure}[!htb]

  \centering
  \begin{minipage}{0.36\textwidth}
    \includegraphics[width=\linewidth]{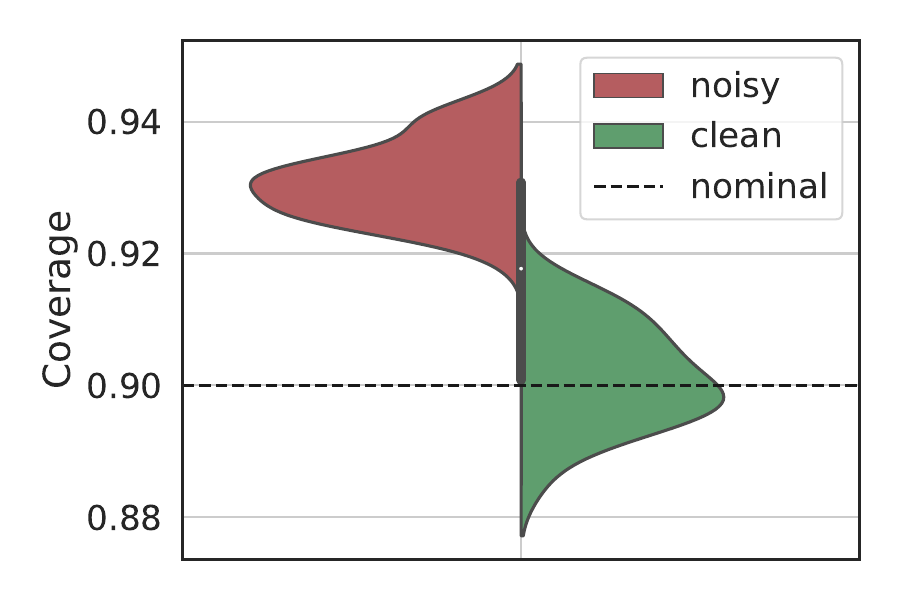}
  \end{minipage}
  \begin{minipage}{0.16\textwidth}
    \includegraphics[trim=26 23 0 0, clip, width=\linewidth]{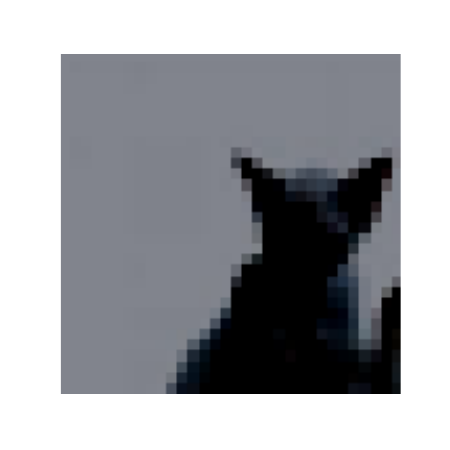}
  \end{minipage}
  \begin{minipage}{0.14\textwidth}
    \begin{itemize}[leftmargin=*]
    \small
    \item True label:\\\ Cat
    \item Noisy: \\\{Cat, Dog\} 
    \item Clean:\\\{Cat\} 
    \end{itemize}
  \end{minipage}
  \begin{minipage}{0.16\textwidth}
    \includegraphics[trim=26 23 0 0, clip, width=\linewidth]{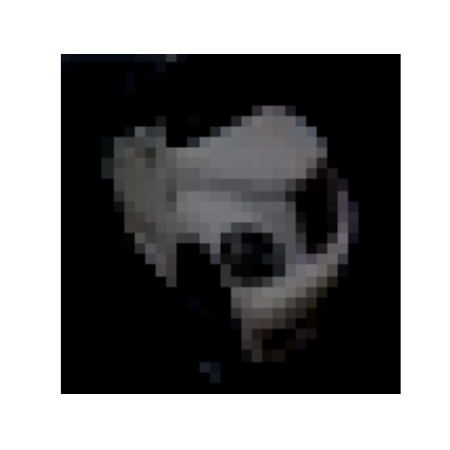}
  \end{minipage}
  \begin{minipage}{0.14\textwidth}
    \begin{itemize}[leftmargin=*]
    \small
    \item True label:\\\ Car
    \item Noisy:\\\{Car, Ship, Cat\} 
    \item Clean:\\\{Car\} 
    \end{itemize}
  \end{minipage}
  
\caption{\textbf{Effect of label noise on CIFAR-10.} Left: distribution of average coverage on a clean test set over 30 independent experiments with target coverage $1-\alpha = 90\%$, using noisy and clean labels for calibration. We use a pre-trained resnet 18 model, which has Top-1 accuracy of 93\% and 90\% on the clean and noisy test set, respectively. The gray bar represents the interquartile range. Center and right: prediction sets achieved using noisy and clean labels for calibration.}
\label{fig:cifar10-clean}
\end{figure}

Using the noisy calibration set (i.e., a calibration set containing these noisy labels), we apply vanilla conformal prediction as if the data were i.i.d, and study the performance of the resulting prediction sets. 
Details regarding the training procedure can be found in Appendix~\ref{app:cifar-10-supp}. 
The fraction of majority vote labels covered is demonstrated in Figure~\ref{fig:cifar10-clean}.
This figure shows that when using the clean calibration set, the marginal coverage is 90\%, as expected. 
When using the noisy calibration set, the coverage on the clean, unknown test point, i.e.,~\eqref{eq:goal} increases to approximately 93\%. Figure~\ref{fig:cifar10-clean} also demonstrates that prediction sets that were calibrated using noisy labels are larger than sets calibrated with clean labels. 
This experiment illustrates the main intuition behind our paper: adding noise will usually increase the variability in the labels, leading to larger prediction sets that retain the coverage property.

\subsubsection*{Regression}\label{sec:real_reg}

In this section, we present a real-world application with a continuous response, using Aesthetic Visual Analysis (AVA) data set, first presented by~\cite{murray2012ava}. This data set contains pairs of images and their aesthetic scores in the range of 1 to 10, obtained by approximately 200 annotators. Following~\cite{kao2015visual, talebi2018nima, murray2012ava}, the task is to predict the average aesthetic score of a given test image. Therefore, we consider the average aesthetic score taken over all annotators as the clean, ground truth response. The noisy response is the average aesthetic score taken over 10 randomly selected annotators only.

We examine the performance of conformal prediction using two different scores: the CQR score \citep{romano2019conformalized}, defined in~\eqref{eq:CQR-score} and the residual magnitude score \citep{papadopoulos2002inductive,lei2018distribution} defined in~\eqref{eq:rm-score}. In our experiments, we set the function $\hat{u}(x)$ of the residual magnitude score as $\hat{u}(x) = 1$. We follow~\cite{talebi2018nima} and take a transfer learning approach to fit the predictive model using a VGG-16 model pretrained on ImageNet data set. 
Details regarding the training strategy are in Appendix~\ref{app:AVA-supp}.

Figure~\ref{fig:AVA-noisy} portrays the marginal coverage and average interval length achieved using CQR and residual magnitude scores. As a point of reference, this figure also presents the performance of the two conformal methods when calibrated with a clean calibration set; as expected, the two perfectly attain 90\% coverage. By constant, when calibrating the same predictive models with a noisy calibration set, the resulting prediction intervals tend to be wider and to over-cover the average aesthetic scores.

\begin{figure}[t]
\begin{center}
{\includegraphics[width=.5\linewidth]{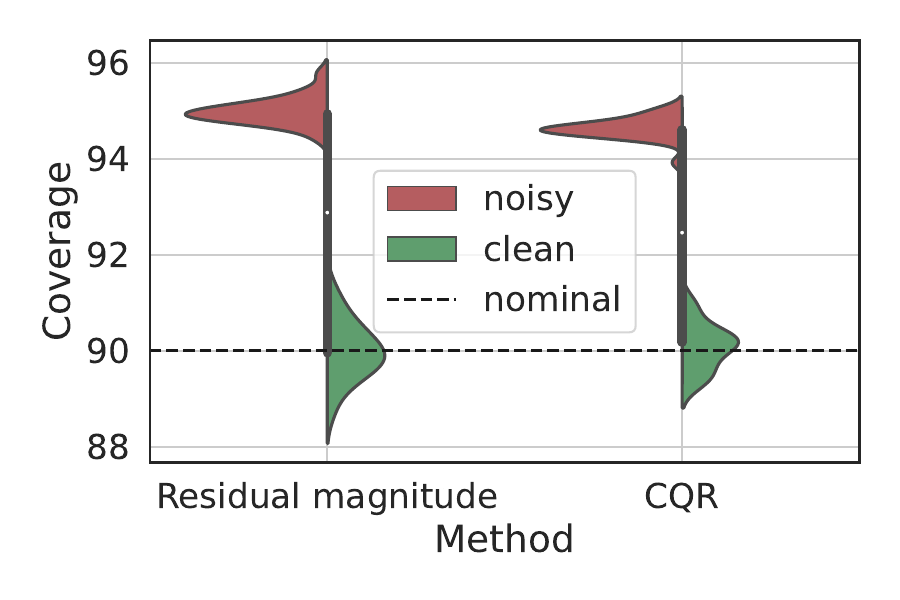}}\hfill
{\includegraphics[width=.5\linewidth]{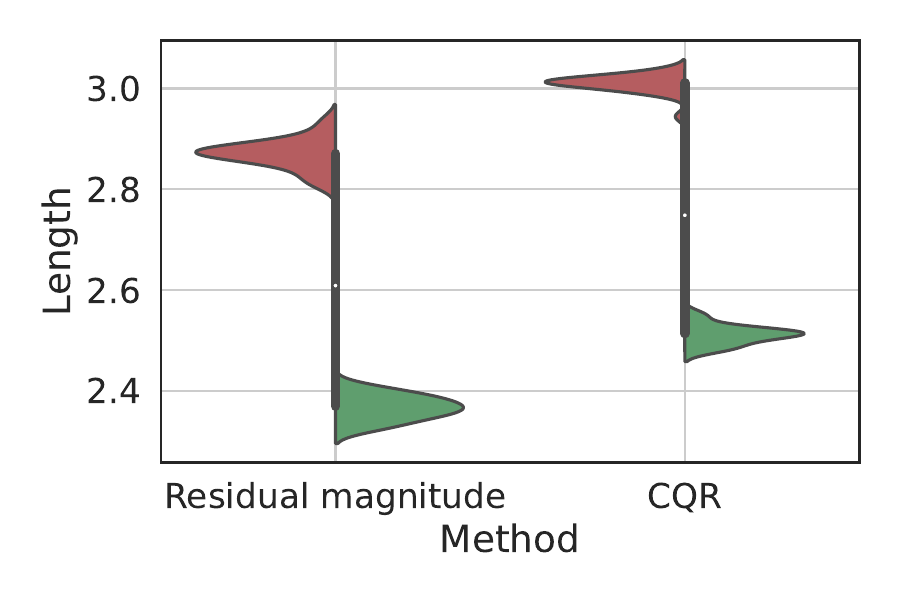}}\par 
\caption{\textbf{Results for real-data regression experiment: predicting aesthetic visual rating.} Performance of conformal prediction intervals with 90\% marginal coverage based on a VGG-16 model using a noisy training set. We compare the residual magnitude score and CQR methods with both noisy and clean calibration sets. Left: Marginal coverage; Right: Interval length.
The results are evaluated over 30 independent experiments and the gray bar represents the interquartile range.}
\label{fig:AVA-noisy}
\end{center}
\vspace{-0.5cm}
\end{figure}

Thus far, we have found in empirical experiments that conservative coverage is obtained in the presence of label noise. In the following sections, our objective is to establish conditions that formally guarantee label-noise robustness.

\subsection{Theoretical Analysis}\label{sec:conformal_analysis}
\subsubsection{General Analysis}
We begin the theoretical analysis with a general statement, showing that Recipe~\ref{recipe} produces valid prediction sets whenever the noisy score distribution stochastically dominates the clean score distribution.
The intuition is that the noise distribution `spreads out' the distribution of the score function such that $\hat{q}_{\rm noisy}$ is (stochastically) larger than $\hat{q}_{\rm clean}$.
\begin{theorem}
    \label{thm:general}
    Assume that 
        $\P(\ts_{\rm test} \leq t) \leq \P(s_{\rm test} \leq t)$
    for all $t$.
    Then,
    \begin{equation}
        \P\left(Y_{\rm test} \in \Chatnoisy(X_{\rm test})\right) \geq 1-\alpha.
    \end{equation}
    Furthermore, for any $u$ satisfying
        $\P(\ts_{\rm test} \leq t) + u \geq \P(s_{\rm test} \leq t),$
    for all $t$, then
    \begin{equation}
        \P\left(Y_{\rm test} \in \Chatnoisy(X_{\rm test})\right) \leq 1-\alpha+\frac{1}{n+1}+u.
    \end{equation}
\end{theorem}
\begin{figure}[ht]
    \centering
    \includegraphics[scale=0.6]{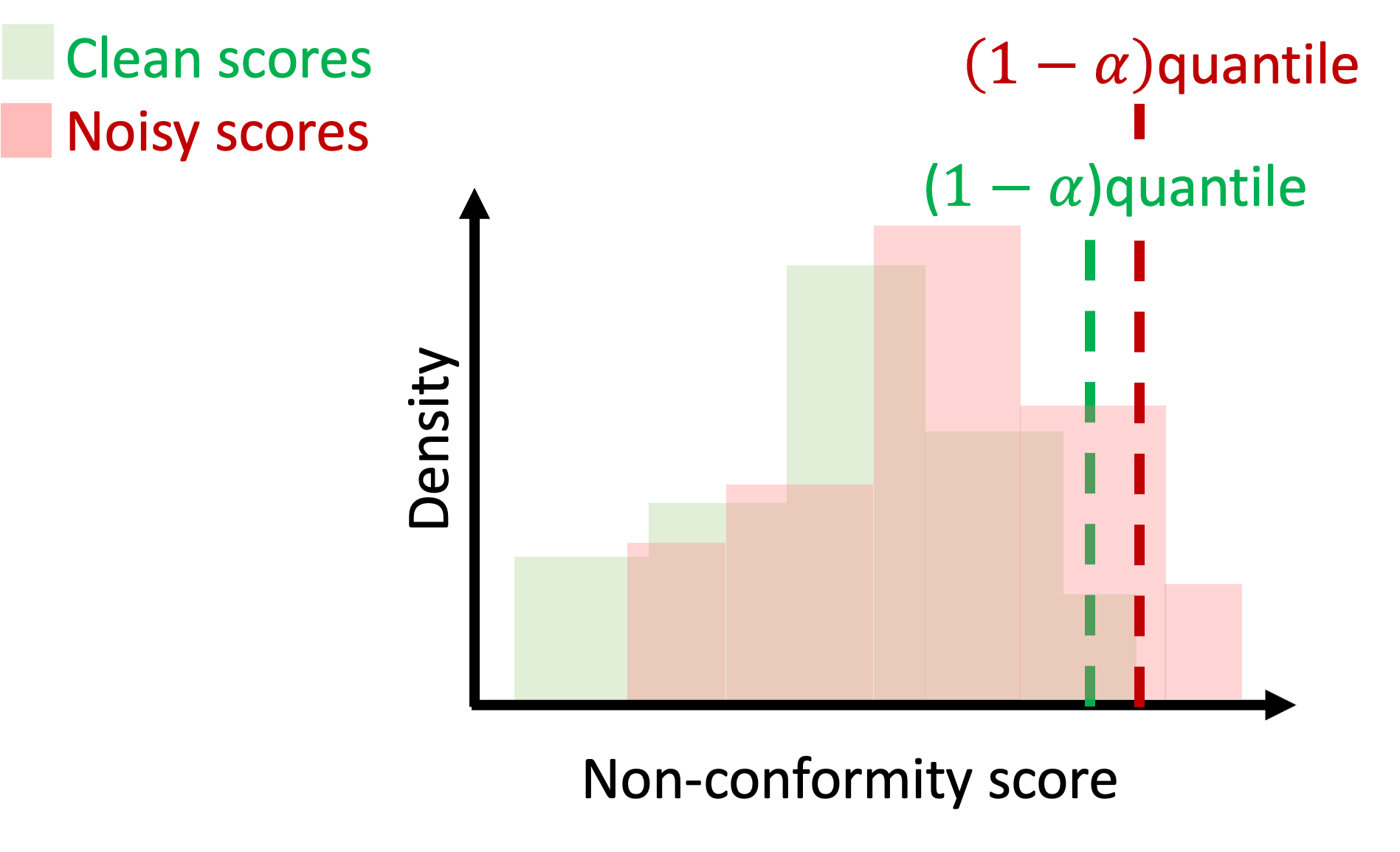}
    \caption{Clean (green) and noisy (red) non-conformity scores under dispersive corruption.}
    \label{fig:thm1_scores}
\end{figure}
Figure~\ref{fig:thm1_scores} illustrates the idea behind Theorem~\ref{thm:general}, demonstrating that when the noisy score distribution stochastically dominates the clean score distribution, then $\qhatnoisy \geq \qhatclean$ and thus uncertainty sets calibrated using noisy labels are more conservative. In practice, however, one does not have access to such a figure, since the scores of the clean labels are unknown. This gap requires an individual analysis for every task and its noise setup, which emphasizes the complexity of this study.
In general, for most commonly used score functions the stochastic dominance assumption holds when the noise is dispersive, meaning it ``flattens'' the density of $Y\mid X$, e.g., when $\text{Var}(\tilde{Y}\mid X)>\text{Var}(Y\mid X)$.
In the following subsections, we present example setups in classification and regression tasks under which this stochastic dominance holds, and conformal prediction with noisy labels succeeds in covering the true, noiseless label.
The purpose of these examples is to illustrate simple and intuitive statistical settings where Theorem~\ref{thm:general} holds.
Under the hood, all setups given in the following subsections are applications of Theorem~\ref{thm:general}.
Though the noise can be adversarially designed to violate these assumptions and cause under-coverage (as in the impossibility result in Proposition~\ref{prop:general-case}), the evidence presented here suggests that in the majority of practical settings, conformal prediction can be applied without modification.
The proof is given in Appendix~\ref{app-general}.

\subsubsection{Regression}
In this section, we analyze a regression task where the labels are continuous-valued and the corruption function is additive:
\begin{equation}\label{eq:reg_g_sym}
    g^{\rm add}(y) = y + Z
\end{equation}
for some independent noise sample $Z$.

We first analyze the setting where the noise $Z$ is symmetric around 0 and the density of $Y\mid X$ is symmetric unimodal. We also assume that the estimated prediction interval contains the true median of $\tilY\mid X = x$, which is a very weak assumption about the fitted model. The following proposition states that such an interval achieves a conservative coverage rate over the clean labels. 
\begin{prop}
  \label{prop:regression_valid_miscoverage}
  Assume a symmetric unimodal distribution of $Y\mid X$ and an independent additive noise that is symmetric around zero. Let $\Chatnoisy(\Xtest)$ be a prediction interval constructed according to Recipe~\ref{recipe} with the corruption function $g^{\rm add}$. Further, suppose that $\Chatnoisy(x)$ contains the true median of $\tilY\mid X = x$ for all $x\in\mathcal{X}$.
  Then
  \begin{equation}
    \P\Big( \Ytest \in \Chatnoisy(\Xtest) \Big) \geq 1-\alpha.
  \end{equation}
\end{prop}
Importantly, the corruption function may depend on the feature vector, as stated next.
\begin{remark}[Corruptions dependent on X]
Proposition~\ref{prop:regression_valid_miscoverage} holds even if the noise $Z$ depends on $X$.
\end{remark}
Furthermore, Proposition~\ref{prop:regression_valid_miscoverage} applies for any non-conformity score function,
specifically for two popular scores we focus on.
The CQR score, developed by~\cite{romano2019conformalized}, measures the distance of an estimated interval's endpoints from the corresponding label $y$, formally defined as:
\begin{equation}
    \label{eq:CQR-score}
    s^{\rm CQR}\left(x,y\right)= \max\{\hat{f}_{\text{lower}}(x)-y, y-\hat{f}_{\text{upper}}(x)\}.
\end{equation}
Above, $\hat{f}_{\text{lower}}$ and $\hat{f}_{\text{upper}}$ are the estimated lower and upper interval's endpoints, e.g., obtained by fitting a quantile regression model to approximate the $\alpha/2$ and $1-\alpha/2$ conditional quantiles of $\tilde{Y}\mid X$ \citep{koenker1978regression}. The residual magnitude (RM) score \citep{papadopoulos2002inductive,lei2018distribution} assesses the normalized prediction error:
\begin{equation}\label{eq:rm-score}
    s^{\rm RM}(x,y) = \big|\hat{f}(x) - y\big| / \hat{u}(x),
\end{equation}
where $\hat{f}$ is a regression model, such as a conditional mean estimator of $\tilde{Y} \mid X$, and $\hat{u}$ is some normalization function, e.g., $\hat{u}(x) \equiv 1$.
Additionally, the distributional assumption on $Y\mid X$ is not necessary to guarantee label-noise robustness under this noise model; in Appendix~\ref{sec:general_regression_result} we show that this distributional assumption can be relaxed at the expense of additional requirements on the estimated interval. 
Furthermore, in Section~\ref{sec:risk_regression} we extend this analysis and formulate bounds for the coverage rate obtained on the clean labels that do not require any distributional assumptions and impose no restrictions on the constructed intervals. 
Subsequently, in Section~\ref{sec:exp_reg_bounds}, we empirically evaluate the proposed bounds and show that they are informative despite their extremely weak assumptions.

\subsubsection{Classification}\label{sec:classification_analysis}
In this section, we formulate the conditions under which conformal prediction is robust to label noise in a $K$-class classification setting where the labels take one of $K$ values, i.e., $Y\in\{1,2,...,K\}$. 
In a nutshell, robustness is guaranteed when the corruption function transforms the labels' distribution towards a uniform distribution while maintaining the same ranking of labels.
Such corruption increases the prediction uncertainty, which drives conformal methods to construct conservative uncertainty sets to achieve the nominal coverage level on the observed noisy labels. We now formalize this intuition. We begin by defining the following noise models:
\begin{itemize}
    \item \texttt{Uniform noise}: A noise model that fulfils the following:
    for all $x\in\mathcal{X}$:
\begin{enumerate}
    \item $\forall i \in \{1,..,K\}: \left|\mathbb{P}(\tilde{Y}_{\rm test}=i \mid \Xtest=x) - \frac{1}{K}\right| \leq \left|\mathbb{P}(\Ytest=i \mid \Xtest=x) - \frac{1}{K}\right|,$
    \item $ \forall i,j\in \{1,..,K\} : \mathbb{P}(\Ytest=i \mid \Xtest=x) \leq \mathbb{P}(\Ytest=j \mid \Xtest=x) \iff \\
    \mathbb{P}(\tilde{Y}_{\rm test}=i \mid \Xtest=x) \leq \mathbb{P}(\tilde{Y}_{\rm test}=j \mid \Xtest=x).$

\end{enumerate}
\item \texttt{Random flip}: A corruption function that follows:
\begin{equation}
    \label{eq:epsilon-class-corruption}
	g^{\rm flip}(y) = \begin{cases} 
						y & \text{w.p } 1-\epsilon \\
						Y' & \text{else},
				\end{cases} 
\end{equation}
where $Y'$ is uniformly drawn from the set $\{1,...,K\}$.

\end{itemize}
\begin{prop}\label{prop:general_classification}
Let $\Chatnoisy$ be constructed as in Recipe~\ref{recipe}.
Then, the coverage rate achieved over the clean labels is upper bounded by:
\begin{equation}
\mathbb{P}\left({Y}_{\rm test}\in \Chatnoisy(X_{\rm test})\right) \leq  1-\alpha + \frac{1}{n+1} + \frac{1}{2}\sum_{i=1}^{K} {\left|\mathbb{P}(\tilde{Y}_{\rm test} = i) - \mathbb{P}({Y}_{\rm test} = i) \right|}.
\end{equation}
Further suppose that $\Chatnoisy$ contains the most likely labels, i.e., $\forall x\in\mathcal{X}, i\in \Chatnoisy(x), j\notin \Chatnoisy(x): \mathbb{P}(\tilde{Y}_{\rm test}=i\mid \Xtest=x) \geq \mathbb{P}(\tilde{Y}_{\rm test}=j\mid \Xtest=x)$.
If the noise follows the \texttt{uniform noise} model, then the coverage rate is guaranteed to be valid:
\begin{equation}
1-\alpha \leq \mathbb{P}\left({Y}_{\rm test}\in \Chatnoisy(X_{\rm test})\right).
\end{equation}
\end{prop}  
\begin{figure}
    \centering
    \includegraphics[scale=0.6]{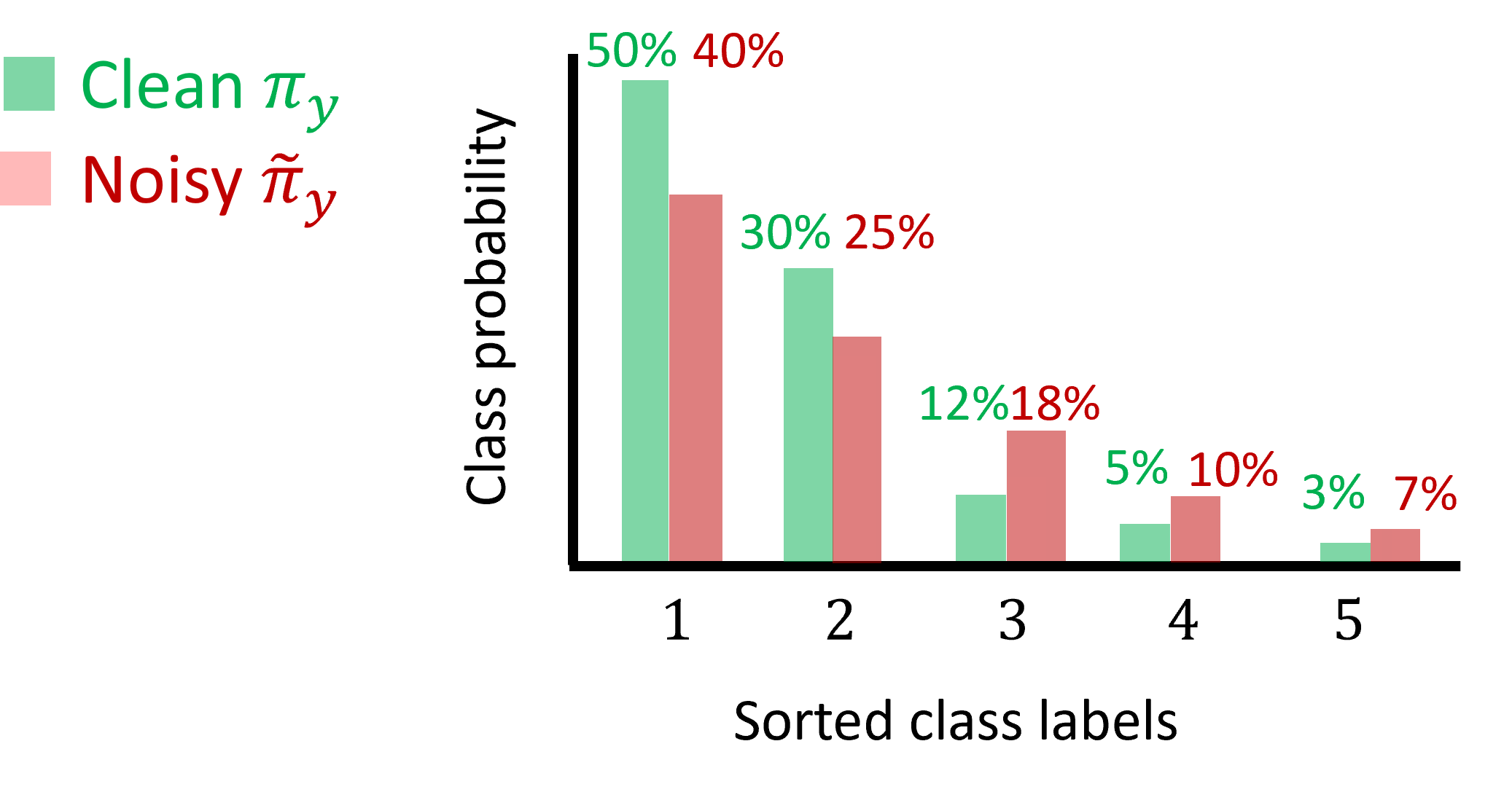}
    \caption{Clean (green) and noisy (red) class probabilities under dispersive corruption.}
    \label{fig:class_probability}
\end{figure}
We emphasize that the key contribution of Proposition~\ref{prop:general_classification} is the lower bound, as it guarantees a valid coverage rate. We also note that~\citet{barber2022conformal} provides a sharper upper bound for the coverage rate which also relies on the TV-distance between $Y_{\rm test}, \tilde{Y}_{\rm test}$. Nevertheless, the lower bound in Proposition~\ref{prop:general_classification} is tighter than the lower bound described in~\citet{barber2022conformal}. Figure~\ref{fig:class_probability} visualizes the essence of Proposition~\ref{prop:general_classification}, showing that as the corruption increases the label uncertainty, the prediction sets generated by conformal prediction get larger, having a conservative coverage rate. 
It should be noted that under the above proposition, achieving valid coverage requires only knowledge of the \emph{noisy} conditional distribution $\mathbb{P}\big(\tilY \; \mid \; X \big)$, so a model trained on a large amount of noisy data should approximately have the desired coverage rate. Moreover, one should notice that without the assumptions in Proposition~\ref{prop:general_classification}, even the oracle model is not guaranteed to achieve valid coverage, as we will show in Section~\ref{sec:syn_classification}.
For a more general analysis of classification problems where the noise model is an arbitrary confusion matrix, see Appendix~\ref{app:thrm_confusion}.

We now turn to examine two conformity scores and show that applying them with noisy data leads to conservative coverage, as a result of Proposition~\ref{prop:general_classification}. The \textit{adaptive prediction sets} (APS) score, first introduced by \cite{romano2020classification}, is defined as
\begin{equation}
    \label{eq:APS-score}
    s^{\rm APS}\left(x,y\right)=\sum_{y'\in\mathcal{Y}}\hat{{\pi}}_{y'}\left(x\right)\mathbb{I}\left\{\hat{{\pi}}_{y'}\left(x\right)>\hat{{\pi}}_{y}\left(x\right)\right\} +\hat{{\pi}}_{y}\left(x\right)\cdot U',
\end{equation}
where $\mathbb{I}$ is the indicator function, $\hat{{\pi}}_{y}\left(x\right)$ is the estimated conditional probability $\mathbb{P}\big(Y = y \; \mid \; X = x \big)$ and $U'\sim \mathrm{Unif}(0,1)$. 
To make this non-random, a variant of the above where $U'=1$ is often used. 
The APS score is one of two popular conformal methods for classification.
The other score from~\cite{vovk2005algorithmic,lei2013distribution}, is referred to as \emph{homogeneous prediction sets} (HPS) score, 
    $s^{\rm HPS}\left(x,y\right)=1-\hat{{\pi}}_{y}\left(x\right)$,
for some classifier $\hat{{\pi}}_y\left(x\right) \in [0,1]$.
The next corollary states that with access to an oracle classifier's ranking, conformal prediction covers the noiseless test label.
\begin{cor}
\label{cor:classification-HPS-APS}
    Let $\Chatnoisy(\Xtest)$ be constructed as in Recipe~\ref{recipe} with either the APS or HPS score functions, with any classifier that ranks the classes in the same order as the oracle classifier $\hat{{\pi}}_y(x) = \mathbb{P}\big(\tilY = y \; \mid \; X = x \big)$.
  Then,
  \begin{equation}
    1 - \alpha \leq \P\Big( \Ytest \in \Chatnoisy(\Xtest) \Big) \leq 1-\alpha+\frac{1}{n+1}+\frac{1}{2}\sum_{i=1}^{K} {\left|\mathbb{P}(\tilde{Y}_{\rm test} = i) - \mathbb{P}({Y}_{\rm test} = i) \right|}.
  \end{equation}
\end{cor}

We now turn to examine the specific \texttt{random flip} noise model in which the noisy label is randomly flipped $\epsilon$ fraction of the time. This noise model is well-studied in the literature; see, for example, \citep{aslam1996sample,angluin1988learning,ma2018dimensionality,jenni2018deep,jindal2016learning,yuan2018iterative}.
\begin{cor}\label{cor:random_flip_upper_bound}
    Let $\Chatnoisy(\Xtest)$ be constructed as in Recipe~\ref{recipe} with the corruption function $g^{\rm flip}$ and either the APS or HPS score functions, with any classifier that ranks the classes in the same order as the oracle classifier $\hat{{\pi}}_y(x) = \mathbb{P}\big(\tilY = y \; \mid \; X = x \big)$.
  Then,
  \begin{equation}
    1 - \alpha \leq \P\Big( \Ytest \in \Chatnoisy(\Xtest) \Big) \leq 1-\alpha+\frac{1}{n+1}+\epsilon\frac{K-1}{K}.
  \end{equation}
\end{cor}
Crucially, the above corollaries apply with any score function that preserves the order of the estimated classifier, which emphasizes the generality of our theory. All proofs are given in Appendix~\ref{app:gen-class}. 
Moreover, although this is not our main focus, in Appendix~\ref{set-sizes-analysis} we investigate the inflation of the prediction set size in the specific case of \texttt{random flip} noise with APS scores and the oracle model.
Table~\ref{tab:summary} summarizes all different settings we examine with their corresponding bounds.
Finally, we note that in Section~\ref{sec:multi_label_theory} we extend the above analysis to multi-label classification, where there may be multiple labels that correspond to the same sample.

\begin{table}[!htb]
\caption{Summary of coverage bounds for different scores and noise models}
\label{tab:summary}
\centering
\begin{tabular}{|c|c|c|c| }
\hline
Task & Score & Noise model & Bounds\\
\hline
\multirow{3}{6em}{Regression} & All & \makecell{Additive symmetric: \\$g^\text{add}$ in~(3)} & \makecell{$\P\Big( \Ytest \in \Chatnoisy(\Xtest) \Big) \geq 1-\alpha$\\ (Proposition 1)}\\ 
& All & \makecell{Additive symmetric: \\$g^\text{add}$ in~(3)}& \makecell{$\mathbb{P}(Y \in C(x) \mid X=x) \geq$\\ $\mathbb{P}(\Tilde{Y} \in C(x) \mid X=x)$ \\(Theorem A4)}\\
\hline
\multirow{3}{6em}{Classification} & All & \texttt{uniform noise} &\makecell{ $1-\alpha \leq \mathbb{P}\left({Y}_{\rm test}\in \Chatnoisy(X_{\rm test})\right) \leq $\\ $1-\alpha +\frac{1}{n+1}$\\ $+ \frac{1}{2}\sum_{i=1}^{K} {\left|\mathbb{P}(\tilde{Y}_{\rm test} = i) - \mathbb{P}({Y}_{\rm test} = i) \right|}$ \\(Proposition 2)}\\ 
& \makecell{HPS, \\APS} & \makecell{Random flip: \\$g^\text{flip}$ in~(6)} & \makecell{$1 - \alpha \leq \P\Big( \Ytest \in \Chatnoisy(\Xtest) \Big)\leq$\\$1-\alpha+\frac{1}{n+1}+\epsilon\frac{K-1}{K}$ \\(Corollary 2)}\\ 
\bottomrule
\end{tabular}
\end{table}

\subsection{Disclaimer: Distribution-Free Results}\label{sec:disclaimer}

Though the coverage guarantee holds in many realistic cases, conformal prediction may generate uncertainty sets that fail to cover the true outcome.
Indeed, in the general case, conformal prediction produces invalid prediction sets, and must be adjusted to account for the size of the noise. 
The following proposition states that for any nontrivial noise distribution, there exists a score function that breaks na\"ive conformal.
\begin{prop} [Coverage is impossible in the general case.] \label{prop:general-case}
    Take any $\tilY \overset{d}{\neq} Y$.
    Then there exists a score function $s$ that yields $\P\left( \Ytest \in \Chatnoisy(\Xtest) \right) < \P\left( \Ytest \in \Chat(\Xtest) \right)$, for $\Chatnoisy$ constructed using noisy samples and $\Chat$ constructed with clean samples.
\end{prop}
The above proposition says that for any noise distribution, there exists an adversarially chosen score function that will disrupt coverage.  
Furthermore, as we discuss in Appendix~\ref{app:general-case}, with a noise of a sufficient magnitude, it is possible to get arbitrarily bad violations of coverage.
In Appendix~\ref{app:general-case} we state an additional impossibility result in which we claim that for any given score function following some conditions, there is an adversarial noise that invalidates the coverage.

Next, we discuss how to adjust the threshold of conformal prediction to account for noise of a known size, as measured by total variation (TV) distance from the clean label.
\begin{cor}[Corollary of~\citet{barber2022conformal}]
    \label{cor:tv}
    Let $\tilY$ be any random variable satisfying $D_{\rm TV}(Y, \tilY) \leq \epsilon$.
    Take $\alpha' = \alpha + 2\frac{n}{n+1} \epsilon$. 
    Letting $\Chatnoisy(\Xtest)$ be the output of Recipe~\ref{recipe} with any score function at level $\alpha'$ yields
    \begin{equation}
        \P\big( \Ytest \in \Chatnoisy(\Xtest) \big) \geq 1-\alpha.
    \end{equation}
\end{cor}
We discuss this strategy more in Appendix~\ref{app:general-case}---the algorithm implied by Corollary~\ref{cor:tv} may not be particularly useful, as the TV distance is a badly behaved quantity that is also difficult to estimate, especially since the clean labels are inaccessible.

As a final note, if the noise is bounded in TV norm, then the coverage is also not too conservative.
\begin{cor}[Corollary of~\cite{barber2022conformal} Theorem 3]
    \label{cor:tv-ub}
    Let $\tilY$ be any random variable satisfying $D_{\rm TV}(Y, \tilY) \leq \xi$.
    Letting $\Chatnoisy(\Xtest)$ be the output of Recipe~\ref{recipe} with any score function at level $\alpha$ yields
    \begin{equation}
        \P\big( \Ytest \in \Chatnoisy(\Xtest) \big) < 1-\alpha+\frac{1}{n+1}+\frac{n}{n+1}\xi.
    \end{equation}
\end{cor}

\section{Risk Control Under Label Noise}
\subsection{Problem Setup}
Up to this point, we have focused on the miscoverage loss, and the analysis conducted thus far applies exclusively to this metric. 
However, in real-world applications, it is often desired to control metrics other than the binary loss $L^\text{miscoverage}(y, C) = \mathbbm{1}\{y\notin C\}$, where $C$ is a set of predicted labels. Examples of such alternative losses include the F1-score or the false negative rate. The latter is particularly relevant for high-dimensional response $Y$, as in tasks like multi-label classification or image segmentation.
To address this need, researchers have developed extensions of the conformal framework that go beyond the miscoverage loss, providing a rigorous \emph{risk control} guarantee for general loss functions \citep{bates2021distributionfree,angelopoulos2021learn, angelopoulos2022conformal}. 

Similarly to the conformal prediction algorithm, in the risk-control setting we post-process the predictions of a model $\hat{f}$ to create a prediction set $\widehat{\mathcal{C}}_\lambda(\Xtest)$ with a parameter $\lambda$ that determines its level of conservativeness: higher values yield larger and nested sets, in the sense that $\widehat{\mathcal{C}}_{\lambda_{2}}(\cdot) \subseteq \widehat{\mathcal{C}}_{\lambda_{1}}(\cdot) $ for $\lambda_2 \leq \lambda_1$.
For instance, if $\hat{f}_y(x)$ is an estimator of the conditional probability of $Y\mid X=x$, then the prediction sets can be defined as $\widehat{\mathcal{C}}_\lambda(\Xtest)=\{y : \hat{f}_y(\Xtest)>\lambda\}$. 
To measure the quality of $\widehat{\mathcal{C}}_\lambda(\Xtest)$, we consider a loss function $L(\Ytest, \Chat_\lambda(\Xtest))$ which we require to be non-increasing as a function of $\lambda$. This is analogous to conformal prediction in which the quantile of the scores, $\hat{q}$, encodes the prediction set sizes, and the error measure is simply the miscoverage loss. In the conformal prediction framework, we use a holdout set $\{ (X_i,Y_i) \}_{i=1}^n$ in order to calibrate $\hat{q}_{\rm clean}$ and achieve valid coverage over a new test point.
Likewise, in the conformal risk control settings we aim to use the observed losses $\{L(Y_i, \Chat_\lambda(X_i))\}_{i=1}^{n}$ derived by the calibration set to find a calibrated threshold $\hat{\lambda}_{\rm clean}$ that will control the risk of an unseen test point at a pre-specified level $\alpha$:
\begin{equation}
    \mathcal{R}(C) = \mathbb{E}[L(\Ytest, \Chat_{\hat{\lambda}_{\rm clean}}(\Xtest))] \leq \alpha.
\end{equation}
See Appendix~\ref{sec:risk_alg} for the conformal risk control procedure. Analogously to conformal prediction, these methods produce valid sets under the i.i.d. assumption, but their guarantees do not hold in the presence of label noise. 
Provided a noisy calibration set, $\{ (X_i,\tilde{Y}_i) \}_{i=1}^n$, the parameter $\hat{\lambda}_{\rm noisy}$ is constructed using the noisy losses $\{L(\tilde{Y}_i, \Chat_\lambda(X_i))\}_{i=1}^{n}$, and therefore the risk of a new clean test point is not guaranteed to be controlled. 

Our main goal is to delineate when it is possible to provide a risk control guarantee of the form
\begin{equation}
    \mathbb{E}[L(\Ytest, \Chat_{\hat{\lambda}_{\rm noisy}}(\Xtest))] \leq \alpha.
\end{equation}

In the next section, we conduct a real-data experiment, demonstrating that conformal risk control is robust to label noise.
In Section~\ref{risk_theoty} we explain this phenomenon and specify the conditions under which conservative risk is obtained.
To ease notation, we will omit the subscript $\lambda$ but directly specify whether the prediction set is constructed by clean or noisy calibration data.

\subsection{Empirical Evidence}
\subsubsection*{Multi-label Classification}\label{sec:real_coco_noise_exp}

In this section, we analyze the robustness of conformal risk control  \citep{angelopoulos2022conformal} to label noise in a multi-label classification setting. We use the MS COCO data set \citep{lin2014microsoft}, in which the input image may contain up to $K=80$ positive labels, i.e., $Y\subseteq\{1,2,...,K\}$. 
In the following experiment, we consider the annotations in this data set as ground-truth labels. We have collected noisy labels from individual annotators who annotated 117 images in total. On average, the annotators missed or mistakenly added approximately 1.75 labels from each image. See Appendix~\ref{sec:real_coco_noise_details} for additional details about this experimental setup and data collection.

\begin{figure}[ht]
    \centering
    \includegraphics[height=0.28\linewidth]{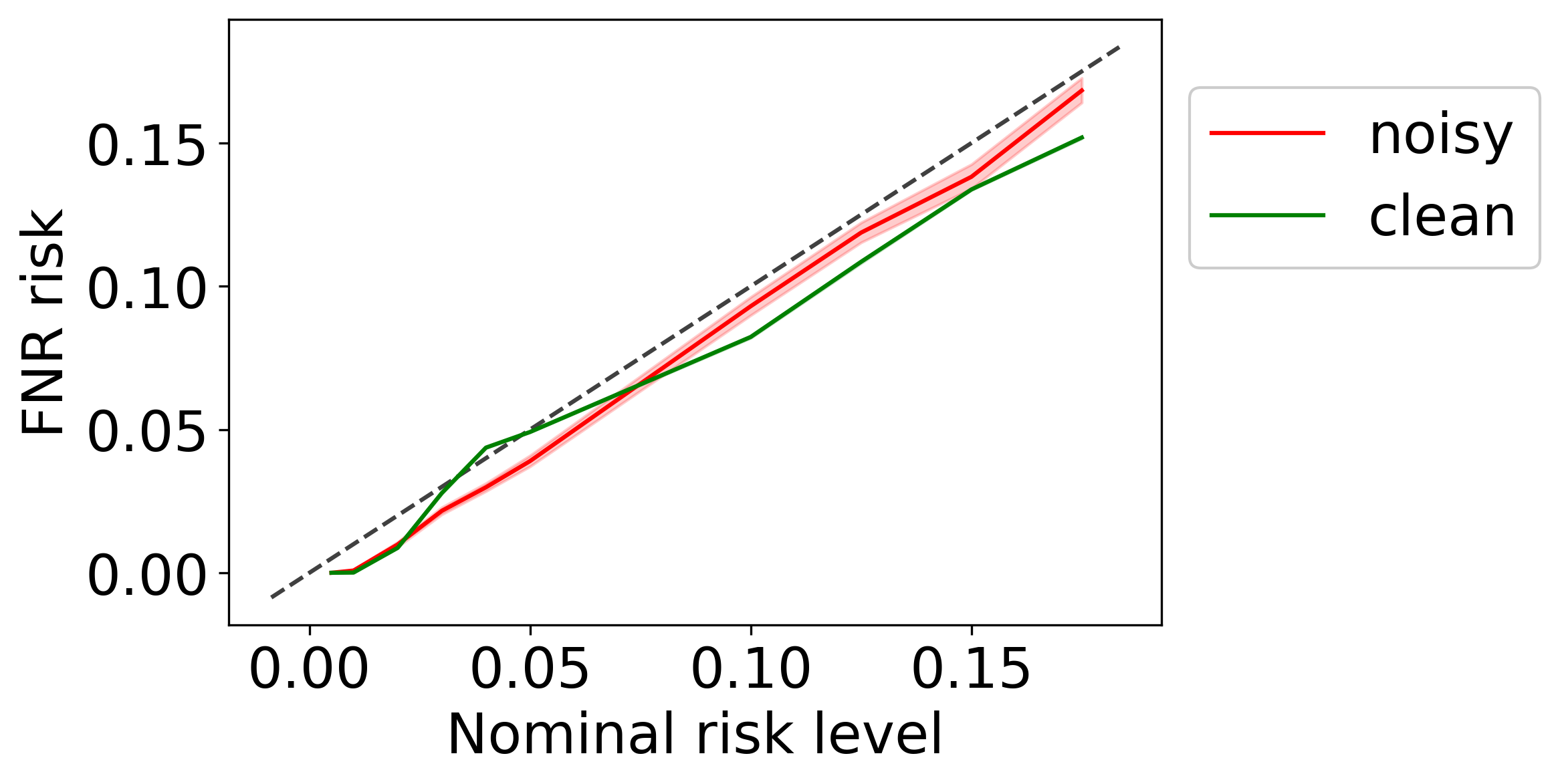}
    \includegraphics[height=0.28\linewidth]{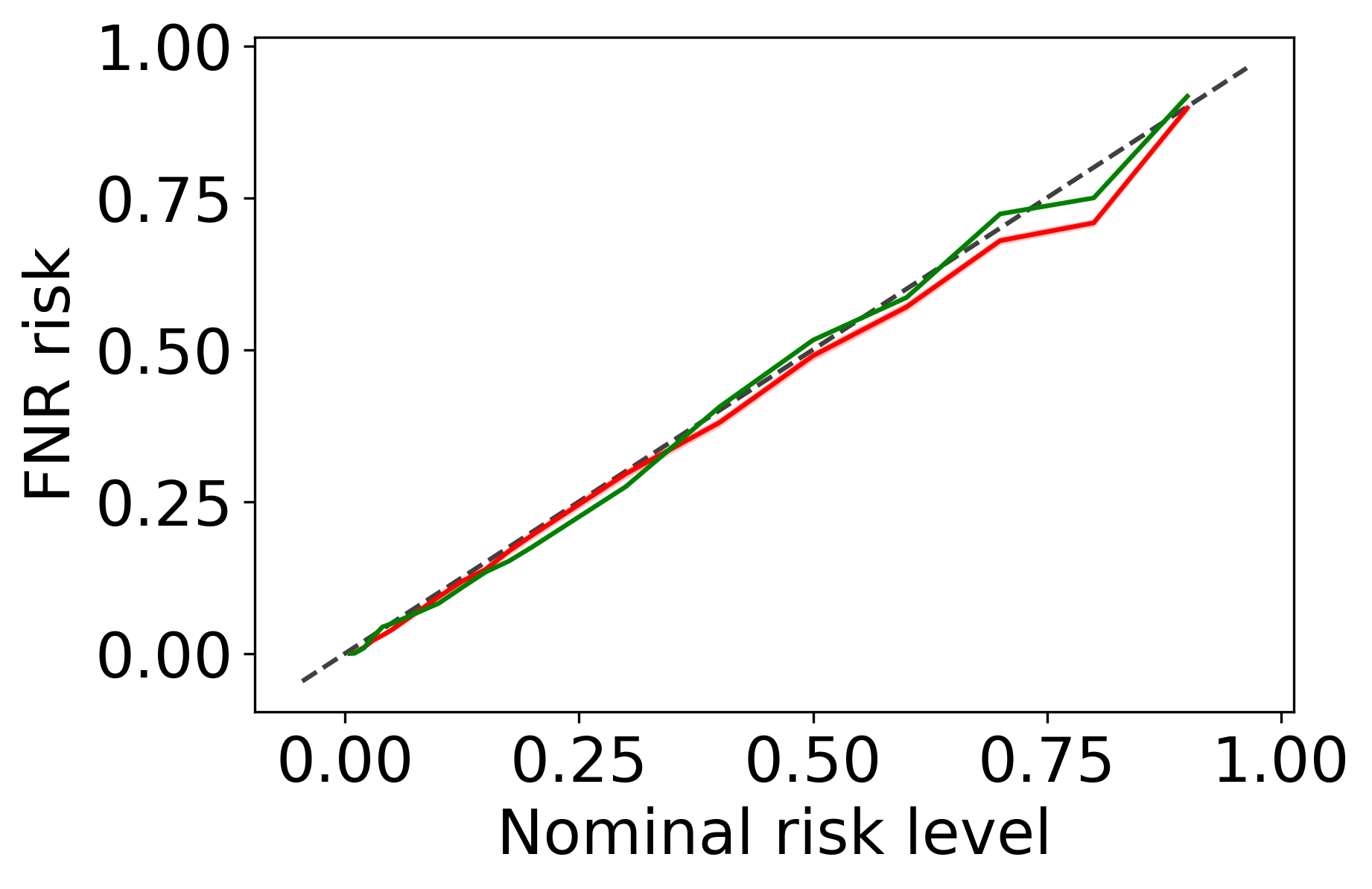}
    \caption{FNR on MS COCO data set, achieved over noisy (red) and clean (green) test sets. The calibration scheme is applied with noisy annotations. Results are averaged over 2000 trials.}
    \label{fig:real_multi_label_exp}
\end{figure}

We fit a TResNet \citep{ridnik2021tresnet} model on 100k clean samples and calibrate it using 105 noisy samples with conformal risk control, as introduced in \cite[Section 3.2]{angelopoulos2022conformal}, to control the false-negative rate (FNR) at different levels. The FNR is the ratio of positive labels of $Y$ that are missed in a prediction set $C$, formally defined as:
\begin{equation}\label{eq:fnr}
L^{\rm FNR}(Y,C) = 1- \frac{|Y \cap C|}{|Y|}.
\end{equation}
We measure the FNR obtained over the clean and noisy test sets which contain 40k and 12 samples, respectively. Figure~\ref{fig:real_multi_label_exp} displays the results, showing that the uncertainty sets attain valid risk even though they were calibrated using corrupted data. In the following section, we aim to explain these results and find the conditions under which label-noise robustness is guaranteed.

\subsection{Theoretical Analysis}
\label{risk_theoty}
\subsubsection{Multi-label Classification}\label{sec:multi_label_theory}
In this section, we study the conditions under which conformal risk control is robust to label noise in a multi-label classification setting. Recall that each sample contains up to $K$ positive labels, i.e., $Y\subseteq\{1,2,...,K\}$. Here, we assume a vector-flip noise model with a binary random variable $\epsilon_i$ that flips the $i$-th label with probability $\mathbb{P}(\epsilon_i=1)$:
\begin{equation}\label{eq:vector_flip}
g^{\rm vector-flip}(y)_i = y_i (1-\epsilon_i) + (1-y_i) \epsilon_i. 
\end{equation}
Above, $y_i$ is an indicator that takes the value 1 if the $i$-th label is present in $y$, and 0, otherwise. Notice that the random variable $\epsilon_i$ takes the value 1 if the $i$-th label in $y$ is flipped and 0 otherwise. We further assume that the noise is not adversarial, i.e., $\mathbb{P}(\epsilon_i=1) < 0.5$ for all $i\in\{1,2,...,K\}$. We now show that a valid FNR risk is guaranteed in the presence of label noise under the following assumptions.
\begin{prop}\label{prop:multi_label}
Let $\Chatnoisy(\Xtest)$ be a prediction set that contains the most likely labels, in the sense of Proposition~\ref{prop:general_classification}, and controls the FNR risk of the noisy labels at level $\alpha$. Assume the multi-label noise model $g^{\rm vector-flip}$. If 
\begin{enumerate}
    \item $\Ytest$ is a deterministic function of $\Xtest$,
    \item The number of positive labels, i.e., $|\tilde{Y}_{\rm test}| \mid \Xtest=x$ is a constant for all $x\in\mathcal{X}$,
\end{enumerate}
  then
  \begin{equation}
    \mathbb{E}\left[L^{\rm FNR}\left(\Ytest,\Chatnoisy(\Xtest)\right)\right] \leq \alpha.
  \end{equation}
\end{prop}
\begin{remark}[Dependent corruptions]
Proposition~\ref{prop:multi_label} holds even if the elements in the noise vector $\epsilon$ depend on each other or on $X$.
\end{remark}
The proof and other additional theoretical results are provided in Appendices~\ref{sec:additional_fnr_results}-~\ref{app:proof-multi-label}. Importantly, the determinism assumption on $Y\mid X=x$ is reasonable as it is simply satisfied when the noiseless response is defined as the consensus outcome.
Nevertheless, this assumption may not always hold in practice.
Thus, in the next section, we propose alternative requirements for the validity of the FNR risk and demonstrate them in a segmentation setting. 

\subsubsection{Segmentation}

In segmentation tasks, the goal is to 
assign labels to every pixel in an input image such that pixels with similar characteristics share the same label. For example, tumor segmentation can be applied to identify polyps in medical images. 
Here, the response is a binary matrix $Y\subseteq \{0,1\}^{W\times H}$ that contains the value $1$ in the $(i,j)$ pixel if it includes the object of interest and $0$ otherwise. The uncertainty is represented by a prediction set $C \subseteq \{1,...,W \} \times \{1,...,H \}$ that includes pixels that are likely to contain the object. Similarly to the multi-label classification problem, here, we assume a vector flip noise model $g^{\rm vector-flip}$, that flips the $(i,j)$ pixel in $Y$, denoted as $Y_{i,j}$, with probability $\mathbb{P}(\epsilon_{i,j}=1)$. We now show that the prediction sets constructed using a noisy calibration set are guaranteed to have conservative FNR if the clean response matrix $Y$ and the noise variable $\epsilon$ are independent given $X$.
\begin{prop}\label{prop:segmentation}
Let $\Chatnoisy(\Xtest)$ be a prediction set that contains the most likely pixels, in the sense of Proposition~\ref{prop:general_classification}, and controls the FNR risk of the noisy labels at level $\alpha$. 
Suppose that:
\begin{enumerate}
  \item The elements of the clean response matrix are independent of each other given $\Xtest$. That is, ${\Ytest}^{i,j}\mid \Xtest=x \indep {\Ytest}^{m,n}\mid \Xtest=x$ for all $(i,j)\ne (m,n)\subseteq \{1,...,W\}\times\{1,...,H\}$ and $x\in \mathcal{X}$.
  \item For a given input $\Xtest$, the noise level is the same for all response elements.
  \item The noise variable $\epsilon$ is independent of $\Ytest\mid \Xtest=x$, and the noises of different labels are independent of each other given $\Xtest$, similarly to condition 1.
\end{enumerate}
  Then,
  \begin{equation}
    \mathbb{E}\left[L^{\rm FNR}\left(\Ytest,\Chatnoisy(\Xtest)\right)\right] \leq \alpha.
  \end{equation}
\end{prop}
We note that a stronger version of this proposition that allows dependence between the elements in $\Ytest$ is given in Appendix~\ref{sec:additional_fnr_results}, and the proof is in Appendix~\ref{app:proof-segmentation}.
The advantage of Proposition~\ref{prop:segmentation} over Proposition~\ref{prop:multi_label} is that here, the response matrix $\Ytest$ and the number of positive labels are allowed to be stochastic. For this reason, we believe that Proposition~\ref{prop:segmentation} is more suited for segmentation tasks, even though Proposition~\ref{prop:multi_label} applies in segmentation settings as well.

\subsubsection{Regression with a General Loss}\label{sec:risk_regression}

This section studies the general regression setting in which $Y\in\mathbb{R}$ takes continuous values and the loss function is an arbitrary function $L(y,\Chatnoisy(x))\in \mathbb{R}$. Here, our objective is to find tight bounds for the risk over the clean labels using the risk observed over the corrupted labels. The main result of this section accomplishes this goal while making minimal assumptions on the loss function, the noise model, and the data distribution.
\begin{prop}\label{prop:general_regression_bounds}
Let $\Chatnoisy(\Xtest)$ be a prediction interval that controls the risk at level $\alpha:= \mathbb{E}\left[L(\tilde{Y}_{\rm test}, \Chatnoisy(\Xtest)\right]$. Suppose that the second derivative of the loss $L(y;C)$ is bounded for all $y,C$: $q \leq  \frac{\partial^2}{\partial y^2} L(y;C) \leq Q$ for some $q,Q\in\mathbb{R}$. If the labels are corrupted by the function $g^{\rm add}$ from~\eqref{eq:reg_g_sym} with a noise $Z$ that satisfies $\mathbb{E}[Z] = 0$, then
\begin{equation}
\begin{split}
\alpha - \frac{1}{2}Q \cdot \textup{Var}(Z) \leq \mathbb{E}\left[L(\Ytest, \Chatnoisy(\Xtest)\right]
\leq 
\alpha - \frac{1}{2}q \cdot \textup{Var}(Z).
\end{split}
\end{equation}
If we further assume that $L$ is convex then we obtain valid risk:
\begin{equation}
\mathbb{E}\left[L(\Ytest, \Chatnoisy(\Xtest)\right] \leq \alpha.
\end{equation}
\end{prop}
The proof is detailed in Appendix~\ref{sec:regression_taylor_risk_bounds}. Remarkably, Proposition~\ref{prop:general_regression_bounds} applies for any predictive model, calibration scheme, distribution of $Y\mid X$, and loss function that is twice differentiable. The only requirement is that the noise must be additive and with zero mean. We now demonstrate this result on a smooth approximation of the miscoverage loss, formulated as:
\begin{equation}\label{eq:smooth_miscoverage}
    L^{sm}(y,[a,b]) = \frac{2}{1+e^{-\left(2\frac{y-a}{b-a}-1\right)^2}}
\end{equation}
\begin{cor}\label{cor:smooth_miscoverage}
Let $\Chatnoisy(\Xtest)$ be a prediction interval. Denote the smooth miscoverage over the noisy labels as $\alpha := \mathbb{E}\left[L^{sm}(\tilde{Y}_{\rm test}, \Chatnoisy(\Xtest)\right]$. Under the assumptions of Proposition~\ref{prop:general_regression_bounds}, the risk of the clean labels is bounded by
\begin{equation}
\alpha - \frac{1}{2}Q\cdot \textup{Var}(Z) \leq \mathbb{E}\left[L^{sm}(\Ytest, \Chatnoisy(\Xtest)\right] \leq \alpha - \frac{1}{2}q \cdot \textup{Var}(Z),
\end{equation}
where $q=\mathbb{E}_X \left[\min_{y}{ {\frac{\partial^2}{\partial y^2} L^{sm}}(y,\Chatnoisy(X))}\right]$ and $Q=\mathbb{E}_X \left[\max_{y}{ {\frac{\partial^2}{\partial y^2} L^{sm}}(y,\Chatnoisy(X))}\right]$ are known constants. 
\end{cor}
We now build upon Corollary~\ref{cor:smooth_miscoverage} and establish a lower bound for the coverage rate achieved by intervals calibrated with corrupted labels.
\begin{prop}\label{prop:cov_bound}
Let $\Chatnoisy(\Xtest)$ be a prediction interval. Suppose that the labels are corrupted by the function $g^{\rm add}$ from~\eqref{eq:reg_g_sym} with a noise $Z$ that satisfies $\mathbb{E}[Z] = 0$, then
\begin{equation}
\P\Big( \Ytest \in \Chatnoisy(\Xtest) \Big) \geq 1-\frac{\mathbb{E}[L^{\rm sm}(\tilde{Y}_{\rm test}, \Chatnoisy(\Xtest))] - 0.5 P \cdot \textup{Var}(Z)}{Q},
\end{equation}
where $P,Q$ are tunable constants.
\end{prop}
In Appendix~\ref{sec:regression_taylor_coverage_bounds} we give additional details about this result as well as formulate a stronger version of it that provides a tighter coverage bound. Finally, in Appendix~\ref{sec:smoothnes_regression_miscoverage_bound} we provide an additional miscoverage bound which is more informative and tight for smooth densities of $Y\mid X=x$. Table~\ref{tab:summary-gen-risk} summarizes all different risk-control settings with their corresponding bounds. Lastly, in Appendix~\ref{sec:image_miscoverage} we analyze label-noise robustness in settings where the response $Y$ is a matrix, as in image-to-image regression tasks.

 


\begin{table}[!htb]
  \caption{Summary of coverage bounds for different risk control tasks and different noise models}
\label{tab:summary-gen-risk}
\centering
\begin{tabular}{|c|c|c|} 
\hline
Task & Noise model  & Bounds\\
\hline
\makecell{Multi-label\\ classification} & \makecell{Random multi-label flip:\\ $g^\text{vector-flip}$ in~(8)} & \makecell{$\mathbb{E}\left[L^{\rm FNR}\left(\Ytest,\Chatnoisy(\Xtest)\right)\right] \leq \alpha$ \\(Proposition 4)}\\ 
\hline
Segmentation & \makecell{Random multi-label flip: \\$g^\text{vector-flip}$ in~(8)}&\makecell{ $\mathbb{E}\left[L^{\rm FNR}\left(\Ytest,\Chatnoisy(\Xtest)\right)\right] \leq \alpha$ \\(Proposition 5)}\\ 
\hline
\makecell{Regression with a \\general loss}
& Additive: $g^{\text{add}}$ in~(3)&\makecell{ $\mathbb{E}\left[L\left(\Ytest,\Chatnoisy(\Xtest)\right)\right] \leq \alpha$ \\(Proposition 6)}\\ 
\hline
\makecell{Regression with \\miscoverage loss}
& Additive: $g^{\text{add}}$ in~(3)& \makecell{$\P\Big( \Ytest \in \Chatnoisy(\Xtest) \Big) \geq$ \\ $1-\frac{\mathbb{E}[L^{\rm sm}(\tilde{Y}_{\rm test}, \Chatnoisy(\Xtest))] - 0.5 P \cdot \textup{Var}(Z)}{Q}$ \\(Proposition 7)}\\
 \bottomrule
\end{tabular}
\end{table}

\subsection{Online Learning Under Label Noise}\label{sec:online_thoery}

In this section, we focus on an online learning setting and show that all theoretical results presented thus far also apply to the online framework.
Here, the data is given as a stream ${(X_t, \tilde{Y}_t)}_{t\in\mathbb{N}}$ in a sequential
fashion. Crucially, we have access only to the noisy labels $\tilde{Y}_t$, and the clean labels $Y_t$ are unavailable throughout the entire learning process.
At time stamp $t\in\mathbb{N}$, our goal is to construct a prediction set $\Chatnoisyt$ given on all previously observed samples ${(X_{t'}, \tilde{Y}_{t'})}_{t'=1}^{t-1}$ along with the test feature vector $X_t$ that achieves a long-range risk controlled at a user-specified level $\alpha$, i.e.,
\begin{equation}\label{eq:online_risk_requirement}
\mathcal{R}(\Chat) = \lim_{T \to \infty} \frac{1}{T} \sum_{t=1}^T {L_t(Y_t, \Chatnoisyt(X_t))} = \alpha.
\end{equation} 
Importantly, in this online learning setting, the loss function $L_t$ might be time-dependent and may vary throughout the learning process. There have been developed calibration schemes that generate uncertainty sets with statistical guarantees in online settings in the sense of~\eqref{eq:online_risk_requirement}. A popular approach is Adaptive conformal inference (\texttt{ACI}), proposed by \cite{aci}, which is an innovative online calibration scheme that constructs prediction sets with a pre-specified coverage rate, in the sense of~\eqref{eq:online_risk_requirement} with the choice of the miscoverage loss. In contrast to \texttt{ACI}, Rolling risk control (\texttt{Rolling RC}) \citep{feldman2022achieving} extends \texttt{ACI} by providing a guaranteed control of a general risk that may go beyond the binary loss. 
The main idea behind both these methods is to tune the calibration parameter that controls the size of the prediction set according to the coverage or risk level achieved in the past. See Appendix~\ref{sec:aci_alg} and Appendix~\ref{sec:rrc_alg} for more details on \texttt{ACI} and \texttt{RRC}, respectively.
Yet, the guarantees of these approaches are invalidated when applied using corrupted data.
Nonetheless, we argue that uncertainty sets constructed using corrupted data attain conservative risk in online settings under the requirements for offline label-noise robustness presented thus far.
\begin{prop}\label{prop:online}
Suppose that for all $t\in\mathbb{N}$ and $x \in \mathcal{X}$ we have a conservative loss:
\begin{equation}
\mathbb{E}\left[L_t\left(Y_t,\Chatnoisyt(X_t)\right) \middle| X_t=x \right] \leq \mathbb{E}\left[L_t\left(\tilde{Y}_t,\Chatnoisyt(X_t)\right) \middle| X_t=x\right].
\end{equation}
Then,
\begin{equation}
\lim_{T \rightarrow \infty} {\frac{1}{T}\sum_{t=1}^T {L_t(Y_t, \Chatnoisyt(X_t))}} \leq \alpha,
\end{equation}
where $\alpha$ is the risk computed over the noisy labels.
\end{prop}
The proof is given in Appendix~\ref{sec:online_risk}. In words, Proposition~\ref{prop:online} states that if the expected loss at every timestamp is conservative, then the risk over long-range windows in time is guaranteed to be valid. Practically, this proposition shows that all theoretical results presented thus far apply also in the online learning setting. We now demonstrate this result in two settings: online classification and segmentation and show that valid risk is obtained under the assumptions of Proposition~\ref{prop:general_classification} and Proposition~\ref{prop:segmentation}, respectively.
\begin{cor}[Valid risk in online classification settings]
Suppose that the distributions of $Y_t\mid X_t$ and $\tilde{Y}_t \mid X_t$ satisfy the assumptions in Proposition~\ref{prop:general_classification} for all $t\in\mathbb{N}$. If $\Chatnoisyt(x)$ contains the most likely labels, in the sense of Proposition~\ref{prop:general_classification}, for every $t\in\mathbb{N}$ and $x\in\mathcal{X}$, 
then
\begin{equation}
\lim_{T \rightarrow \infty} {\frac{1}{T}\sum_{t=1}^T {\mathbbm{1}\{Y_t\notin \Chatnoisyt(X_t)\}}} \leq \alpha.
\end{equation}
\end{cor}

\begin{cor}[Valid risk in online segmentation settings]
Suppose that the distributions of $Y_t\mid X_t$ and $\tilde{Y}_t \mid X_t$ satisfy the assumptions in Proposition~\ref{prop:segmentation} for all $t\in\mathbb{N}$. If $\Chatnoisyt(x)$ contains the most likely labels, in the sense of Proposition~\ref{prop:segmentation}, for every $t\in\mathbb{N}$ and $x\in\mathcal{X}$ 
then
\begin{equation}
\lim_{T \rightarrow \infty} {\frac{1}{T}\sum_{t=1}^T {L^\text{FNR}(Y_t, \Chatnoisyt(X_t))}} \leq \alpha.
\end{equation}
\end{cor}
Finally, in Appendix~\ref{sec:miscoverage_counter} we analyze the effect of label noise on the \emph{miscoverage counter} loss \citep{feldman2022achieving}. This loss assesses conditional validity in online settings by counting occurrences of consecutive miscoverage events. In a nutshell, Proposition~\ref{prop:miscoverage_counter} claims that with access to corrupted labels, the miscoverage counter is valid when the miscoverage risk is valid. This is an interesting result, as it connects the validity of the miscoverage counter to the validity of the miscoverage loss, where the latter is guaranteed under the conditions established in Section~\ref{sec:conformal_analysis}.

\section{Experiments}
Software is available online at \url{https://github.com/bat-sheva/Conformal-Label-Noise}, with all code needed to reproduce the numerical experiments.

\subsection{Synthetic Classification}\label{sec:syn_classification}
In this section, we focus on multi-class classification problems, where we study the validity of conformal prediction using different types of label noise distributions, described below. 
        \paragraph{Class-independent noise.}
        This noise model, which we call \texttt{uniform flip}, randomly flips the ground truth label into a \emph{different} one with probability $\epsilon$. Notice that this noise model slightly differs from the random flip $g^{\text{flip}}$ from~\eqref{eq:epsilon-class-corruption}, since in this \texttt{uniform flip} setting, a label cannot be flipped to the original label. Nonetheless, Proposition~\ref{prop:general_classification} states that the coverage achieved by an oracle classifier is guaranteed to increase in this setting as well.
        \paragraph{Class-dependent noise.}
        In contrast to \texttt{uniform flip} noise, here, we consider a more challenging setup in which the probability of a label to be flipped depends on the ground truth class label $Y$. Such a noise label is often called Noisy at Random (NAR) in the literature, where certain classes are more likely to be mislabeled or confused with similar ones.
        Let $T$ be a row stochastic transition matrix of size $K \times K$ such that $T_{i,j}$ is the probability of a point with label $i$ to be swapped with label $j$. In what follows, we consider three possible strategies for building the transition matrix $T$.
        (1) \texttt{Confusion matrix}~\citep{algan2020label}: we define $T$ as the oracle classifier's confusion matrix, up to a proper normalization to ensure the total flipping probability is $\epsilon$. We provide a theoretical study of this case in Appendix~\ref{app:thrm_confusion}.
        (2) \texttt{Rare to most frequent class}~\citep{xu2019l_dmi}: here, we flip the labels of the least frequent class with those of the most frequent class. This noise model is not uncommon in medical applications: imagine a setting where only a small fraction of the observations are abnormal, and thus likely to be annotated as the normal class.  If switching between the rare and most frequent labels does not lead to a total probability of $\epsilon$, we move to the next least common class, and so on.

\begin{figure}[ht]
\begin{center}
            {\includegraphics[width=.7\linewidth]{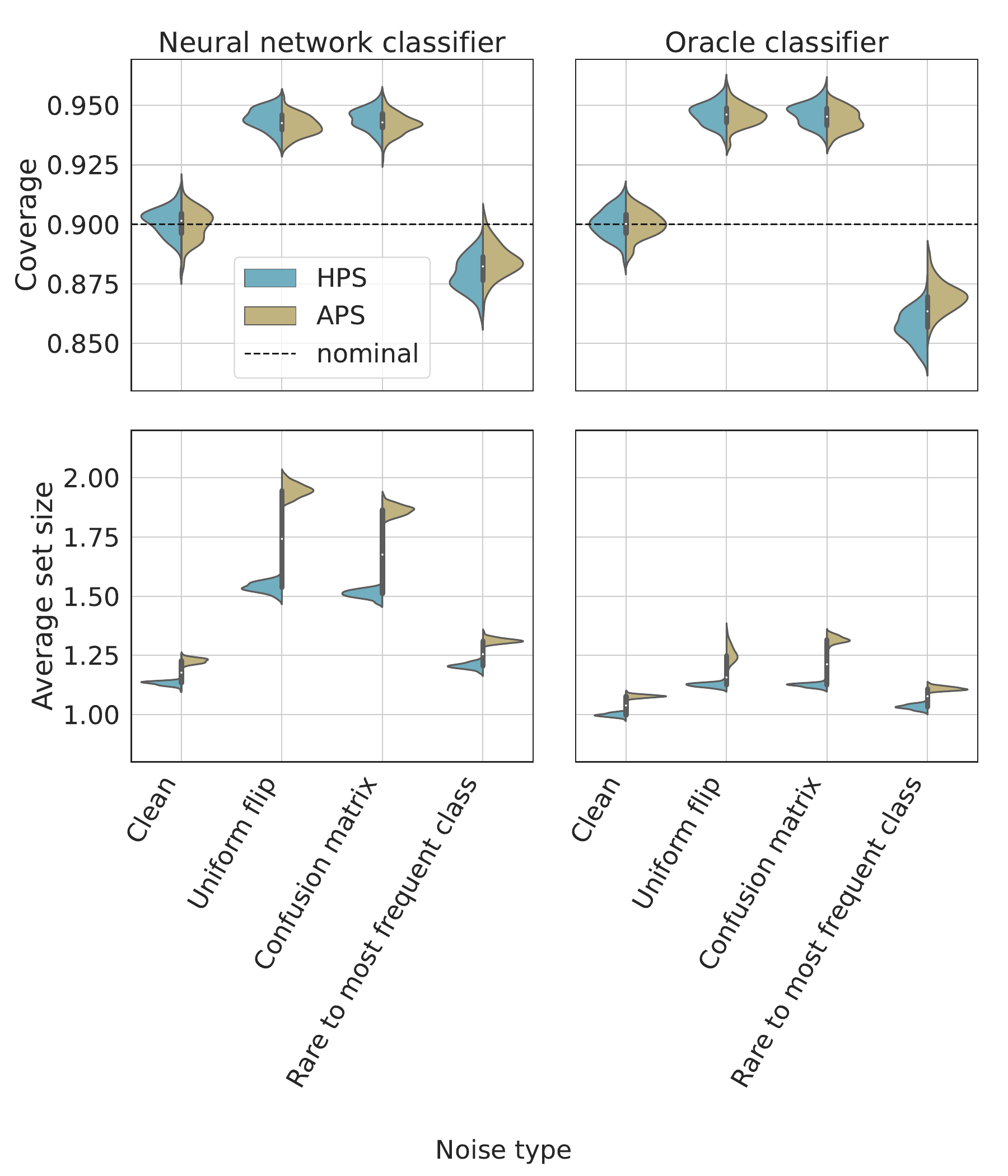}}\par 
            \caption{\textbf{Effect of label noise on synthetic multi-class classification data}. Performance of conformal prediction sets with target coverage $1-\alpha = 90\% $, using a noisy training set and a noisy calibration set. Top: Marginal coverage; Bottom: Average size of predicted sets. The results are evaluated over 100 independent experiments and the gray bar represents the interquartile range.}
            \label{fig:syntethic_classification}
        \end{center}
        \vspace{-0.5cm}
        \end{figure}

To set the stage for the experiments, we generate synthetic data with $K=10$ classes as follows. The features $X\in\mathbb{R}^d$ follow a standard multivariate Gaussian distribution of dimension $d=100$. The conditional distribution of $Y \mid X$ is multinomial with weights $w_j(x)=\exp((x^\top B)_j)/\sum_{i=1}^{K}\exp((x^\top B)_i)$, where $B\in\mathbb{R}^{d\times K}$ whose entries are sampled independently from the standard normal distribution. In our experiments, we generate a total of $60,000$ data points, where $50,000$ are used to fit a classifier, and the remaining ones are randomly split to form calibration and test sets, each of size $5,000$. The training and calibration data are corrupted using the label noise models we defined earlier, with a fixed flipping probability of $\epsilon=0.05$. Of course, the test set is not corrupted and contains the ground truth labels.
We apply conformal prediction using both the HPS and the APS score functions, with a target coverage level $1-\alpha$ of $90\%$. We use two predictive models: a two-layer neural network and an oracle classifier that has access to the conditional distribution of $\tilde{Y} \mid X$.
Finally, we report the distribution of the coverage rate as in~\eqref{eq:goal} and the prediction set sizes across $100$ random splits of the calibration and test data. As a point of reference, we repeat the same experimental protocol described above on clean data; in this case, we do not violate the i.i.d. assumption required to grant the marginal coverage guarantee in~\eqref{eq:marginal-coverage}.

The results are depicted in Figure~\ref{fig:syntethic_classification}. As expected, in the clean setting all conformal methods achieve the desired coverage of $90\%$. Under the \texttt{uniform flip} noise model, the coverage of the oracle classifier increases to around $94\%$, supporting our theoretical results from Section~\ref{sec:classification_analysis}. The neural network model follows a similar trend. Although not supported by a theoretical guarantee, when corrupting the labels using the more challenging \texttt{confusion matrix} noise, we can see a conservative behavior similar to the \texttt{uniform flip}. By contrast, under the  \texttt{rare to most frequent class} noise model, we can see a decrease in coverage, which is in line with our disclaimer from Section~\ref{sec:disclaimer}. Yet, observe how the APS score tends to be more robust to label noise than HPS, which emphasizes the role of the score function.

In Appendix~\ref{app:adv-supp} we provide additional experiments with adversarial noise models that more aggressively reduce the coverage rate. Such adversarial cases are more pathological and less likely to occur in real-world settings, unless facing a malicious attacker.

\subsection{Regression}
\label{sec:synthetic-reg}

\begin{figure}[ht]
\begin{center}
{\includegraphics[width=.5\linewidth]{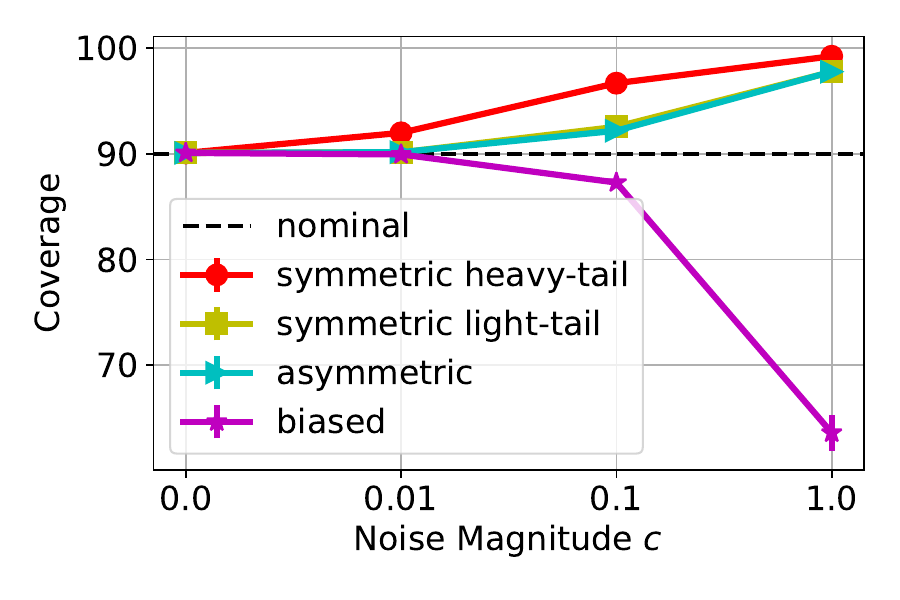}}\hfill
{\includegraphics[width=.5\linewidth]{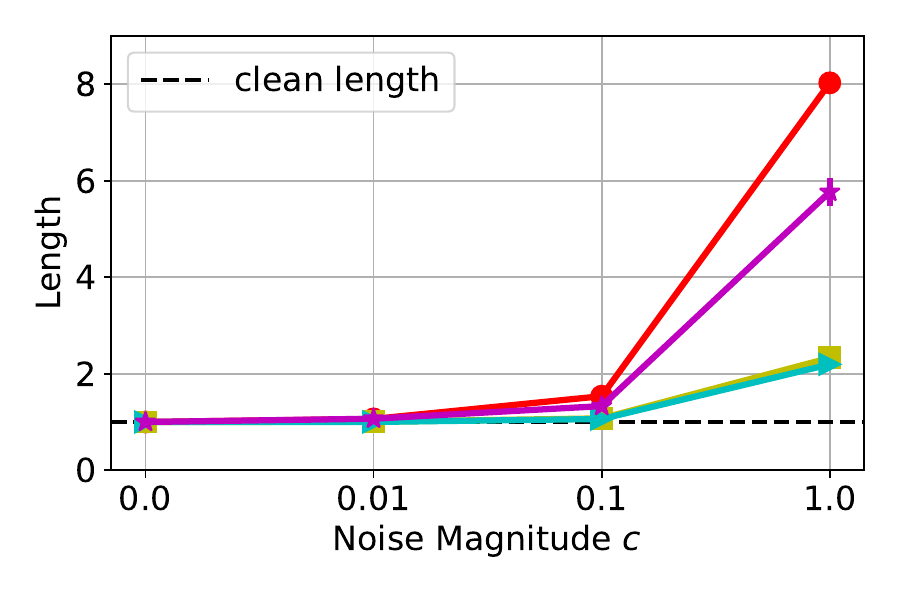}}
\caption{\textbf{Response-independent noise.} Performance of conformal prediction intervals with target coverage $1-\alpha=90\%$, using a noisy training set and a noisy calibration set. Left: Marginal coverage; Right: Length of predicted intervals (divided by the average clean length) using symmetric, asymmetric and biased noise with a varying magnitude. The results are evaluated over 50 independent experiments, with error bars showing one standard deviation. 
The standard deviation of $Y$ (without noise) and the square root of $\mathbb{E}\left[ \textnormal{Var}\left[ Y \mid X \right] \right]$ are both approximately 2.6.}
\label{fig:sym_noisy_train}
\end{center}
\vspace{-0.5cm}
\end{figure}

Similarly to the classification experiments, we study two types of noise distributions.
\paragraph{Response-independent noise.} We consider an additive noise of the form: $\tilde{Y}=Y+c\cdot Z$, where $c$ is a parameter that allows us to control the noise level. The noise component $Z$ is a random variable sampled from the following distributions.
(1) \texttt{Symmetric light tailed}: standard normal distribution; 
(2) \texttt{Symmetric heavy tailed}: t-distribution with one degree of freedom;
(3) \texttt{Asymmetric}: standard Gumbel distribution, normalized to have zero mean and unit variance; and
(4) \texttt{Biased}: positive noise formulated as the absolute value of the \texttt{symmetric heavy tailed noise} above. 

\paragraph{Response-dependent noise.} Analogously to the class-dependent noise from Section~\ref{sec:syn_classification}, we define more challenging noise models as follows.
(1) \texttt{Contractive}: this corruption pushes the ground truth response variables towards their mean. Formally, $\tilde{Y}_i=Y_i-\left( Y_i-\frac{1}{n} \sum_{i=1}^{n}Y_i\right)\cdot U$, where $U$ is a random uniform variable defined on the segment [0,0.5], and $n$ is the number of calibration points.
(2) \texttt{Dispersive}: this noise introduces some form of a dispersion effect on the ground truth response, which takes the opposite form of the \texttt{contractive} model, given by $\tilde{Y}_i=Y_i+\left( Y_i-\frac{1}{n} \sum_{i=1}^{n}Y_i\right)\cdot U$.

Having defined the noise models, we turn to describe the data-generating process. We simulate a 100-dimensional $X$ whose entries are sampled independently from a uniform distribution on the segment $[0, 5]$. Following~\cite{romano2019conformalized}, the response variable is generated as follows:
\begin{equation} \label{eq:reg_syn_model}
    Y \sim \textrm{Pois}(\sin^{2}(\bar{X})+0.1)+0.03\cdot \bar{X}\cdot \eta_{1} + 25\cdot \eta_{2} \cdot \mathbbm{1} \left\{ U<0.01\right\},
\end{equation}
where $\bar{X}$ is the mean of the vector $X$, and $\textrm{Pois}(\lambda)$ is the Poisson distribution with mean $\lambda$. Both $\eta_{1}$ and $\eta_{2}$ are i.i.d. standard Gaussian variables, and $U$ is a uniform random variable on $[0, 1]$. The right-most term in~\eqref{eq:reg_syn_model} creates a few but large outliers. Figure~\ref{fig:data-example} in the appendix illustrates the effect of the noise models discussed earlier on data sampled from~\eqref{eq:reg_syn_model}.

\begin{figure}[t]
\begin{center}
{\includegraphics[width=.5\linewidth]{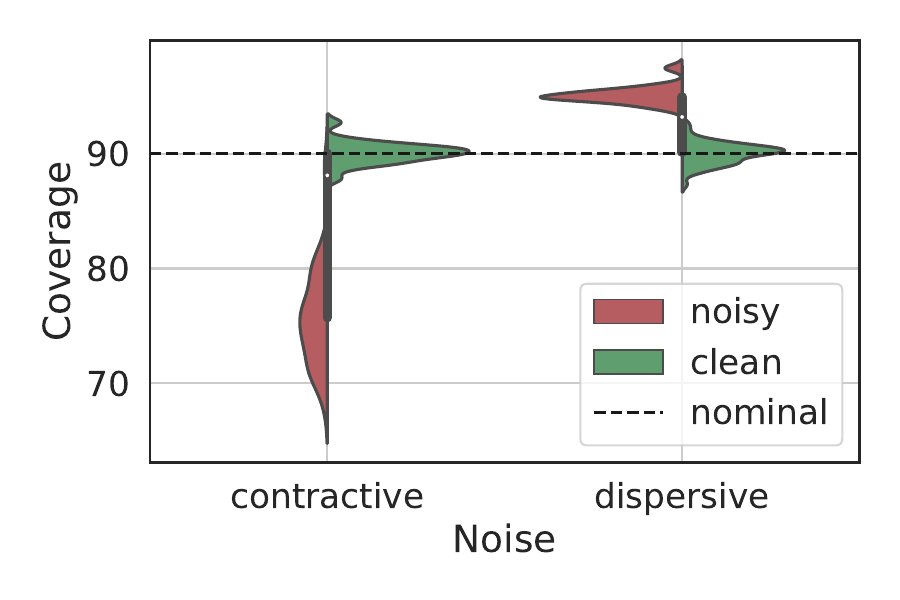}}\hfill
{\includegraphics[width=.5\linewidth]{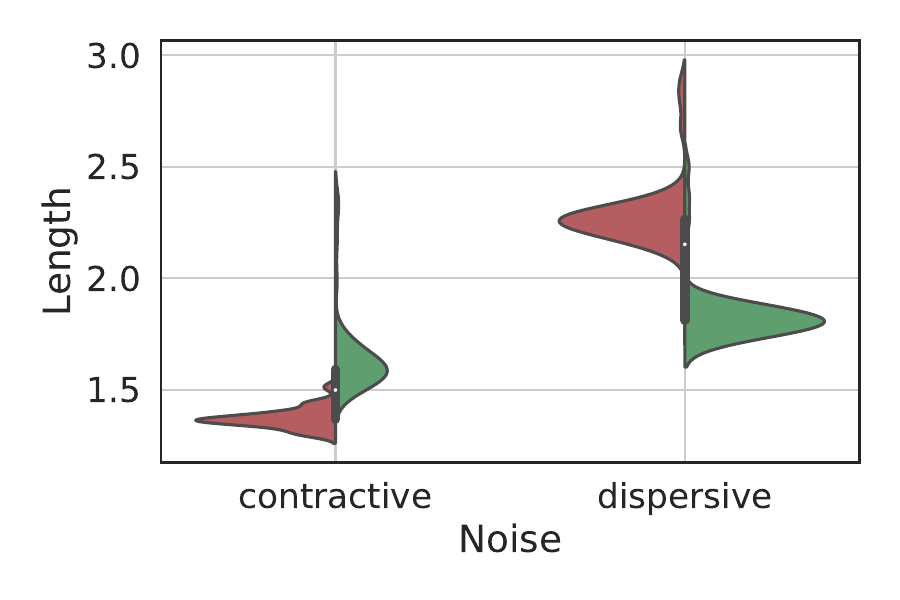}}
\caption{\textbf{Dispersive versus contractive noise regression experiment.} Performance of conformal prediction intervals with target coverage $1-\alpha=90\%$, using a noisy training set and a noisy calibration set. Left: Marginal coverage; Right: Length of predicted intervals. The results are evaluated over 50 independent experiments and the gray bar represents the interquartile range.}
\label{fig:center_noisy_train}
\end{center}
\vspace{-0.5cm}
\end{figure}

We apply conformal prediction with the CQR score \citep{romano2019conformalized} for each noise model as follows.
First, we fit a quantile random forest model on $8,000$ noisy training points; we then calibrate the model using $2,000$ fresh noisy samples; and, lastly, test the performance on additional $5,000$ clean, ground truth samples. The results are summarized in Figures~\ref{fig:sym_noisy_train} and~\ref{fig:center_noisy_train}. Observe how the prediction intervals tend to be conservative under \texttt{symmetric}, both for light- and heavy-tailed noise distributions, \texttt{asymmetric}, and \texttt{dispersive} corruption models. Intuitively, this is because these noise models increase the variability of $Y$; in Proposition~\ref{prop:regression_valid_miscoverage} we prove this formally for any \texttt{symmetric} independent noise model, whereas here we show this result holds more generally even for response-dependent noise. By contrast, the prediction intervals constructed under the \texttt{biased} and \texttt{contractive} corruption models tend to under-cover the response variable. This should not surprise us: following Figure~\ref{fig:data-example}(c), the \texttt{biased} noise shifts the data `upwards', and, consequently, the prediction intervals are undesirably pushed towards the positive quadrants. Analogously, the \texttt{contractive} corruption model pushes the data towards the mean, leading to intervals that are too narrow. Figure~\ref{fig:scores-noisy_train} in the appendix illustrates the scores achieved when using the different noise models and the 90\%'th empirical quantile of the CQR scores. This figure supports the behavior witnessed in Figures~\ref{fig:sym_noisy_train},~\ref{fig:noise_clean_train} and~\ref{fig:w2r_and_r2c}: over-coverage is achieved when $\qhatnoisy$ is larger than $\qhatclean$, and under-coverage is obtained when $\qhatnoisy$ is smaller.

In Appendix~\ref{app:synthetic-reg-supp} we study the effect of the predictive model on the coverage property, for all noise models. To this end, we repeat similar experiments to the ones presented above, however, we now fit the predictive model on clean training data; the calibration data remains noisy. We also provide an additional adversarial noise model that reduces the coverage rate, but is unlikely to appear in real-world settings. Figures~\ref{fig:noise_clean_train} and~\ref{fig:w2r_and_r2c} in the appendix depict a similar behaviour for most noise models, except the \texttt{biased} noise for which the coverage requirement is not violated. This can be explained by the improved estimation of the low and high conditional quantiles, as these are fitted on clean data and thus less biased.

\subsection{Multi-label Classification}

\begin{figure}[!ht]

  \centering
  \hspace{0.05\linewidth}
  \begin{minipage}{0.4\textwidth}
    \includegraphics[width=\linewidth]{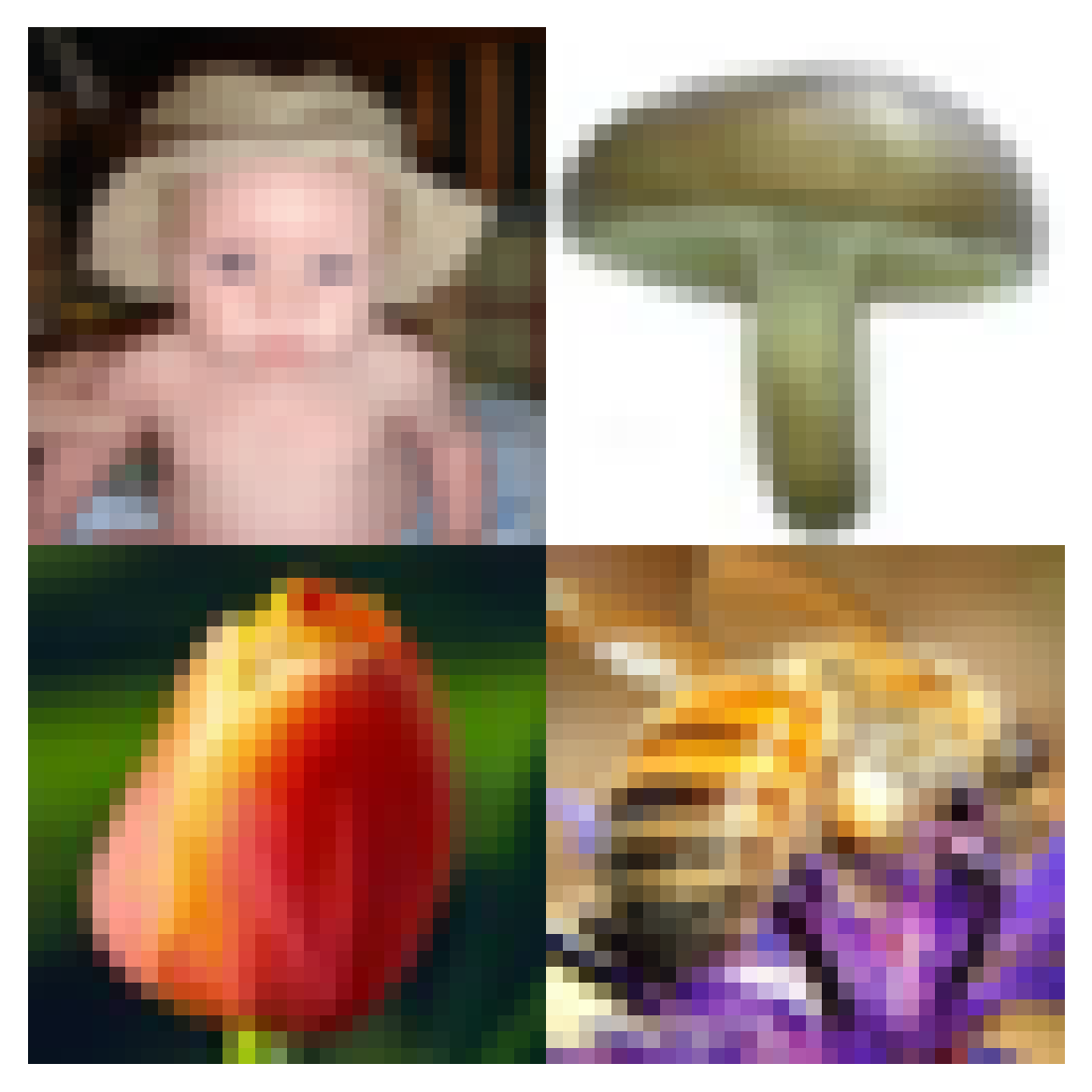}
  \end{minipage}
    \hspace{0.02\linewidth}
  \begin{minipage}{0.5\textwidth}
  True labels:\\\ baby, mushroom, tulip, bee. \\
  \\
  Noisy labels:\\\ baby, mushroom, sweet pepper, bee.
  \end{minipage}
  
\caption{Visualization of the multi-label CIFAR-100N data.}
\label{fig:cifar100N-example}
\end{figure}

In this section, we analyze conformal risk control in a multi-label classification task. For this purpose, we use the CIFAR-100N data set \citep{wei2022learning}, which contains 50K colored images. Each image belongs to one of a hundred fine classes that are grouped into twenty mutually exclusive coarse super-classes. Furthermore, every image has a noisy and a clean label, where the noise rate of the fine categories is $40\%$ and of the coarse categories is $25\%$. We turn this single-label classification task into a multi-label classification task by merging four random images into a 2 by 2 grid. Every image is used once in each position of the grid, and therefore this new data set consists of 50K images, where each is composed of four sub-images and thus associated with up to four labels. Figure~\ref{fig:cifar100N-example} displays a visualization of this new variant of the CIFAR-100N data set.

\begin{figure}[ht]
\begin{center}
{\includegraphics[height=.3\linewidth]{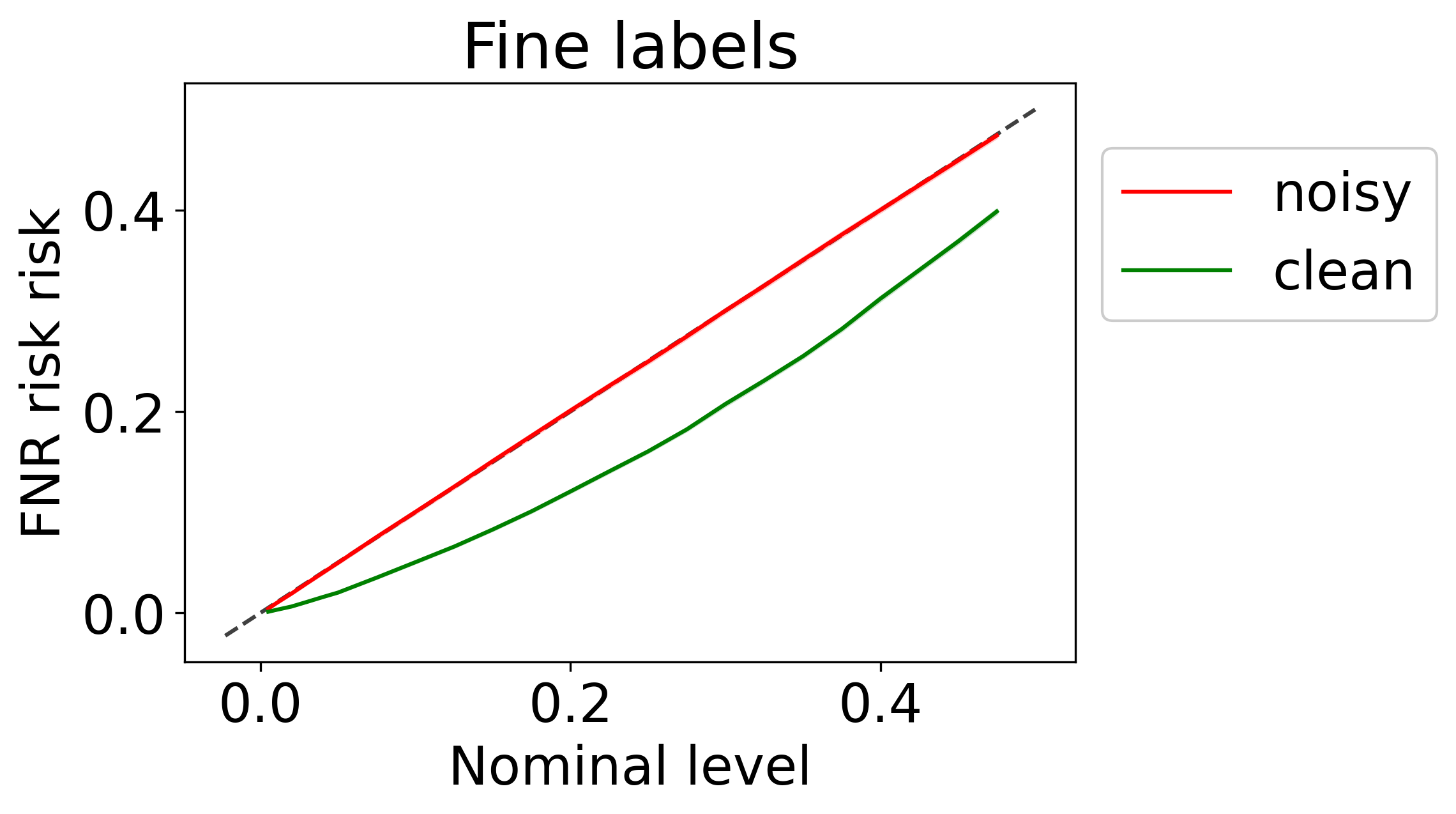}}
{\includegraphics[height=.3\linewidth]{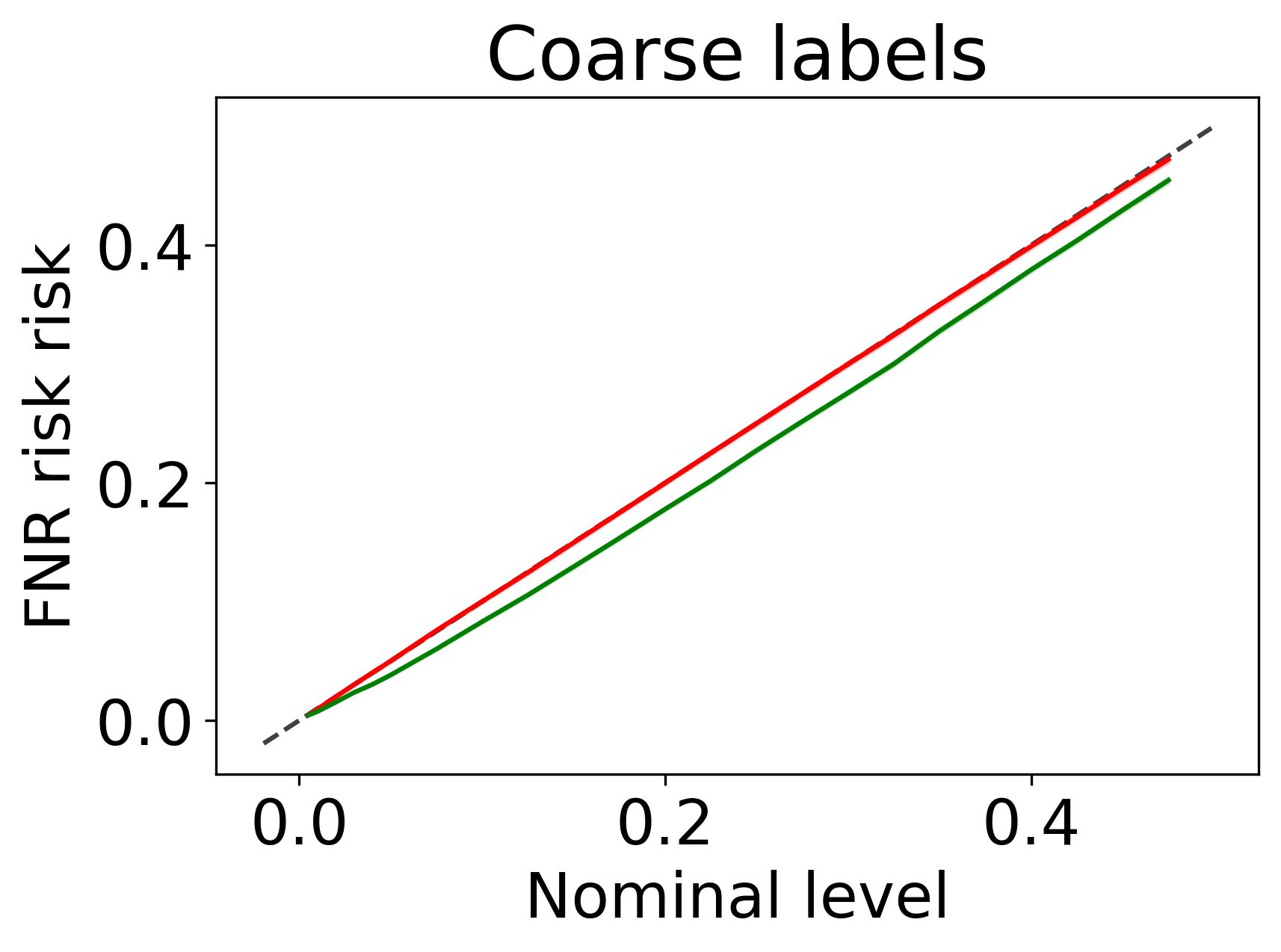}}
\caption{FNR achieved over noisy (red) and clean (green) test sets of the on CIFAR-100N data set. Left: fine labels. Right: coarse labels. The calibration scheme is applied using noisy annotations in all settings. Results are averaged over 50 trials.}
\label{fig:cifar_fnr}
\end{center}
\end{figure}

We fit a TResNet \citep{ridnik2021tresnet} model on 40k noisy samples and calibrate it using 2K noisy samples with conformal risk control, as outlined in \cite[Section 3.2]{angelopoulos2022conformal}. We control the false-negative rate (FNR) defined in~\eqref{eq:fnr} at different levels and measure the FNR obtained over clean and noisy versions of the test set, which contains 8k samples. We conducted this experiment twice: once with the fine-classed labels and once with the super-classed labels. Figure~\ref{fig:cifar_fnr} presents the results in both settings, showing that the risk obtained over the clean labels is valid for every nominal level. Importantly, this corruption setting violates the assumptions of Proposition~\ref{prop:multi_label}, as the positive label count may vary across different noise instantiations. This experiment reveals that valid risk can be achieved in the presence of label noise even when the corruption model violates the requirements of our theory.

\subsection{Segmentation}
In this section, we follow a common artificial label corruption methodology, following~\cite{zhao2021evaluating, kumar2020robust}, and analyze three corruption setups that are special cases of the vector-flip noise model from~\eqref{eq:vector_flip}: independent, dependent, and partial. In the independent setting, each pixel's label is flipped with probability $\beta$, independently of the others. In the dependent setting, however, two rectangles in the image are entirely flipped with probability $\beta$, and the other pixels are flipped independently with probability $\beta$. Finally, in the partial noise setting, only one rectangle in the image is flipped with probability $\beta$, and the other pixels are unchanged. 

\begin{figure}[ht]
\begin{center}
{\includegraphics[height=.27\linewidth]{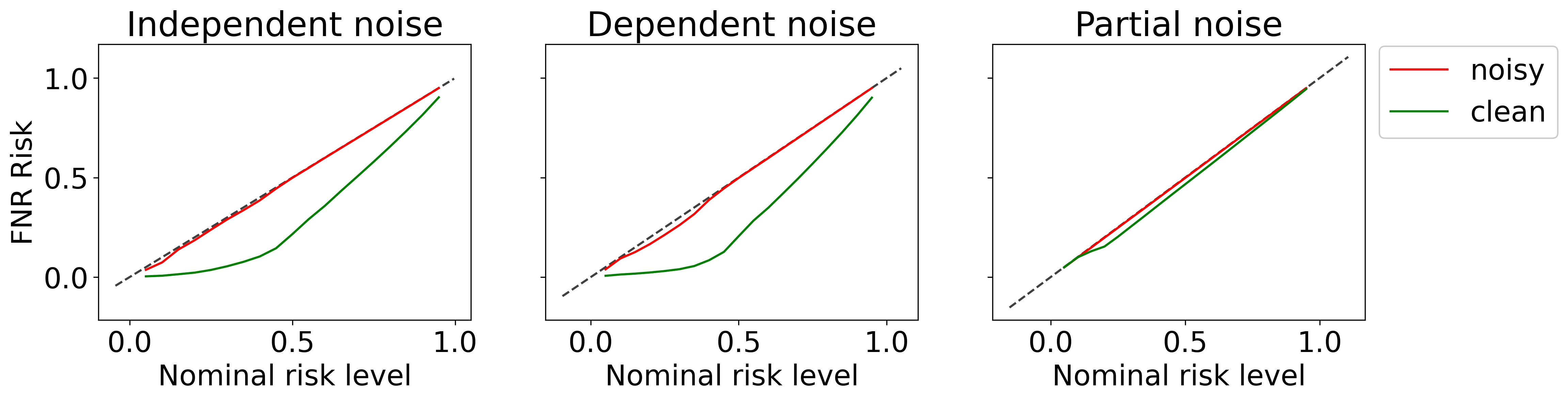}}
\caption{FNR on a polyp segmentation data set, achieved over noisy (red) and clean (green) test sets. Left: independent noise. Middle: dependent noise. Right: partial noise. The predictive model is calibrated using noisy annotations in all settings, where the noise level is set to $\beta=0.1$. Results are averaged over 1000 trials.}
\label{fig:segmentation_exp}
\end{center}
\end{figure}

We experiment on a polyp segmentation task, pooling data from several polyp data sets: Kvasir, CVC-ColonDB, CVC-ClinicDB, and ETIS-Larib. We consider the annotations given in the data as ground-truth labels and artificially corrupt them according to the corruption setups described above, to generate noisy labels. We use PraNet \citep{fan2020pranet} as a base model and fit it over 1450 noisy samples. Then, we calibrate it using 500 noisy samples with conformal risk control, as outlined in \cite[Section 3.2]{angelopoulos2022conformal} to control the false-negative rate (FNR) from~\eqref{eq:fnr} at different levels.
Finally, we evaluate the FNR over clean and noisy versions of the test set, which contains 298 samples, and report the results in Figure~\ref{fig:segmentation_exp}. This figure indicates that conformal risk control is robust to label noise, as the constructed prediction sets achieve conservative risk in all experimented noise settings.
This is not a surprise, as it is guaranteed by  Propositions~\ref{prop:segmentation} and~\ref{prop:fnr_independent}.

\subsection{Regression Risk Bounds}\label{sec:exp_reg_bounds}
\begin{figure}[ht]
\begin{center}
{\includegraphics[height=.3\linewidth]{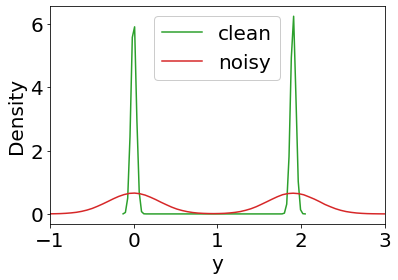}}
\caption{Visualization of the marginal density function of the adversarial synthetic data.}
\label{fig:synthetic_adversarial_reg}
\end{center}
\end{figure}

We now turn to demonstrate the coverage rate bounds derived in Section~\ref{sec:risk_regression} on real and synthetic regression data sets.
We examine two real benchmarks: \texttt{\cite{meps19_data}} and \texttt{\cite{bio_data}} used in~\citep{romano2019conformalized}, and one synthetic data set that was generated from a bimodal density function with a sharp slope, as visualized in Figure~\ref{fig:synthetic_adversarial_reg}. The simulated data was deliberately designed to invalidate the assumptions of our label-noise robustness requirements in Proposition~\ref{prop:regression_valid_miscoverage}. Consequentially, prediction intervals that do not cover the two peaks of the density function might undercover the true outcome, even if the noise is dispersive. 
Therefore, this gap calls for our distribution-free risk bounds from Section~\ref{sec:risk_regression}, which are applicable in this setup, in contrast to Proposition~\ref{prop:regression_valid_miscoverage}. The former approach can be used to assess the worst risk level that may be obtained in practice.
We consider the labels given in the real and synthetic data sets as ground truth and artificially corrupt them according to the additive noise model~\eqref{eq:reg_g_sym}. The added noise is independently sampled from a normal distribution with mean zero and variance $0.1\text{Var}(Y)$. 

For each data set and nominal risk level $\alpha$, fit a quantile regression model on 12K samples and learn the $\alpha/2, 1-\alpha/2$ conditional quantiles of the noisy labels. Then, we calibrate its outputs using another 12K samples of the data with conformal risk control, by~\cite{angelopoulos2022conformal},
to control the smooth miscoverage~\eqref{eq:smooth_miscoverage} at level $\alpha$. Finally, we evaluate the performance on the test set which consists of 6K samples. We also compute the smooth miscoverage risk bounds according to Corollary~\ref{cor:smooth_miscoverage} with a noise variance set to $0.1$.

\begin{figure}[ht]
\begin{center}
{\includegraphics[width=.32\linewidth]{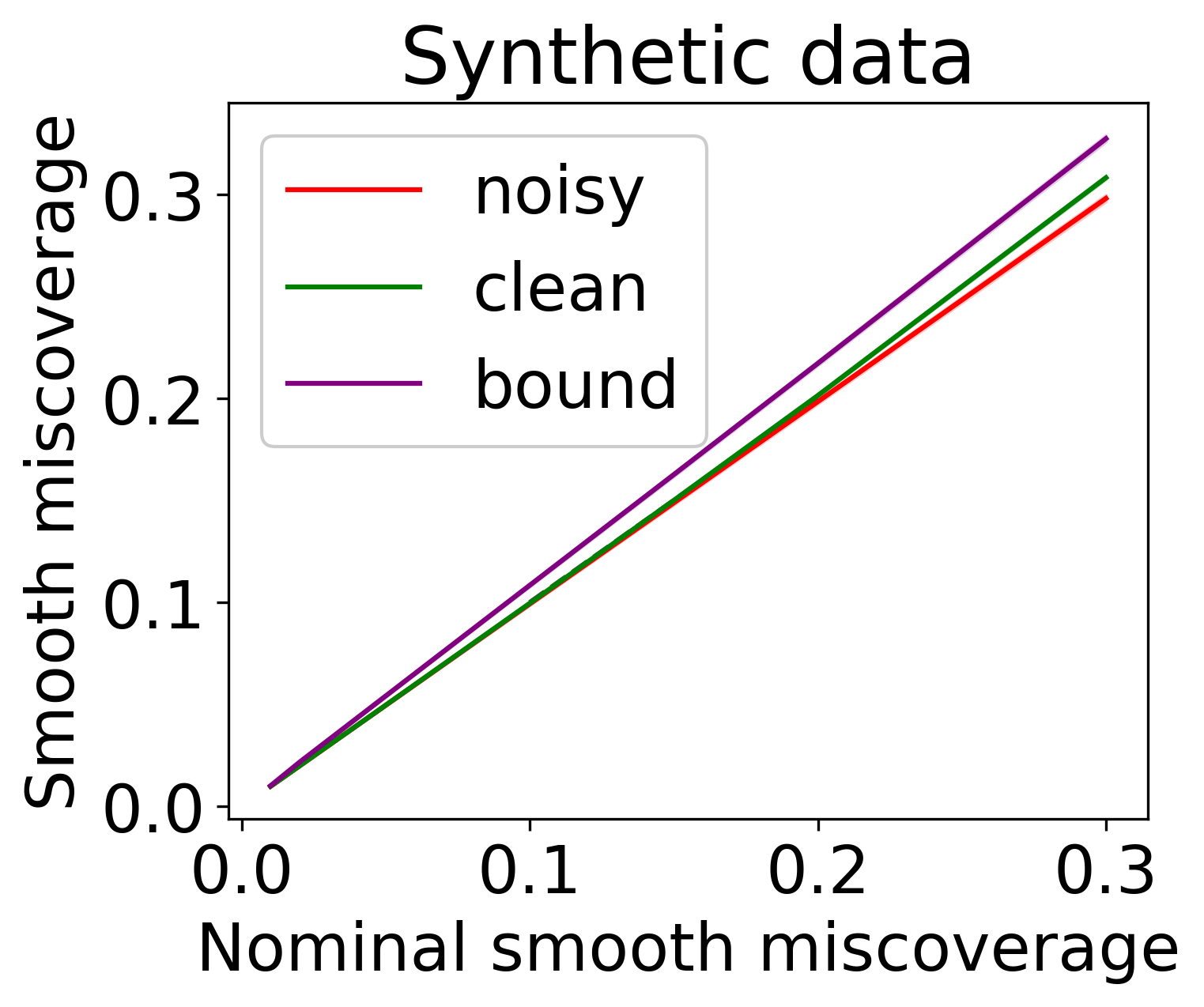}}
{\includegraphics[width=.32\linewidth]{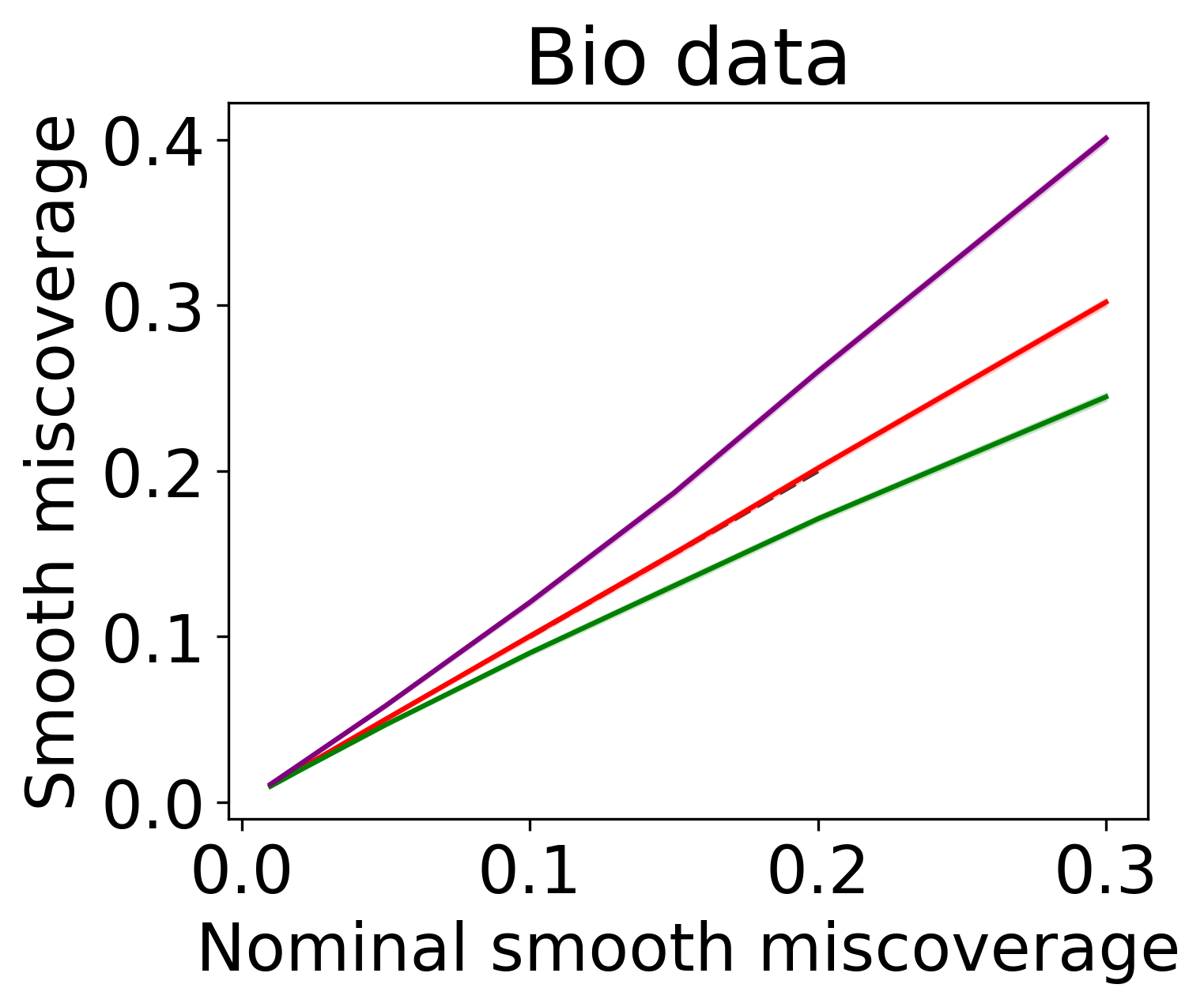}}
{\includegraphics[width=.32\linewidth]{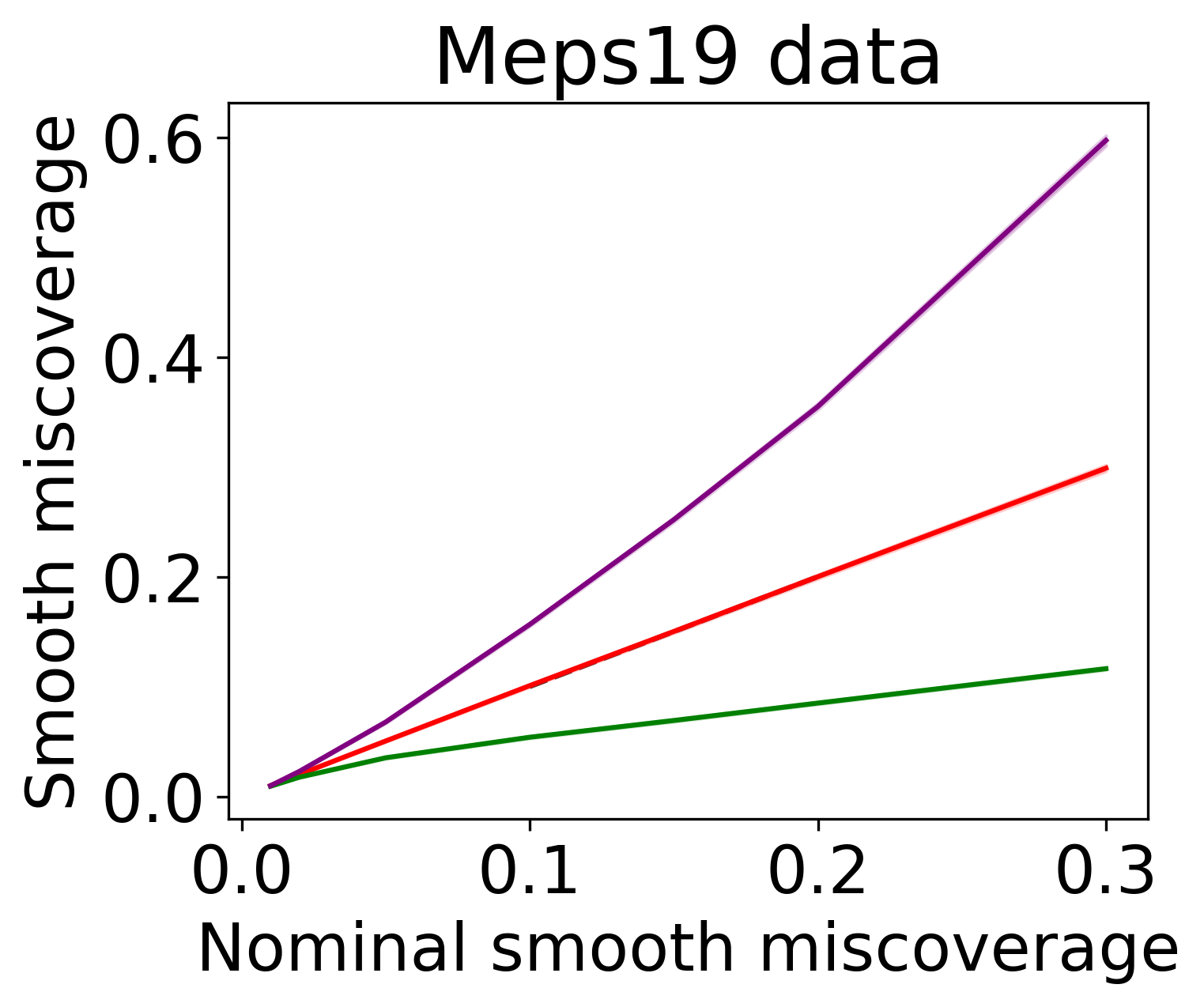}}
\caption{Smooth miscoverage rate achieved over noisy (red) and clean (green) test sets. The calibration scheme is applied using noisy annotations to control the smooth miscoverage level. Results are averaged over 10 random splits of the calibration and test sets.}
\label{fig:smooth_miscoverage_bounds}
\end{center}
\end{figure}

Figure~\ref{fig:smooth_miscoverage_bounds} presents the risk bound along with the smooth miscoverage obtained over the clean and noisy versions of the test set. This figure indicates that conformal risk control generates invalid uncertainty sets when applied on the simulated noisy data, as we anticipated. Additionally, this figure shows that the proposed risk bounds are valid and tight, meaning that these are informative and effective. Moreover, this figure highlights the main advantage of the proposed risk bounds: their validity is universal across all distributions of the response variable and the noise component. Lastly, we note that in Appendix~\ref{sec:miscoverage_bounds_exp} we repeat this experiment with the miscoverage loss and display the miscoverage bound derived from Corollary~\ref{prop:cov_bound}.

\subsection{Online Learning}\label{sec:online_exp}

This section studies the effect of label noise on uncertainty quantification methods in an online learning setting, as formulated in Section~\ref{sec:online_thoery}. 
We experiment on a depth estimation task~\citep{kitti}, where the objective is to predict a depth map given a colored image. In other words, $X\in \mathbb{R}^{W \times H \times 3}$ is an input RGB image of size $W\times H$ and $Y\in \mathbb{R}^{W \times H}$ is its corresponding depth map.
We consider the original depth values given in this data as ground truth and artificially corrupt them according to the additive noise model defined in~\eqref{eq:reg_g_sym} to produce noisy labels. Specifically, we add to each depth pixel an independent random noise drawn from a normal distribution with zero mean and 0.7 variance.
Here, the depth uncertainty of the $i,j$ pixel is represented by a prediction interval $C^{i,j}(X)\subseteq \mathbb{R}$. Ideally, the estimated intervals should contain the correct depth values at a pre-specified level $1-\alpha$. In this high-dimensional setting, this requirement is formalized as controlling the \emph{image miscoverage} loss, defined as:
\begin{equation}\label{eq:image_miscoverage_loss}
L^{\text{im}}(Y,C(X)) = 
\frac{1}{WH}\sum_{i=1}^W {\sum_{j=1}^H{1\{Y^{i,j} \notin C^{i,j}(X) \}}}. 
\end{equation}
In words, the image miscoverage loss measures the proportion of depth values that were not covered in a given image.

For this purpose, we employ the calibration scheme \texttt{Rolling RC}~\citep{feldman2022achieving}, which constructs uncertainty sets in an online setting with a valid risk guarantee in the sense of~\eqref{eq:online_risk_requirement}. We follow the experimental protocol outlined in \cite[Section 4.2]{feldman2022achieving} and apply \texttt{Rolling RC} with an exponential stretching to control the image miscoverage loss at different levels on the observed, noisy, labels. We use LeReS~\citep{leres} as a base model, which was pre-trained on a clean training set that corresponds to timestamps 1,...,6000. 
We continue training it and updating the calibration scheme in an online fashion on the following 2000 timestamps. We consider these samples, indexed by 6001 to 8000, as a validation set and use it to choose the calibration's hyperparameters, as explained in \cite[Section 4.2]{feldman2022achieving}. 
Finally, we continue the online procedure on the test samples whose indexes correspond to 8001 to 10000, and measure the performance on the clean and noisy versions of this test set. 

\begin{figure}[ht]
\centering
            {\includegraphics[width=.4\linewidth]{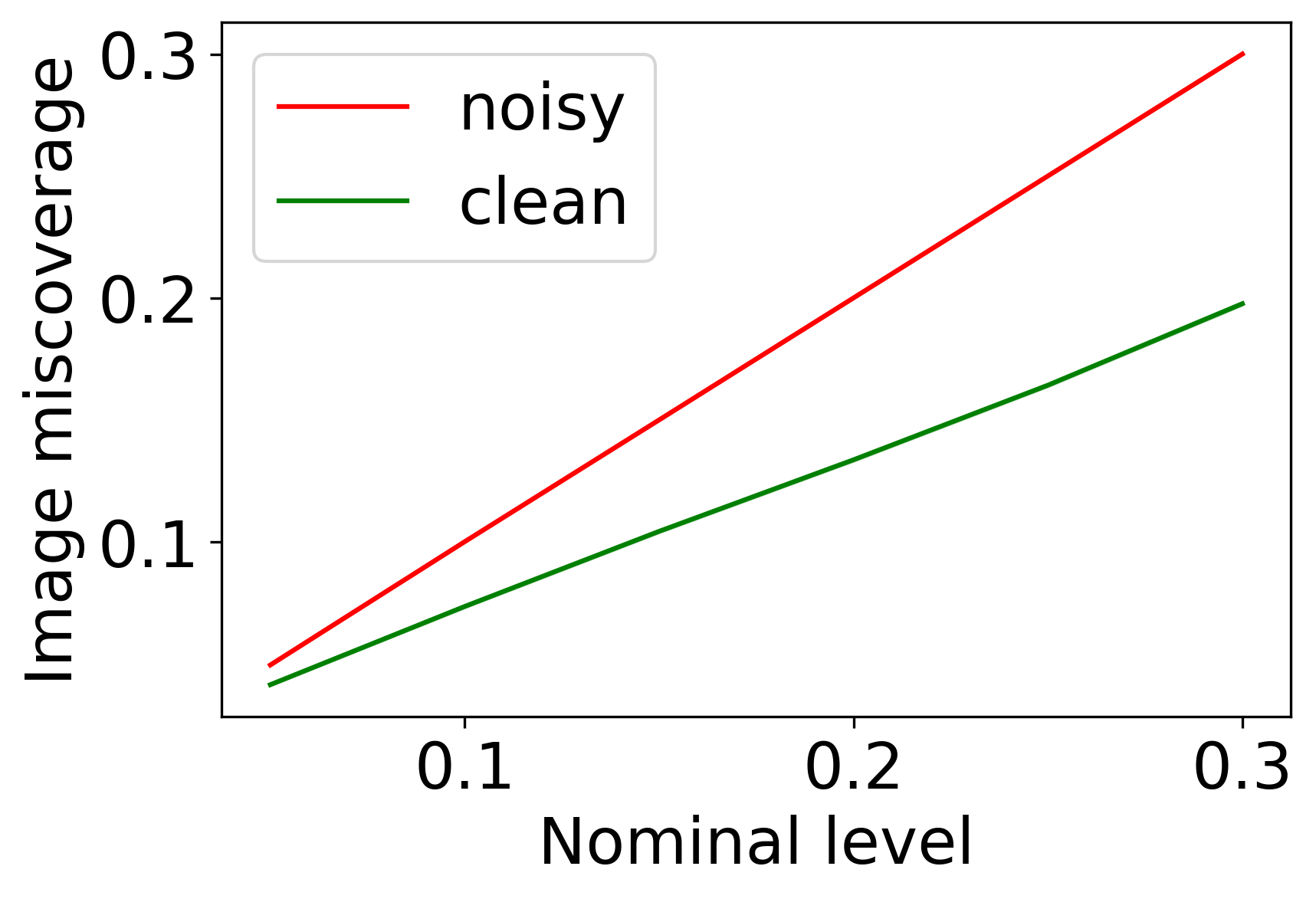}}
            \caption{Image miscoverage achieved by \texttt{Rolling RC}. Results are averaged over 10 random trials.}
            \label{fig:online_image_miscoverage}
\end{figure}

Figure~\ref{fig:online_image_miscoverage} displays the risk achieved by this technique over the clean and corrupted labels. This figure indicates that \texttt{Rolling RC} attain valid image miscoverage over the unknown noiseless labels. This is not a surprise, as it is supported by Proposition~\ref{prop:image_miscoverage} which guarantees conservative image miscoverage under label noise in an offline setting, and Proposition~\ref{prop:online} which states that the former result applies to an online learning setting as well. 

\newpage

\section{Discussion}

\subsection{Related Work}\label{sec:related_work}
Label noise independently of conformal prediction has been well-studied, especially for training more robust predictive models; see, for example~\citep{angluin1988learning,frenay2013classification, tanno2019learning, algan2020label, kumar2020robust}.
Recently, there has been a body of work studying the statistical properties of conformal prediction~\citep{lei2018distribution,barber2020distributionfree} and its performance under deviations from exchangeability~\citep{tibshirani2019conformal, podkopaev2021distribution}. This line of work is relevant to us since the label noise setting violates the exchangeability assumption between the training and testing data, where the latter is clean while the former is noisy. However, these works cannot be applied in our setting since they assume covariate or label shift only. The work by~\citet{barber2022conformal} refers to any distribution shift, and we build upon some of its results in our general disclaimer in Section~\ref{sec:disclaimer}. 
Another relevant work is \citep{farinhas2024nonexchangeable}, which studies the effect of a general distribution shift on the obtained risk of risk-controlling techniques, thus extending the work of by~\citet{barber2022conformal} to more general loss functions than the miscoverage loss. Additionally, \cite{angelopoulos2022conformal} analyzes their proposed risk-controlling method in a covariate shift and a general distribution shift settings.
Close works to ours include~\citep{stutz2023conformal}, which analyses the performance of conformal prediction under ambiguous ground truth, and~\citet{cauchois2022weak}, that studies conformal prediction with weak supervision, which could be interpreted as a type of noisy label.

Lastly, a follow-up work to ours has recently been published~\citep{sesia2023adaptive}. Similarly to our work, it begins by studying the effect of label noise on the coverage achieved by standard conformal prediction. Then, an explicit factor that depicts the inflation or deflation of the coverage is estimated to adjust the desired coverage rate. The theoretical analysis requires no assumptions
on the contamination process, but in order to achieve an applicable method that can automatically adapt to label noise, some mild assumptions on the relation between the clean and observable labels are used.
Indeed, the presented experiments demonstrate less conservative coverage compared to the standard method. However, there are two key distinctions between this work and ours. It focuses on controlling the coverage rate in classification tasks, whilst our analysis extends to regression tasks, general risk control, and online settings. Additionally, it aims to modify the calibration algorithm to account for label noise, whereas our goal is to test the limits of conformal prediction in the label noise setting and reveal the conditions on the scores, noise models, and predictive models, under which the standard algorithm remains valid despite the presence of noisy labels.

\subsection{Future Research Direction}
Our work raises many new questions. First, one can try and define a score function that is more robust to label noise, continuing the line of~\citet{gendler2021adversarially, frenay2013classification, cheng2022many}.
Second, an important remaining question is how to achieve exact risk control on the clean labels using minimal information about the noise model.
Lastly, it would be interesting to analyze the robustness of alternative conformal methods such as cross-conformal and jackknife+~\citep{vovk2015cross, barber2019jackknife} that do not require data-splitting.

\acks{
Y.R., A.G., B.E., and S.F. were supported by the ISRAEL SCIENCE FOUNDATION (grant No. 729/21). Y.R. thanks the Career Advancement Fellowship, Technion, for providing research support. 
A.N.A. was supported by the National Science Foundation Graduate Research Fellowship Program under Grant No. DGE 1752814. 
S.F. thanks Aviv Adar, Idan Aviv, Ofer Bear, Tsvi Bekker, Yoav Bourla, Yotam Gilad, Dor Sirton, and Lia Tabib for annotating the MS COCO data set.} 






\clearpage
\onecolumn
\appendix

\section{Mathematical Proofs} \label{app:thrm-proof}
\subsection{General Analysis} \label{app-general}

We begin by proving Theorem~\ref{thm:general}.

\begin{proof}[Proof of Theorem~\ref{thm:general}]
Our assumption states that
\begin{equation}
    \P(\ts_{\rm test} \leq t) \leq \P(s_{\rm test} \leq t).
\end{equation}
Note that the probability is only taken over $\ts_{\rm test}$. 
Since $\hat{q}_{\rm noisy}$ is constant (measurable) with respect to this probability, we have that, for any $\alpha \in (0,1)$,
\begin{equation}
    \P(s_{\rm test} \leq \hat{q}_{\rm noisy}) \geq \P(\ts_{\rm test} \leq \hat{q}_{\rm noisy}) \geq 1-\alpha.
\end{equation}
This implies that $Y_{\rm test} \in \Chatnoisy(X_{\rm test})$ with probability at least $1-\alpha$, completing the proof of the lower bound.

Regarding the upper bound, by the same argument,
\begin{equation}
    \P(s_{\rm test} \leq \hat{q}_{\rm noisy}) \leq \P(\ts_{\rm test} \leq \hat{q}_{\rm noisy}) + u \leq 1-\alpha+\frac{1}{n+1} + u.
\end{equation}

\end{proof}

\subsection{Regression}
\subsubsection{General Regression Result}\label{sec:general_regression_result}

Here we provide an extension to Proposition~\ref{prop:regression_valid_miscoverage} 
 and prove it.
\begin{theorem}\label{thm:general_regression}
Suppose an additive noise model $g^{\rm add}$ with a noise that has mean 0. Denote the prediction interval as $C(x) = [a_x, b_x]$. If for all $x\in\mathcal{X}$ density of $Y\mid X=x$ is peaked inside the interval:
\begin{enumerate}
    \item $\forall \varepsilon \geq 0: f_{Y\mid X=x}(b_x + \varepsilon) \leq  f_{Y\mid X=x}(b_x - \varepsilon).$
    \item $\forall \varepsilon \geq 0: f_{Y\mid X=x}(a_x - \varepsilon) \leq  f_{Y\mid X=x}(a_x + \varepsilon).$
\end{enumerate}
then, we obtain valid conditional coverage:
\begin{equation}
    \mathbb{P}(Y \in C(x) \mid X=x) \geq \mathbb{P}(\Tilde{Y} \in C(x) \mid X=x).
\end{equation}
\end{theorem}

\begin{proof}
For ease of notation, we omit the conditioning on $X=x$. In other words, we treat $Y$ as $Y\mid X=x$ for some $x \in \mathcal{X}$.
We begin by showing $\mathbb{P}(\tilde{Y} \leq b) \geq \mathbb{P}(Y \leq b)$.

\begin{equation}
\begin{split}
&\mathbb{P}(\tilde{Y} \leq b) \\
&= \mathbb{P}(Y +\varepsilon \leq b) \\
&= \mathbb{P}(Y + \varepsilon \leq b \mid \varepsilon \geq 0) \mathbb{P}(\varepsilon \geq 0) + \mathbb{P}(Y + \varepsilon \leq b \mid \varepsilon \leq 0) \mathbb{P}(\varepsilon \leq 0) \\
&= \frac{1}{2}\mathbb{P}(Y + \varepsilon \leq b \mid \varepsilon \geq 0)  + \frac{1}{2}\mathbb{P}(Y + \varepsilon \leq b \mid \varepsilon \leq 0) \\
&= \frac{1}{2}\mathbb{P}(Y \leq b -\varepsilon \mid \varepsilon \geq 0) + \frac{1}{2}\mathbb{P}(Y \leq b +\varepsilon \mid \varepsilon \geq 0) \\
&= \frac{1}{2}\mathbb{E}_{\varepsilon \geq 0} \left[\mathbb{P}(Y \leq b -\varepsilon) + \mathbb{P}(Y \leq b +\varepsilon) \right] \\
&= \mathbb{P}(Y \leq b) + \frac{1}{2}\mathbb{E}_{\varepsilon \geq 0}\left[\mathbb{P}(Y \leq b -\varepsilon)-\mathbb{P}(Y \leq b)  + \mathbb{P}(Y \leq b +\varepsilon) - \mathbb{P}(Y \leq b) \right] \\
&= \mathbb{P}(Y \leq b) + \frac{1}{2}\mathbb{E}_{\varepsilon \geq 0}\left[\mathbb{P}(Y \leq b +\varepsilon) - \mathbb{P}(Y \leq b) - (\mathbb{P}(Y \leq b) - \mathbb{P}(Y \leq b -\varepsilon)) \right] \\
&= \mathbb{P}(Y \leq b) + \frac{1}{2}\mathbb{E}_{\varepsilon \geq 0}\left[\mathbb{P}(b\leq Y \leq b +\varepsilon) - \mathbb{P}(b -\varepsilon \leq Y \leq b)  \right] \\
& \leq \mathbb{P}(Y \leq b) + \frac{1}{2}\mathbb{E}_{\varepsilon \geq 0}\left[ 0 \right] \\
&= \mathbb{P}(Y \leq b).
\end{split}
\end{equation}
The last inequality follows from the assumption that $\forall \varepsilon \geq 0: f_{Y}(b + \varepsilon) \leq  f_{Y}(b - \varepsilon)$.
The proof for $\mathbb{P}(\tilde{Y} \geq a) \leq \mathbb{P}(Y \geq a)$ is similar and hence omitted. We get that:
\begin{equation}
\begin{split}
&\mathbb{P}(\tilde{Y} \in C(x)) \\
& = \mathbb{P}(a \leq \tilde{Y} \leq b ) \\
& = \mathbb{P}(\tilde{Y} \leq b) - \mathbb{P}( \tilde{Y} \geq a ) \\
& \leq \mathbb{P}(Y \leq b) - \mathbb{P}(Y \geq a)  \text{ (follows from the above)} \\
& = \mathbb{P}(a \leq Y \leq b) \\
& = \mathbb{P}(Y \in C(x)).
\end{split}
\end{equation}
\end{proof}

We now turn to prove Proposition~\ref{prop:regression_valid_miscoverage} using Theorem~\ref{thm:general_regression}
\begin{proof}
Since the density of $Y\mid X=x$ is symmetric and unimodal for all $x\in\mathcal{X}$ it is peaked inside any prediction interval that contains its median. Therefore, the prediction interval achieves valid conditional coverage:
\begin{equation}
    \forall x\in\mathcal{X}: \mathbb{P}(\tilde{Y} \in C(x) \mid X=x) \geq  \mathbb{P}(Y \in C(x) \mid X=x).
\end{equation}
By taking the expectation over $X\sim P_X$ we obtain valid marginal coverage.
\end{proof}

\subsection{Classification}
\subsubsection{General Classification Result} \label{app:gen-class}

We commence by providing an extension to Proposition~\ref{prop:general_classification} and proving it.
\begin{theorem} \label{thm:general_class}
Suppose that $\Chatnoisy(x)\subseteq \{1,...,K\}$ is a prediction set. Denote $\beta := \mathbb{P}(\tilde{Y}\in \Chatnoisy(x) \mid X=x)$. First, the coverage rate achieved over the clean labels is upper bounded by:
\begin{equation}
\mathbb{P}(Y\in \Chatnoisy(x) \mid X=x) \leq  \beta + \frac{1}{2}\sum_{i=1}^{K} {|\mathbb{P}(\tilde{Y} = i \mid X=x) - \mathbb{P}({Y} = i \mid X=x) |},
\end{equation}
Further suppose that for all $x\in\mathcal{X}$: 
\begin{equation}
    \sum_{i\in \Chatnoisy(x)} {\left( \mathbb{P}(Y = i \mid X=x) - \mathbb{P}(\tilde{Y} = i \mid X=x)\right)} \geq 0.
\end{equation}
Then, the coverage rate is lower bounded by:
\begin{equation}
\mathbb{P}(\tilde{Y}\in \Chatnoisy(x) \mid X=x) \leq \mathbb{P}(Y\in \Chatnoisy(x) \mid X=x).
\end{equation}
\end{theorem}

\begin{proof}
First, for ease of notation, we omit the conditioning on $x$ and consider a prediction $C \subseteq \{1,...,K\}$. 
Notice that if $C=\emptyset$ the proposition is trivially satisfied.

We begin by proving the following lower bound $\mathbb{P}({Y}\in C) \geq \beta := \mathbb{P}(\tilde{Y}\in C)$.
Denote: $\delta_i := \mathbb{P}(Y = i) - \mathbb{P}(\tilde{Y} = i)$. Since $\sum_{i\in C} { \delta_i} \geq 0$ we get that:
\begin{equation}
\begin{split}
\mathbb{P}(Y \in C) = \sum_{i\in C} { \mathbb{P}(Y = i) } = \sum_{i\in C} { \mathbb{P}(\tilde{Y} = i)} + \sum_{i\in C} { \delta_i} \geq \sum_{i\in C} { \mathbb{P}(\tilde{Y} = i)} + 0 = \mathbb{P}(\tilde{Y} \in C) = \beta.
\end{split}
\end{equation}
We now turn to prove the upper bound.
\begin{equation}
\begin{split}
\sum_{i\in C} { \delta_i} &= \frac{1}{2}\left(\sum_{i\in C} {\delta_i} + \sum_{i\in C} {\delta_i}\right) \\
&= \frac{1}{2}\left(\sum_{i\in C} {\delta_i} + \sum_{i\in \{1,...,K\}-C} {-\delta_i}\right) \\
&\leq \frac{1}{2}\sum_{i=1}^{K} {|\delta_i|} \\
&= \frac{1}{2}\sum_{i=1}^{K} {|\mathbb{P}(\tilde{Y} = i) - \mathbb{P}({Y} = i) |}.
\end{split}
\end{equation}
This gives us:
\begin{equation}
\mathbb{P}(Y \in C) = \mathbb{P}(\tilde{Y} \in C) + \sum_{i\in C} { \delta_i} \leq \mathbb{P}(\tilde{Y} \in C) + \frac{1}{2}\sum_{i=1}^{K} {|\mathbb{P}(\tilde{Y} = i) - \mathbb{P}({Y} = i) |}.
\end{equation}
And this concludes the proof.

\end{proof}

We now turn to prove Proposition~\ref{prop:general_classification} using Theorem~\ref{thm:general_class}.

\begin{proof}
We suppose without loss of generality that the labels are ranked from the most likely to the least likely, i.e., $\mathbb{P}(Y=i) \geq \mathbb{P}(Y=i+1)$. Since the prediction set contains the most likely labels, there exists some $1 \leq m \leq K$ such that the prediction set is $C=\{1,2,...,m\}$. 
We only need to show $\sum_{i=1}^{m} {\left( \mathbb{P}(Y = i) - \mathbb{P}(\tilde{Y} = i )\right)} \geq 0$ for $m\in\{1,...,K\}$ and the result will follow directly from Theorem~\ref{thm:general_class}. We follow the notations in the proof of Theorem~\ref{thm:general_class}.
First, we observe that under the assumed noise model:
\begin{equation}
\mathbb{P}(Y = i) \geq \frac{1}{K} \iff \mathbb{P}(Y = i) \geq \mathbb{P}(\tilde{Y} = i) \iff \delta_i \geq 0.
\end{equation}
Denote by $1 \leq g\in \mathbb{N}$ the largest index for which $\mathbb{P}(Y = g) \geq \frac{1}{K}$. It follows that:
\begin{equation}
\forall i \in \{1,...,K\}: i \leq g \iff \delta_i \geq 0.
\end{equation}
If $m\leq g$ then all summands in $\sum_{i=1}^{m} { \delta_i}$ are positive since $\delta_i \geq 0$ for every $i \leq m \leq g$. Thus, $\sum_{i=1}^{m} { \delta_i} \geq 0$ as required. Otherwise, 
\begin{equation}
\sum_{i=1}^{m} { \delta_i} = \sum_{i=1}^{g} { \delta_i} + \sum_{i=g+1}^{m} { \delta_i} \underset{i > g \iff \delta_i \leq 0}{\geq} \sum_{i=1}^{g} { \delta_i} + \sum_{i=g+1}^{K} { \delta_i} = \sum_{i=1}^{K} { \delta_i}= 0.
\end{equation}
Therefore, from Theorem~\ref{thm:general_class} we get:
\begin{equation}
\beta \leq \mathbb{P}(Y\in \Chatnoisy(x) \mid X=x) \leq  \beta + \frac{1}{2}\sum_{i=1}^{K} {|\mathbb{P}(\tilde{Y} = i \mid X=x) - \mathbb{P}({Y} = i \mid X=x) |}.
\end{equation}
By taking the expectation over $X\sim P_X$ we obtain the desired marginal coverage bounds. The only non-trivial transition is marginalizing the TV-distance, which follows from the integral absolute value inequality:
\begin{equation}
\begin{split}
&\int_{x\in\mathcal{X}}{|\mathbb{P}(\tilde{Y} = i \mid X=x) - \mathbb{P}({Y} = i \mid X=x) |} \geq \\ 
&{\left|\int_{x\in\mathcal{X}}{\left(\mathbb{P}(\tilde{Y} = i \mid X=x) - \mathbb{P}({Y} = i \mid X=x) \right)}\right|} =\\
&{\left|{\mathbb{P}(\tilde{Y}=i) - \mathbb{P}({Y} = i)}\right|} \\
\end{split}
\end{equation}
\end{proof}
Finally, we turn to prove Corollary~\ref{cor:random_flip_upper_bound}.
\begin{proof}
We follow the notations and assumptions in Theorem~\ref{thm:general_class}. We only need to show that $\frac{1}{2}\sum_{i=1}^{K} {|\delta_i|} \leq \varepsilon\frac{K-1}{K}$ and the result will follow directly from Theorem~\ref{thm:general_class}.

In the random flip setting, $|\delta_i| = \left|\mathbb{P}({Y} = i)\varepsilon -  \frac{\varepsilon}{K}\right| = \varepsilon\left|\mathbb{P}({Y} = i) -  \frac{1}{K}\right|$. Therefore:
\begin{equation}
\begin{split}
\frac{1}{2}\sum_{i=1}^{K} {|\delta_i|} &= \frac{1}{2}\sum_{i=1}^{K} { \varepsilon\left|\mathbb{P}({Y} = i) -  \frac{1}{K}\right| }\\
&= \frac{1}{2}\varepsilon\left(\sum_{i=1}^{g} { \left|\mathbb{P}({Y} = i) -  \frac{1}{K}\right| } + \sum_{i=g+1}^{K} { \left|\mathbb{P}({Y} = i) -  \frac{1}{K}\right| } \right) \\
& = \frac{1}{2}\varepsilon\left(\sum_{i=1}^{g} { \left(\mathbb{P}({Y} = i) -  \frac{1}{K}\right) } + \sum_{i=g+1}^{K} { \left(\frac{1}{K} - \mathbb{P}({Y} = i)\right) } \right) \\
& = \frac{1}{2}\varepsilon\left(\sum_{i=1}^{g} { \mathbb{P}({Y} = i) } - \sum_{i=1}^{g} { \frac{1}{K} } + \sum_{i=g+1}^{K} { \frac{1}{K} } - \sum_{i=g+1}^{K} { \mathbb{P}({Y} = i) } \right) \\
& \leq \frac{1}{2}\varepsilon\left(1 - \frac{g}{K} + \frac{K-g}{K} - 0 \right) \\
& \leq \frac{1}{2}\varepsilon\left(\frac{K-g}{K} + \frac{K-g}{K} \right) \\
& \leq \varepsilon\frac{K-g}{K}\\
& \leq \varepsilon\frac{K-1}{K}.
\end{split}
\end{equation}
Thus, we get:
\begin{equation}
\mathbb{P}(\tilde{Y} \in C) \leq \mathbb{P}(Y \in C) \leq \mathbb{P}(\tilde{Y} \in C) + \frac{1}{2}\sum_{i=1}^{K} {|\mathbb{P}(\tilde{Y} = i) - \mathbb{P}({Y} = i) |} \leq \mathbb{P}(\tilde{Y} \in C) + \varepsilon\frac{K-1}{K},
\end{equation}
which concludes the proof.
\end{proof}

\subsubsection{Confusion Matrix}
\label{app:thrm_confusion}
The \texttt{confusion matrix} noise model is more realistic than the \texttt{random flip}. 
However, there exists a score function that causes conformal prediction to fail for any non-identity confusion matrix.  We define the corruption model as follows: consider a matrix $T$ in which $(T)_{i,j}=\P(\tilY = j \mid Y = i)$.
\begin{equation}
	g^{\rm confusion}(y) = \begin{cases} 
						1 & \text{w.p. } T_{1,y} \\
						\ldots & \\
						K & \text{w.p. } T_{K,y}.
				\end{cases} 
\end{equation}

\begin{prop}
     Let $\Chatnoisy$ be constructed as in Recipe~\ref{recipe} with any score function $s$ and the corruption function $g^{\rm confusion}$.
    Then,
    \begin{align} 
    \mathbb{P}\left(Y_{\rm test} \in \Chatnoisy (X_{\rm test}) \right) \geq 1-\alpha.
\end{align}
if and only if for all classes $j \in \{1, \ldots, K\}$,
\begin{equation}
    \label{eq:general-classification}
    \sum\limits_{j' = 1}^K \P(\tilY = j' \mid Y = j)\P(\ts \leq t \mid \tilY = j') \geq \P(s \leq t \mid Y = j).
\end{equation}
\end{prop}
The proof is below.

\begin{proof}
By law of total probability, $\P(s \leq t) = \E\left[ \P(s \leq t \mid Y = j) \right] = \sum_{j = 1}^K w_j \P(s \leq t \mid Y = j)$.
But under the noisy model, we have instead that 
\begin{equation}
    \P(\ts \leq t) = \E\left[ \P(\ts \leq t \mid \tilY = j') \right] =  \sum\limits_{j = 1}^K
    \sum\limits_{j' = 1}^K
    w_{j} \P(\tilY = j' \mid Y = j) \P(\ts \leq t \mid \tilY = j').
\end{equation}

We can write
\begin{equation}
    \P(s \leq t) - \P(\ts \leq t) = \sum\limits_{j = 1}^K
    \sum\limits_{j' = 1}^K
    w_{j} \P(\tilY = j' \mid Y = j) \P(\ts \leq t \mid \tilY = j') - \sum\limits_{j = 1}^K w_j \P(s \leq t \mid Y = j).
\end{equation}
Combining the sums and factoring, the above display equals
\begin{equation}
    \sum\limits_{j = 1}^K w_{j} \left( \sum\limits_{j' = 1}^K \left(\P(\tilY = j' \mid Y = j) \P(\ts \leq t \mid \tilY = j')\right) -  \P(s \leq t \mid Y = j)\right).
\end{equation}
We can factor this expression as
\begin{equation}
    \sum\limits_{j = 1}^K w_{j} \P(s \leq t \mid Y = j) \sum\limits_{j' = 1}^K \left(\P(\tilY = j' \mid Y = j) \frac{\P(\ts \leq t \mid \tilY = j')}{\P(s \leq t \mid Y = j)} - \frac{1}{K}\right).
\end{equation}
The stochastic dominance condition holds uniformly over all choices of base probabilities $w_j$ if and only if for all $j \in [K]$,
\begin{equation}
    \sum\limits_{j' = 1}^K \P(\tilY = j' \mid Y = j)\P(\ts \leq t \mid \tilY = j') \geq \P(s \leq t \mid Y = j).
\end{equation}
\end{proof}
Notice that the left-hand side of the above display is a convex mixture of the quantiles $\P(\ts \leq t \mid \tilY = j')$ for $j' \in [K]$.
Thus, the necessary and sufficient condition is for the noise distribution $\P(\tilY = j' \mid Y = j)$ to place sufficient mass on the classes $j'$ whose quantiles are larger than $\P(s \leq t \mid Y = j)$.
But of course, without assumptions on the model and score, the latter is unknown, so it is impossible to say which noise distributions will preserve coverage.

\subsubsection{Prediction Set Size Analysis}
\label{set-sizes-analysis}
When using the non-random APS scores with the oracle model, the prediction set size for a given $x$ can be expressed as $$k^*=\left\{ \min
        k: \sum_{j=1}^k\pi_{(j)}(x)\geq 1-\alpha\right\},$$ where $\pi_{y}(x)=\mathbb{P}(Y\mid X)$, the desired coverage level is $1-\alpha$, and $\pi_{(1)}(x)\geq \pi_{(2)}(x)\geq \dots \geq \pi_{(K)}(x)$ are the order statistics of $\pi_{1}(x),\pi_{2}(x), \dots ,\pi_{K}(x)$. Under the random flip noise, the noisy conditional class probabilities are given by $$\tilde{\pi}_y(x)=(1-\epsilon)\pi_y(x)+\frac{\epsilon}{K},$$ where $K$ is the number of labels and $\epsilon$ is the fraction of flipped labels. Therefore, the noisy set size is given by:   
        \begin{align}
        k^{*,\text{noisy}}
        &=
        \left\{ \min
        k: \sum_{j=1}^k\tilde{\pi}_{(j)}(x)\geq 1-\alpha\right\} \\ 
        &=\left\{ \min 
        k: \sum_{j=1}^k\pi_{(j)}(x)+\epsilon\left(\frac{k}{K}-\sum_{j=1}^k\pi_{(j)}(x) \right)\geq 1-\alpha\right\}\geq k^*.
        \end{align} As a result, we get $k^{*,\text{noisy}} \geq k^*$, where the term that controls the inflation of the noisy set size $\epsilon\left(\frac{k}{K}-\sum_{j=1}^k\pi_{(j)}(x)\right)$ is non-positive, and is a function of the noise level and the oracle conditional class probabilities.
\subsection{Distribution-Free Results}
\label{app:general-case}
We begin by proving Proposition~\ref{prop:general-case}.
\begin{proof}[Proof of Proposition~\ref{prop:general-case}]
    For convenience, assume the existence of probability density functions $\tilde{p}$ and $p$ for $\tilY$ and $Y$ respectively (these can be taken to be probability mass functions if $\Y$ is discrete).
    Also define the multiset of $Y$ values $E = \{ Y_1, ..., Y_n \}$ and the corresponding multiset of $\tilY$ values $\tilde{E}$.
    Take the set
    \begin{equation}
        \mathcal{A} = \{ y : \tilde{p}(y) > p(y) \}.
    \end{equation}
    Since $Y \overset{d}{\neq} \tilY$, we know that the set $\mathcal{A}$ is nonempty and $\P(\tilY \in \mathcal{A}) = \delta_1 > \P(Y \in \mathcal{A}) = \delta_2 \geq 0$.
    The adversarial choice of score function will be $s(x,y) = \ind{y \in \mathcal{A}^c}$; it puts high mass wherever the ground truth label is more likely than the noisy label.
    The crux of the argument is that this design makes the quantile smaller when it is computed on the noisy data than when it is computed on clean data, as we next show.
    
    Begin by noticing that, because $s(x,y)$ is binary, $\hat{q}_{\rm clean}$ is also binary, and therefore $\hat{q}_{\rm clean} > t \Longleftrightarrow \hat{q}_{\rm clean} = 1$.
    Furthermore, $\hat{q}_{\rm clean} = 1$ if and only if $| E \cap \mathcal{A} | \big| < \lceil (n+1)(1-\alpha) \rceil$.
    Thus, these events are the same, and for any $t \in (0,1]$,
    \begin{equation}
        \P\left( \qhatclean \geq t \right) = \P\Big(\big| E \cap \mathcal{A} \big| < \lceil (n+1)(1-\alpha) \rceil \Big).
    \end{equation}
    By the definition of $\mathcal{A}$, we have that $\P\Big(\big| E \cap \mathcal{A} \big| < \lceil (n+1)(1-\alpha) \rceil \Big) > \P\Big(\big| \tilde{E} \cap \mathcal{A} \big| < \lceil (n+1)(1-\alpha) \rceil \Big)$.
    Chaining the inequalities, we get
    \begin{equation}
        \P\left( \qhatclean \geq t \right) > \P\Big(\big| \tilde{E} \cap \mathcal{A} \big| < \lceil (n+1)(1-\alpha) \rceil \Big)   = \P\left( \qhat \geq t \right).
    \end{equation}
    Since $s_{n+1}$ is measurable with respect to $E$ and $\tilde{E}$, we can plug it in for $t$, yielding the conclusion.
\end{proof}
\begin{remark}
    In the above argument, if one further assumes continuity of the (ground truth) score function and $\P( \tilY \in \mathcal{A} ) = \P( Y \in \mathcal{A} ) + \rho$ for
    \begin{align}
        \rho = \inf\Big\{ \rho' > 0 \; : \; &\mathrm{BinomCDF}(n, \delta_1, \lceil (n+1)(1-\alpha) \rceil - 1) + \frac{1}{n} < \\ &\mathrm{BinomCDF}(n, \delta_2 + \rho', \lceil (n+1)(1-\alpha) \rceil - 1) \Big\},
    \end{align}
    then
    \begin{equation}
        \P(s_{n+1} \leq \qhat) < 1-\alpha.
    \end{equation}
\end{remark}
In other words, the noise must have some sufficient magnitude in order to disrupt coverage.

Next, we provide an additional impossibility result, similar to Proposition~\ref{prop:general-case}.
\begin{prop}
Given a score $s(x,y)=|\hat{f}(x) - y|$ and $\tilde{Y}=\beta Y,$ for $\beta\in(0,1)$. Assume further that $\hat{f}(x)=\mathbb{E}[\tilde{Y}\mid X=x],$ then there exists $\alpha_0 \in [0,1]$ such that for any $\alpha$ satisfying $1-\alpha \geq 1-\alpha_0: \mathbb{P}(Y\in \widehat{\mathcal{C}}_{\rm noisy} (X)) \leq 1-\alpha$.
\end{prop}
The proof is provided below: 
\begin{proof}
Denote $\mu_x := \mathbb{E}[Y\mid X=x]$. Then $$\hat{f}(x)=\mathbb{E}[\tilde{Y}\mid X=x]=\mathbb{E}[\beta Y\mid X=x]=\beta \mathbb{E}[Y\mid X=x]= \beta \mu_x.$$
$$\tilde{s}=|\hat{f}(x)-\tilde{Y}|=|\beta \mu_x-\beta Y|=\beta |\mu_x-Y| $$
$$s=|\hat{f}(x)-Y|=|\beta \mu_x-Y|$$
We will show that $F_{\tilde{s}}(t)\geq F_s(t), \ \forall t \in \mathbb{R}$. Notice that this is true conditional on $X$ (we do not write this explicitly to enhance clarity.)

Denote $G=\mu_x-Y, H=|G|, J=\beta \mu_x-Y$ and then $\tilde{s}=\beta H, s=|J|$. 
$$F_{G}(g)=\mathbb{P}(G\leq g)=\mathbb{P}(\mu_x-Y\leq g)=1-\mathbb{P}(Y\leq \mu_x-g)=1-F_{Y}(\mu_x-g).$$ 
$$F_{H}(h)=\mathbb{P}(H\leq h)=\mathbb{P}(|G|\leq h)=\mathbb{P}(-h \leq G \leq h)=F_{G}(h)-F_{G}(-h).$$
$$F_{J}(j)=\mathbb{P}(J\leq j)=\mathbb{P}(\beta \mu_x-Y\leq j)=1-\mathbb{P}(Y\leq \beta \mu_x-j)=1-F_{Y}(\beta \mu_x-j).$$
$$F_{\tilde{s}}(t)=F_{H}(\frac{t}{\beta})=F_G(\frac{t}{\beta})-F_{G}(-\frac{t}{\beta})=F_{Y}(\mu_x+\frac{t}{\beta})-F_{Y}(\mu_x-\frac{t}{\beta})$$
$$F_{s}(t)=F_{J}(t)-F_{J}(-t)=F_{Y}(\beta\mu_x+t)-F_{Y}(\beta \mu_x-t)$$
Let $t\geq 0,$ then:
\begin{align}
F_{\tilde{s}}(t)-F_s(t)=
&F_Y(\mu_x+\frac{t}{\beta})-F_Y(\mu_x-\frac{t}{\beta}) -\left( F_Y(\beta\mu_x+t)-F_Y(\beta \mu_x-t)\right) \\
&=F_Y(\mu_x+\frac{t}{\beta})-F_Y(\mu_x-\frac{t}{\beta}) -\left( F_Y(\beta(\mu_x+\frac{t}{\beta}))-F_Y(\beta(\mu_x-\frac{t}{\beta}))\right)
\end{align}

For $t\geq |\beta \mu_x|$ from the monotonicity of $F_Y$ we get: $ F_Y(\mu_x+\frac{t}{\beta})\geq F_Y(\beta(\mu_x+\frac{t}{\beta}))$ and $F_Y(\beta(\mu_x-\frac{t}{\beta}))\geq F_Y(\mu_x-\frac{t}{\beta})$. Therefore: $F_{\tilde{s}}(t)\geq F_s(t)$ or equivalently $\mathbb{P}(\tilde{s}\leq t)\geq \mathbb{P}(s\leq t).$ 
Denote $\alpha_0(x) := 1-\mathbb{P}(\tilde{s} \leq |\beta \mu_x|)$. Suppose that $\alpha$ satisfies $1-\alpha \geq 1-\alpha_0(x)$. Denote by $\hat{q}_{\rm noisy}(x)$ the $1-\alpha$ quantile of the noisy scores $\tilde{s}$, conditional on $X=x$. Then, we get that $\hat{q}_{\rm noisy}(x) \geq |\beta \mu_x|$, and thus:
$$1-\alpha = \mathbb{P}(\tilde{s} \leq \hat{q}_{\rm noisy}(x) ) \geq \mathbb{P}(s \leq \hat{q}_{\rm noisy}(x)).$$

Lastly, we return to consider the notations of conditioning on $X$. Denote $\alpha_0 = \inf_{x\in \mathcal{X}}(\alpha_0 (x))$. Suppose that $\alpha \in [0,1]$ satisfies $1-\alpha \geq 1-\alpha_0$. Denote by $\hat{q}_{\rm noisy}$ the $1-\alpha$ quantile of the noisy scores $\tilde{s}$, marginally on $x\in \mathcal{X}$. Then, we obtain:
$$1-\alpha = \mathbb{P}(\tilde{Y}\in \mathcal{C}_{\rm noisy} (X)) = \mathbb{P}(\tilde{s}\leq \hat{q}_{\rm noisy}) \geq \mathbb{P}(s\leq \hat{q}_{\rm noisy}) = \mathbb{P}(Y\in \mathcal{C}_{\rm noisy} (X)),$$
where above, the probability is taken marginally over $x\in \mathcal{X}$.


\end{proof}

We commence by proving Corollary~\ref{cor:tv}.
\begin{proof}[Proof of Corollary~\ref{cor:tv}]
    This a consequence of the TV bound from~\cite{barber2022conformal} with weights identically equal to $1$.
\end{proof}
Unfortunately, getting such a TV bound requires a case by case analysis.
It's not even straightforward to get a TV bound under strong Gaussian assumptions.
\begin{prop}[No general TV bound]
    Assume $Y \sim \mathcal{N}(0,\tau^2)$ and $\tilY = y + Z$, where $Z \sim \mathcal{N}(0,\sigma^2)$.
    Then $D_{\rm TV}(Y,\tilY) \overset{\tau \to 0}{\to} 1$.
\end{prop}
\begin{proof}
  \begin{equation}
      \mathrm{TV}(\mathcal{N}(0,\tau^2),\mathcal{N}(0,\tau^2+\sigma^2)) = \int\limits_{-\infty}^\infty \Big|e^{-x^2/\tau^2}-e^{-x^2/(\tau^2+\sigma^2)}\Big| dx \overset{\tau \to 0}{\to} 1.
  \end{equation}
\end{proof}

\subsection{False-Negative Risk}

In this section, we prove all FNR robustness propositions. Here, we suppose that the response $Y\in[0,1]^n$ is a binary vector of size $n\in\mathbb{N}$, where $Y_i=1$ indicates that the $i$-th label is present. We further suppose a vector-flip noise model from \eqref{eq:vector_flip}, where $\varepsilon$ is a binary random vector of size $n$ as well. These notations apply for segmentation tasks as well, by flattening the response matrix into a vector. The prediction set $C(X)\subseteq \{1,...,n \}$ contains a subset of the labels. We begin by providing additional theoretical results and then turn to the proofs.

\subsubsection{Additional Theoretical Results}\label{sec:additional_fnr_results}

\begin{prop}\label{prop:fnr_independent}
Let $\Chatnoisy(\Xtest)$ be a prediction set that controls the FNR risk of the noisy labels at level $\alpha$. Suppose that
\begin{enumerate}
    \item The prediction set contains the most likely labels in the sense that for all $x\in \mathcal{X}$, $k\in \Chatnoisy(x)$, $i\notin \Chatnoisy(x)$ and $m\in \mathbb{N}$:
  \begin{equation}
    \mathbb{P}\left({Y}_k=1 \middle| \sum_{j \ne i,k} {{Y}_j}=m, X=x\right) \geq \mathbb{P}\left({Y}_i=1 \middle| \sum_{j \ne i,k} {{Y}_j}=m, X=x \right).
  \end{equation}
    \item For a given input $X=x$, the noise level of all response elements is the same, i.e., for all $i\ne j\in\{1\dots K\}$: $\mathbb{P}(\epsilon_i=1 \mid X=x)=\mathbb{P}(\epsilon_j=1 \mid X=x)$.
    \item The noise is independent of $Y\mid X=x$ in the sense that $\epsilon \indep Y \mid X=x$ for all  $x\in\mathcal{X}$.
    \item The noises of different labels are independent of each other given $X$, i.e., $\epsilon_i \indep \epsilon_j \mid X=x$ for all $i\ne j\in \{1, \dots K \}$ and $x\in\mathcal{X}$.
\end{enumerate}
  Then
  \begin{equation}
    \mathbb{E}\left[L^{\rm FNR}\left(\Ytest,\Chatnoisy(\Xtest)\right)\right] \leq \alpha.
  \end{equation}
\end{prop}

\subsubsection{General Derivations}
\label{app:general-derivations}
In this section, we formulate and prove a general lemma that is used to prove label-noise robustness with the false-negative rate loss. 


\begin{lemma}\label{lem:general_fnr}
Suppose that $x\in\mathcal{X}$ is an input variable and $C(x)$ is a prediction set.
Denote by $\beta_i$ the noise level at the $i$-th element: $\beta_i := \mathbb{P}(\varepsilon_i=1 \mid X=x)$.
We define:
\begin{equation}
    e_{k,i}(y):= \beta_i {{y_{k}\mathbb{E} \left[\frac{1 } {\sum_{j=1}^n {\tilde{Y}_{j}}}\middle| \varepsilon_i = 1, Y=y \right]}}.
\end{equation}
If for all $k\in C(x)$ and $i\notin C(x)$:
\begin{equation}
\mathbb{E}\left[  \frac{e_{k,i}(Y)-e_{i,k}(Y)}{\sum_{j=1}^n Y_{j}} \middle| X=x\right] \geq 0
\end{equation}
Then $C(x)$ achieves conservative conditional risk:
\begin{equation}
\mathbb{E}\left[L^{\text{FNR}}(Y, C(X)) \mid X=x \right] \leq \mathbb{E}\left[L^{\text{FNR}}(\tilde{Y}, C(X)) \mid X=x\right]. 
\end{equation}
Furthermore, $C(X)$ achieves valid marginal risk:
\begin{equation}
\mathbb{E}\left[L^{\text{FNR}}(Y, C(X))  \right] \leq \mathbb{E}\left[L^{\text{FNR}}(\tilde{Y}, C(X))\right]. 
\end{equation}
\end{lemma}
\begin{proof}
For ease of notation, we omit the conditioning on $X=x$. That, we take some $x \in \mathcal{X}$ and treat $Y$ as $Y\mid X=x$. We also denote the prediction set as $C = C(x)$. Denote:
\begin{equation}
\delta := L^{\text{FNR}}(\tilde{Y},C) - L^{\text{FNR}}({Y},C). 
\end{equation}
Our goal is to show that:
\begin{equation}
\mathbb{E}\left[\delta\right] \geq 0. 
\end{equation}
If $C =\emptyset$ or $C= \{1,...,n\}$ then the proposition is trivially satisfied.
Therefore, for the rest of this proof, we assume that $C\ne \emptyset,\{1,...,n\}$. We begin by developing $\delta$ as follows.

\begin{equation}
\begin{split}
\delta &= L^{\text{FNR}}(\tilde{Y},C) - L^{\text{FNR}}({Y},C) \\
&= \frac{\sum_{j\in C} {Y_{j}}}{ \sum_{j} Y_{j} } - \frac{\sum_{j \in C} {[Y_{j} - 2\varepsilon_{j}Y_{j} + \varepsilon_{j}] }}{ \sum_{j} {Y_{j} - 2\varepsilon_{j}Y_{j} + \varepsilon_{j}} }\\
& = \frac{\sum_{i}  \sum_{j\in C} { Y_{j}[Y_{i} - 2\varepsilon_{i}Y_{i} + \varepsilon_{i}] } - \sum_{i} \sum_{j \in C} {Y_i[Y_{j} - 2\varepsilon_{j}Y_{j} + \varepsilon_{j}] }}{(\sum_{j} Y_{j})(\sum_{j} {Y_{j} - 2\varepsilon_{j}Y_{j} + \varepsilon_{j}})} \\
& = \frac{\sum_{i}  \sum_{j\in C} { Y_{j}Y_{i} - 2 Y_{j}\varepsilon_{i}Y_{i} +  Y_{j}\varepsilon_{i} } - \sum_{i} \sum_{j \in C} {Y_i Y_{j} - 2Y_i\varepsilon_{j}Y_{j} + Y_i\varepsilon_{j} }}{(\sum_{j} Y_{j})(\sum_{j} {Y_{j} - 2\varepsilon_{j}Y_{j} + \varepsilon_{j}})} \\
& = \frac{\sum_{i}  \sum_{j\in C} { Y_{j}Y_{i} - 2 Y_{j}\varepsilon_{i}Y_{i} +  Y_{j}\varepsilon_{i} - [Y_i Y_{j} - 2Y_i\varepsilon_{j}Y_{j} + Y_i\varepsilon_{j}] }}{(\sum_{j} Y_{j})(\sum_{j} {Y_{j} - 2\varepsilon_{j}Y_{j} + \varepsilon_{j}})} \\
& = \frac{\sum_{i}  \sum_{j\in C} { Y_{j}\varepsilon_{i} - Y_i\varepsilon_{j} }}{(\sum_{j} Y_{j})(\sum_{j} {Y_{j} - 2\varepsilon_{j}Y_{j} + \varepsilon_{j}})} \\
& = \frac{(\sum_{i} {\varepsilon_{i}})  (\sum_{j\in C} { Y_{j}) } - (\sum_{i} {Y_{i}})  (\sum_{j\in C} { \varepsilon_{j})} } {(\sum_{j} Y_{j})(\sum_{j} {Y_{j} - 2\varepsilon_{j}Y_{j} + \varepsilon_{j}})}. \\
\end{split}
\end{equation}
Without loss of generality, we assume that $C=\{1,...,p\}$. We now compute the expectation of each term separately.
\begin{equation}
\begin{split}
&\mathbb{E} \left[\frac{(\sum_{i} {\varepsilon_{i}}) (\sum_{k=1}^p{Y_{k}}) } {(\sum_{j} Y_{j})(\sum_{j} {Y_{j} - 2\varepsilon_{j}Y_{j} + \varepsilon_{j}})} \middle| Y=y \right]\\
&=
\sum_{i} {\mathbb{E} \left[\frac{{\varepsilon_{i}}(\sum_{k=1}^p{Y_{k}}) } {(\sum_{j} Y_{j})(\sum_{j} {Y_{j} - 2\varepsilon_{j}Y_{j} + \varepsilon_{j}})}\middle| Y=y\right]}  \\
&= \sum_{i} {\frac{(\sum_{k=1}^p{y_{k}})}{\sum_{j} y_{j}}\mathbb{E}\left[\frac{\varepsilon_{i} } {\sum_{j} {Y_{j} - 2\varepsilon_{j}Y_{j} + \varepsilon_{j}}}\middle|  Y=y\right]}   \\
&= \sum_{i} {\frac{(\sum_{k=1}^p{y_{k}})}{\sum_{j} y_{j}}\mathbb{E} \left[\frac{\varepsilon_{i} } {\sum_{j} {Y_{j} - 2\varepsilon_{j}Y_{j} + \varepsilon_{j}}} \middle| \varepsilon_i=1, Y=y \right]\mathbb{P}(\varepsilon_i=1)}   \\
&= \frac{1}{\sum_{j} y_{j}}\sum_{i} {\beta_i\left(\sum_{k=1}^p{y_{k}}\right) \mathbb{E} \left[\frac{1 } {(1-Y_i) + \sum_{j\ne i} {Y_{j} - 2\varepsilon_{j}Y_{j} + \varepsilon_{j}}} \middle| \varepsilon_i = 1, Y=y\right]}   \\
&=  \frac{1}{\sum_{j} y_{j}}\sum_{i} {{\sum_{k=1}^p \beta_i y_{k}\mathbb{E}\left[\frac{1 } {(1-Y_i) + \sum_{j\ne i} {Y_{j} - 2\varepsilon_{j}Y_{j} + \varepsilon_{j}}}\middle| \varepsilon_i = 1, Y=y \right]}}   \\
\end{split}
\end{equation}

\begin{equation}
\begin{split}
&\mathbb{E}\left[\frac{(\sum_{i} {Y_{i}}) (\sum_{k=1}^p\varepsilon_{k}) } {(\sum_{j} Y_{j})(\sum_{j} {Y_{j} - 2\varepsilon_{j}Y_{j} + \varepsilon_{j}})}\middle| Y=y\right] \\
&=\sum_{i} {\mathbb{E}\left[\frac{{Y_{i}} (\sum_{k=1}^p\varepsilon_{k}) } {(\sum_{j} Y_{j})(\sum_{j} {Y_{j} - 2\varepsilon_{j}Y_{j} + \varepsilon_{j}})}\middle| Y=y\right]}  \\
&= \sum_{i} {\frac{y_i}{\sum_{j} y_{j}}\mathbb{E}\left[\frac{(\sum_{k=1}^p\varepsilon_{k}) } {\sum_{j} {Y_{j} - 2\varepsilon_{j}Y_{j} + \varepsilon_{j}}}\middle| Y=y\right]}   \\
&= \sum_{i} {\frac{y_i}{\sum_{j} y_{j}}\sum_{k=1}^p\mathbb{E}\left[\frac{\varepsilon_{k}} {\sum_{j} {Y_{j} - 2\varepsilon_{j}Y_{j} + \varepsilon_{j}}}\middle| Y=y\right]}   \\
&=  \sum_{i} {\frac{y_i}{\sum_{j} y_{j}}\sum_{k=1}^p { \beta_k \mathbb{E} \left[\frac{\varepsilon_{k}} {\sum_{j} {Y_{j} - 2\varepsilon_{j}Y_{j} + \varepsilon_{j}}} \middle| \varepsilon_k = 1, Y=y \right]}}   \\
&= \sum_{i} {\frac{y_i}{\sum_{j} y_{j}}\sum_{k=1}^p {\beta_k \mathbb{E} \left[\frac{1} {(1-Y_k)+ \sum_{j\ne k} {Y_{j} - 2\varepsilon_{j}Y_{j} + \varepsilon_{j}}} \middle| \varepsilon_k = 1, Y=y \right]}}   \\
&= \frac{1}{\sum_{j} y_{j}}\sum_{i} {\sum_{k=1}^p \beta_k y_i \mathbb{E} \left[\frac{1} {(1-Y_k)+ \sum_{j\ne k} {Y_{j} - 2\varepsilon_{j}Y_{j} + \varepsilon_{j}}}\middle| \varepsilon_k = 1, Y=y \right]}   \\
\end{split}
\end{equation}
We now compute the expected 
value of $\delta$ for a given vector of labels $Y=y$.
\begin{equation}
\begin{split}
\mathbb{E} \left[\delta \middle| Y=y \right]&= \mathbb{E} \left[\frac{(\sum_{i} {\varepsilon_{i}}) (\sum_{k=1}^p{Y_{k}}) } {(\sum_{i} Y_{i})(\sum_{j} {Y_{j} - 2\varepsilon_{j}Y_{j} + \varepsilon_{j}})} \middle| Y=y \right]-\mathbb{E} \left[\frac{(\sum_{i} {Y_{i}}) (\sum_{k=1}^p\varepsilon_{k}) } {(\sum_{i} Y_{i})(\sum_{j} {Y_{j} - 2\varepsilon_{j}Y_{j} + \varepsilon_{j}})} \middle|  Y=y \right] \\
&= \frac{1}{\sum_{j} y_{j}} \left( \sum_{i} {{\sum_{k=1}^p e_{k,i}(y)}}  - e_{i,k}(y) \right) \\
&= \frac{1}{\sum_{j} y_{j}} \left[ \left(\sum_{i=1}^p {\sum_{k=1}^p  e_{k,i}(y)} +\sum_{i=p+1}^{n} {\sum_{k=1}^p  e_{k,i}(y)} \right) - \left(\sum_{i=1}^p {\sum_{k=1}^p e_{i,k}(y)} + \sum_{i=p+1}^{n} {\sum_{k=1}^p e_{i,k}(y)} \right) \right] \\
&= \frac{1}{\sum_{j} y_{j}} \left[ \sum_{i=p+1}^{n} {\sum_{k=1}^p  e_{k,i}(y)}  - \sum_{i=p+1}^{n} {\sum_{k=1}^p e_{i,k}(y)} \right] \\
&= \frac{1}{\sum_{j} y_{j}} \left[ \sum_{i=p+1}^{n} {\sum_{k=1}^p  e_{k,i}(y)-e_{i,k}(y)} \right]\\
&=  \sum_{i=p+1}^{n} {\sum_{k=1}^p  \frac{e_{k,i}(y)-e_{i,k}(y)}{\sum_{j} y_{j}}}\\
\end{split}
\end{equation}
Finally, we get valid risk conditioned on $X=x$:

\begin{equation}
\mathbb{E} \left[\delta \right]= \mathbb{E} \left[\sum_{i=p+1}^{n} {\sum_{k=1}^p  \frac{e_{k,i}(Y)-e_{i,k}(Y)}{\sum_{j} Y_{j}}}\right] =\sum_{i=p+1}^{n} {\sum_{k=1}^p  \mathbb{E} \left[ \frac{e_{k,i}(Y)-e_{i,k}(Y)}{\sum_{j} Y_{j}}\right]}\geq 0.
\end{equation}
Above, the last inequality follows from the assumption of this lemma. We now marginalize the above result to obtain valid marginal risk:
\begin{equation}
\mathbb{E} \left[\delta \right]=\mathbb{E}_{X} \left[ \mathbb{E}_{\tilde{Y},Y\mid X=x} \left[\delta \mid X=x\right]\right]\geq \mathbb{E}_{X} \left[ 0\right] \geq 0.
\end{equation}
\end{proof}

\subsubsection{Proof of Proposition~\ref{prop:fnr_independent}}
\begin{proof}
As in Lemma~\ref{lem:general_fnr}, we omit the conditioning on $X$ and treat $Y$ as $Y \mid X=x$. 
Without loss of generality, we suppose that $C=\{1,...,p\}$, where $\mathbb{P}(\tilde{Y}_i=1)\geq \mathbb{P}(\tilde{Y}_{i+1}=1)$ for all $i\in \{1,...,n-1\}$. Since $\forall i\in \{1,...,n-1\}: \mathbb{P}(\varepsilon_i =1) \leq 0.5$, the order of class probabilities is preserved under the corruption, meaning that 
$$\forall i\in \{1,...,n-1\}: \mathbb{P}({Y}_i=1)\geq \mathbb{P}({Y}_{i+1}=1).$$ 
We now compute $\mathbb{E} \left[{\frac{e_{k,i}(Y)-e_{i,k}(Y)}{\sum_{j} Y_{j}}} \right]$ for $k < i$:
\begin{equation}
\begin{split}
& \mathbb{E} \left[\frac{e_{k,i}(Y)-e_{i,k}(Y)}{\sum_{i} Y_{i}}  \right] \\
& = \mathbb{E} \left[\frac{{\beta_i{Y_k}\mathbb{E} \left[\frac{1 } {(1-Y_i) + \sum_{j\ne i} {Y_{j} - 2\varepsilon_{j}Y_{j} + \varepsilon_{j}}} \middle| Y=Y\right]} - {\beta_k{Y_i}\mathbb{E} \left[\frac{1 } {(1-Y_k) + \sum_{j\ne k} {Y_{j} - 2\varepsilon_{j}Y_{j} + \varepsilon_{j}}} \mid Y=Y\right] }}{\sum_{j} Y_j}   \right] \\
&=\sum_{y\in\mathcal{Y}}\mathbb{P}(Y=y  )\left[{\frac{\beta_i y_k}{\sum_{j} {y_j}}\mathbb{E} \left[\frac{1 } {(1-y_i) + \sum_{j\ne i} 
{y_{j} - 2\varepsilon_{j}y_{j} + \varepsilon_{j}}} \middle| Y=y\right]} \right. \\
& \left. - {\frac{\beta_k y_i}{\sum_{j} y_j}\mathbb{E}\left[\frac{1 } {(1-y_k) + \sum_{j\ne k} {y_{j} - 2\varepsilon_{j}y_{j} + \varepsilon_{j}}}\right] } \middle| Y=y \right] \\
&=\sum_{\{y\in\mathcal{Y}: y_i \ne y_k\} }\mathbb{P}(Y=y  )\left[{\frac{\beta_i y_k}{\sum_{j} {y_j}}\mathbb{E} \left[\frac{1 } {(1-y_i) + \sum_{j\ne i} 
{y_{j} - 2\varepsilon_{j}y_{j} + \varepsilon_{j}}} \middle| Y=y \right]}\right] \\
&- \sum_{\{y\in\mathcal{Y}: y_i \ne y_k\} }\mathbb{P}(Y=y  )\left[{\frac{\beta_k y_i}{\sum_{j} {y_j}}\mathbb{E} \left[\frac{1 } {(1-y_k) + \sum_{j\ne k} 
{y_{j} - 2\varepsilon_{j}y_{j} + \varepsilon_{j}}} \middle| Y=y \right]}\right]\\
&=\sum_{\{y\in\mathcal{Y}: y_k=1,y_i=0\}}\mathbb{P}(Y=y  ){\frac{\beta_i}{\sum_{j} {y_j}}\mathbb{E} \left[\frac{1 } {2 - \varepsilon_{k}+ \sum_{j\ne i,k} 
{y_{j} - 2\varepsilon_{j}y_{j} + \varepsilon_{j}}}\middle| Y=y\right]}\\
&- \sum_{\{y\in\mathcal{Y}: y_i=1, y_k=0\} }\mathbb{P}(Y=y  ){\frac{\beta_k}{\sum_{j} {y_j}}\mathbb{E} \left[\frac{1} {2 -\varepsilon_i + \sum_{j\ne k,i} 
{y_{j} - 2\varepsilon_{j}y_{j} + \varepsilon_{j}}} \middle| Y=y\right]}
\end{split}
\end{equation}
To simplify the equation, denote by $y^*$ the vector $y$ with indexes $i,k$ swapped, that is:
\begin{equation*}
y^*_j = \begin{cases}
y_j &j \ne i,k,\\
y_i &j = k,\\
y_k &j = i.\\
\end{cases}
\end{equation*}
Notice that $\beta_k \mathbb{E} \left[\frac{1} {2 -\varepsilon_i + \sum_{j\ne k,i} 
{y_{j} - 2\varepsilon_{j}y_{j} + \varepsilon_{j}}} \middle| Y=y\right]=\beta_i \mathbb{E} \left[\frac{1} {2 -\varepsilon_k+ \sum_{j\ne k,i} 
{y_{j} - 2\varepsilon_{j}y_{j} + \varepsilon_{j}}} \middle| Y=y\right]$ since $\varepsilon_i \stackrel{d}{=} \varepsilon_k$ and these variables are independent of $Y$ and of $\varepsilon_j$ for $j \ne i,k$.
Further denote $\gamma_m = \beta \left( \sum_{j}{y_j}\right) {\mathbb{E} \left[\frac{1} { 2- \varepsilon_{k}+ \sum_{j\ne i,k} {y_{j} - 2\varepsilon_{j}y_{j} + \varepsilon_{j}}} \middle| Y=y\right]}$ for $\sum_{j\ne i,k}y_j = m$ and $y_i \ne y_k$. Notice that $\gamma_m$ is well defined as $\beta {\mathbb{E} \left[\frac{1} { 2- \varepsilon_{k}+ \sum_{j\ne i,k}
{y_{j} - 2\varepsilon_{j}y_{j} + \varepsilon_{j}}} \middle| Y=y \right]}$ has the same value for every $y$ such that $\sum_{j\ne i,k}y_j = m$ since $y$ and $\varepsilon$ are independent. Also $\sum_{j}{y_j}=m+1$ if $\sum_{j\ne i,k}y_j = m$ and $y_i \ne y_k$. Lastly, we denote  ${\gamma'}_m = \gamma_m\mathbb{P}\left(\sum_{j\ne i,k}Y_j = m \right)$.
We continue computing $\mathbb{E} \left[\frac{e_{k,i}(Y)-e_{i,k}(Y)}{\sum_{j}{Y_j}}\right]$ for $k < i$:
\begin{equation}
\begin{split}
& \mathbb{E}\left[\frac{e_{k,i}(Y)-e_{i,k}(Y)}{\sum_{j}{Y_j}} \right]\\
&=\sum_{\{y\in\mathcal{Y}:y_k=1,y_i=0\}}(\mathbb{P}(Y=y)-\mathbb{P}(Y=y^*)){\frac{\beta}{\sum_{j}{y_j}}\mathbb{E} \left[\frac{1 } { 2- \varepsilon_{k}+ \sum_{j\ne i,k} 
{y_{j} - 2\varepsilon_{j}y_{j} + \varepsilon_{j}}} \middle| Y=y \right]}\\
&=\sum_{m}\sum_{\{y\in\mathcal{Y}, y_k=1,y_i=0, \sum_{j\ne i,k} y_j=m\}}(\mathbb{P}(Y=y)-\mathbb{P}(Y=y^*)){\gamma_m}\\
&=\sum_{m}\gamma_m\sum_{\{y\in\mathcal{Y}, y_k=1,y_i=0, \sum_{j\ne i,k} y_j=m\}}\left[\mathbb{P}(Y=y)-\mathbb{P}(Y=y^*)\right]\\
&=\sum_{m}\gamma_m \left[\sum_{\{y\in\mathcal{Y}, y_k=1,y_i=0, \sum_{j\ne i,k} y_j=m\}}\mathbb{P}(Y=y) - \sum_{\{y\in\mathcal{Y}, y_i=1,y_k=0, \sum_{j\ne i,k} y_j=m\}}\mathbb{P}(Y=y) \right]\\
&=\sum_{m}\gamma_m \left[ \mathbb{P}\left(Y_k=1,Y_i=0, \sum_{j\ne i,k}Y_j=m\right) -  \mathbb{P}\left(Y_k=0,Y_i=1, \sum_{j\ne i,k}Y_j=m\right) \right]\\
&=\sum_{m}{\gamma'}_m \left[ \mathbb{P}\left(Y_k=1,Y_i=0\mid \sum_{j\ne i,k}Y_j=m\right) -  \mathbb{P}\left(Y_k=0,Y_i=1 \mid \sum_{j\ne i,k}Y_j=m \right) \right]\\
&=\sum_{m}{\gamma'}_m \left[\mathbb{P}\left(Y_k=1,Y_i=0\middle| \sum_{j\ne i,k}Y_j=m \right) + \mathbb{P}\left(Y_k=1,Y_i=1\middle| \sum_{j\ne i,k}Y_j=m \right) \right. \\ &- \left. \mathbb{P}\left(Y_k=1,Y_i=1\middle| \sum_{j\ne i,k}Y_j=m \right) -  \mathbb{P}\left(Y_k=0,Y_i=1 \middle| \sum_{j\ne i,k}Y_j=m\right) \right] \\
&=\sum_{m}{\gamma'}_m \left[ \mathbb{P}\left(Y_k=1\middle| \sum_{j\ne i,k}Y_j=m \right) -  \mathbb{P}\left(Y_i=1 \middle| \sum_{j\ne i,k}Y_j=m \right) \right]\\
\end{split}
\end{equation}
We assume that $\forall m: \mathbb{P}\left(Y_k=1\middle| \sum_{j\ne i,k}Y_j=m \right) \geq  \mathbb{P}\left(Y_i=1 \middle| \sum_{j\ne i,k}Y_j=m \right)$ and therefore:
\begin{equation}
\begin{split}
    & \mathbb{E} \left[\frac{e_{k,i}(Y)-e_{i,k}(Y)}{\sum_{j}{Y_j}}\right] \\
    &=\sum_{m}{\gamma'}_m \left[ \mathbb{P}\left(Y_k=1\middle| \sum_{j\ne i,k}Y_j=m \right) -  \mathbb{P}\left(Y_i=1 \middle| \sum_{j\ne i,k}Y_j=m \right) \right] \\
    &\geq 0.
\end{split}
\end{equation}
According to Lemma~\ref{lem:general_fnr}, the above concludes the proof.
\end{proof}

\subsubsection{Multi-label Classification} \label{app:proof-multi-label}
Here we prove Proposition~\ref{prop:multi_label}.

\begin{proof}[Proof of Proposition~\ref{prop:multi_label}]
As in Lemma~\ref{lem:general_fnr}, we omit the conditioning on $X$ and treat $Y$ as $Y \mid X=x$. 
Without loss of generality, we suppose that $C=\{1,...,p\}$, where $\mathbb{P}(\tilde{Y}_i=1)\geq \mathbb{P}(\tilde{Y}_{i+1}=1)$ for all $i\in \{1,...,n-1\}$. Since $\forall i\in \{1,...,n-1\}: \mathbb{P}(\varepsilon_i =1) \leq 0.5$, the order of class probabilities is preserved under the corruption, meaning that 
$$\forall i\in \{1,...,n-1\}: \mathbb{P}({Y}_i=1)\geq \mathbb{P}({Y}_{i+1}=1).$$
Since $Y$ is a deterministic function of $X$, we get that
$Y$ is some non-increasing constant binary vector. Suppose that $k\in C$ and $i\notin C$. Since $C$ contains the most likely labels, these indexes satisfy $k < i$. We now compare $e_{k,i}(y)$ to $e_{i,k}$. According to the definition of $e_{k,i}$:
\begin{equation}
e_{k,i}:=\beta_i {{y_{k}\mathbb{E} \left[\frac{1 } {\sum_{j=1}^n {\tilde{Y}_{j}}}\middle| \varepsilon_i = 1, Y=y \right]}}.
\end{equation}
Since ${\sum_{j=1}^n {\tilde{Y}_{j}}}$ is a assumed to be a constant, $e_{k,i}(y)$ and $e_{i,k}$ may differ only in the values of $y_k, y_i$ and $\beta_k, \beta_i$. We go over all four combinations of $y_k$ and $y_i$.
\begin{enumerate}
    \item If $y_k=0, y_i=0$, then $e_{k,i}(y)=e_{i,k}(y)=0$.
    \item If $y_k=1, y_i=0$, then $e_{k,i}(y) \geq 0$ and $e_{i,k}(y)=0$.
    \item If $y_k=0, y_i=1$, then we get a contradiction to the property of $Y$ being non-increasing, and thus this case is not possible.
    \item If $y_k=1, y_i=1$, then $e_{k,i}(y)\geq e_{i,k}(y)=0$ since $\beta_i \geq \beta_k$:
    \begin{equation}
\begin{split}
&\mathbb{P}(\varepsilon_k = 0)=\mathbb{P}(\tilde{Y}_k = 1) \geq \mathbb{P}(\tilde{Y}_i = 1) = \mathbb{P}(\varepsilon_i = 0) \\
& \Rightarrow  1- \mathbb{P}(\varepsilon_k = 1) \geq 1- \mathbb{P}(\varepsilon_i = 1)\\
& \Rightarrow \beta_k = \mathbb{P}(\varepsilon_k = 1) \leq \mathbb{P}(\varepsilon_i = 1) = \beta_i.
\end{split}
\end{equation}
Therefore, under all cases, we get that  $e_{k,i}(y) \geq e_{i,k}(y)$ for all $k\in C, i \notin C$, and thus:
\begin{equation}
\mathbb{E}\left[  \frac{e_{k,i}(Y)-e_{i,k}(Y)}{\sum_{j=1}^n Y_{j}} \middle| X=x\right] \geq 0.
\end{equation}
Finally, by applying Lemma~\ref{lem:general_fnr}, we obtain valid conditional risk.
\end{enumerate}

\end{proof}

\subsubsection{Segmentation} \label{app:proof-segmentation}

\begin{proof}[Proof of Proposition~\ref{prop:segmentation}]
This proposition is a special case of Proposition~\ref{prop:fnr_independent} and thus valid risk follows directly from this result.
\end{proof}

\subsection{Risk Control for Regression Tasks}

\subsubsection{General Risk Bound for Regression Tasks}\label{sec:regression_taylor_risk_bounds}

Here we prove Proposition~\ref{prop:general_regression_bounds}.
\begin{proof}[Proof of Proposition~\ref{prop:general_regression_bounds}]
For ease of notation, we omit the conditioning on $X=x$. In other words, we treat $Y$ as $Y\mid X=x$ for some $x \in \mathcal{X}$. Given a prediction set $\hat{C}(x)$, we consider the loss as a function of $Y$, where  $\hat{C}(x)$ is fixed, and denote it by:
\begin{equation}
L(Y) := L(Y, \hat{C}(x)).
\end{equation}

We expand $L(Y+\varepsilon)$ using Taylor's expansion:
\begin{equation}
\begin{split}
L(Y+\varepsilon) &= L(Y) + \varepsilon L'(Y) +  \frac{1}{2}\varepsilon^2 L''(\xi),
\end{split}
\end{equation}
where $\xi$ is some real number between $Y$ and $Y+\varepsilon$.
\begin{equation}
\begin{split}
\delta &:= L(\Tilde{Y}) -  L(Y) \\
&= L(Y+\varepsilon) -  L(Y) \\
&= \varepsilon L'(Y) +  \frac{1}{2}\varepsilon^2 L''(\xi). \\
\end{split}
\end{equation}
We now develop each term separately. Since $\varepsilon \indep Y$ it follows that $\varepsilon \indep L'(Y)$.
\begin{equation}
\begin{split}
\mathbb{E}\left[\varepsilon L'(Y)\right] = \mathbb{E}\left[\varepsilon\right] \mathbb{E}\left[L'(Y)\right]= 0 \mathbb{E}\left[L'(Y)\right] = 0
\end{split}
\end{equation}
Since $L''(y)$ is bounded by $q \leq L''(y) \leq Q$, we get the following:
\begin{equation}
\begin{split}
\frac{1}{2}q\text{Var}\left[\varepsilon  \right] =
\frac{1}{2}q\mathbb{E}\left[\varepsilon^2   \right]  
=\mathbb{E}\left[\frac{1}{2} \varepsilon^2  q_x  \right]  
\leq \mathbb{E}\left[\frac{1}{2} \varepsilon^2  L''(\xi)  \right] \leq 
\mathbb{E}\left[\frac{1}{2} \varepsilon^2  Q_x  \right] =
\frac{1}{2}Q\mathbb{E}\left[\varepsilon^2   \right] =
\frac{1}{2}Q\text{Var}\left[\varepsilon  \right]
\end{split}
\end{equation}
We get that:
\begin{equation}
\begin{split}
\mathbb{E}[\delta] = \mathbb{E}\left[\varepsilon L'(Y) +  \frac{1}{2}\varepsilon^2 L''(\xi) \right]
= 0 + \mathbb{E}\left[\frac{1}{2}\varepsilon^2 L''(\xi) \right].
\end{split}
\end{equation}
Therefore:
\begin{equation}
\begin{split}
& \frac{1}{2}q\text{Var}\left[\varepsilon  \right] 
\leq 
\mathbb{E}[\delta]  
\leq
\frac{1}{2}Q\text{Var}\left[\varepsilon \right]. \\
& \Rightarrow
\frac{1}{2}q\text{Var}\left[\varepsilon \right] 
\leq 
\mathbb{E}\left[L(\tilde{Y}, \hat{C}(X)) - L({Y}, \hat{C}(X))\right] 
\leq
\frac{1}{2}Q\text{Var}\left[\varepsilon \right].
\end{split}
\end{equation}
We now turn to consider the conditioning on $X=x$ and obtain marginalized bounds by taking the expectation over all $X$:
\begin{equation}
\begin{split}
 \alpha - \frac{1}{2}Q\text{Var}\left[\varepsilon  \right] 
\leq 
 \mathbb{E}\left[L({Y}, \hat{C}(X)) \right]
\leq
\alpha - \frac{1}{2}q\text{Var}\left[\varepsilon  \right] 
\end{split}
\end{equation}
where $\alpha := \mathbb{E}\left[L({Y}, \hat{C}(X)) \right]$, $q:=\mathbb{E}_X[q_X]$, and $Q:=\mathbb{E}_X[Q_X]$.
Additionally, if $L(y, C(x))$ is convex for all $x\in{\mathcal{X}}$, then $q_x\geq 0$, and we get conservative coverage:
\begin{equation}
\begin{split} 
\mathbb{E}\left[L({Y}, \hat{C}(X)) \right] \leq \mathbb{E}\left[L(\Tilde{Y}, \hat{C}(X)) \right] = \alpha.
\end{split}
\end{equation}
\end{proof}

\subsubsection{Deriving a Miscoverage Bound from the General Risk Bound in Regression tasks}\label{sec:regression_taylor_coverage_bounds}

In this section we prove Proposition~\ref{prop:cov_bound}
and show how to obtain tight coverage bounds.
First, we define a parameterized smoothed miscoverage loss:
\begin{equation}
    L^{sm}_{d,c}(y,[a,b]) = \frac{2}{1+e^{-d\left(\left(2\frac{y-a}{b-a}-1\right)^2\right)^c}}.
\end{equation}
Above, $c,d\in\mathbb{R}$ are parameters that affect the loss function. We first connect between the smooth miscoverage and the standard miscoverage functions:
\begin{equation}\label{eq:smooth_miscoverage_connection}
\begin{split}
L(y, [a,b]) = \mathbbm{1}\{L^{sm}_{d,c}(y,[a,b]) \geq h \}.
\end{split}
\end{equation}
We now invert the above equation to find $h$:
\begin{equation}
\begin{split}
h &= L^{sm}_{d,c}\left( \min_{L(y, [a,b]) =0} {y},[a,b] \right) \\
&= L^{sm}_{d,c}\left( \max_{L(y, [a,b]) =0} {y},[a,b] \right) \\
&= L^{sm}_{d,c}\left(a, [a,b] \right) \\
&= L^{sm}_{d,c}\left(b, [a,b] \right) \\
&= \frac{2}{1+e^{-d\left(\left(2\frac{a-a}{b-a}-1\right)^2\right)^c}} \\
&= \frac{2}{1+e^{-d\left(\left(-1\right)^2\right)^c}} \\
&= \frac{2}{1+e^{-d}}.
\end{split}
\end{equation}
Therefore, $h(d)$ is a function that depends only on $d$. We now denote the second derivative of the smoothed loss by:
\begin{equation}
\begin{split}
q_x(c,d) = \min_{y}{ \frac{\partial^2}{\partial y^2} L^{sm}_{c,d}(y, C(x))}.
\end{split}
\end{equation}
Importantly, $q_x(c,d)$ can be empirically computed by sweeping over all $y\in\mathbb{R}$ and computing the second derivative of $L^{sm}_{c,d}$ for each of them.

We obtain an upper bound for the miscoverage of $C(x)$ by applying Markov's inequality using~\eqref{eq:smooth_miscoverage_connection}:
\begin{equation}\label{eq:miscov_markov}
\begin{split}
\mathbb{P}(Y \notin C(X) \mid X=x) = \mathbb{P}(L^{sm}_{d,c}(Y, C(X)) \geq h(d) \mid X=x) \leq \frac{\mathbb{E}\left[ L^{sm}_{c,d}({Y}, {C}(X)) \mid X=x\right]}{h(d)}.
\end{split}
\end{equation}
Next, we employ smoothed miscoverage upper bound provided by Proposition~\ref{prop:cov_bound}:
\begin{equation}\label{eq:smooth_miscoverage_upper_bound}
\begin{split}
 \mathbb{E}\left[ L^{sm}_{c,d}({Y}, {C}(X)) \mid X=x\right]
\leq
\mathbb{E}\left[ L^{sm}_{c,d}(\tilde{Y}, {C}(X)) \mid X=x\right]- \frac{1}{2}q_x(c,d)\text{Var}\left[\varepsilon  \right] 
\end{split}
\end{equation}
Finally, we combine~\eqref{eq:miscov_markov} and~\eqref{eq:smooth_miscoverage_upper_bound} and derive the following miscoverage bound:
\begin{equation}
\begin{split}
\mathbb{P}(Y \notin C(X) \mid X=x) \leq \frac{\mathbb{E}\left[ L^{sm}_{c,d}(\tilde{Y}, {C}(X)) \mid X=x\right] - \frac{1}{2}q_x(c,d)\text{Var}\left[\varepsilon  \right] }{h(d)},
\end{split}
\end{equation}
which can be restated to:
\begin{equation}
\begin{split}
\mathbb{P}(Y \in C(X) \mid X=x) \geq 1-\frac{\mathbb{E}\left[ L^{sm}_{c,d}(\tilde{Y}, {C}(X)) \mid X=x\right] - \frac{1}{2}q_x(c,d)\text{Var}\left[\varepsilon  \right] }{h(d)}.
\end{split}
\end{equation}
Finally, we take the expectation over all $X$ to obtain marginal coverage bound:
\begin{equation}\label{eq:coverage_taylor_bound}
\begin{split}
\mathbb{P}(Y \in C(X)) \geq 1-\frac{\mathbb{E}\left[ L^{sm}_{c,d}(\tilde{Y}, {C}(X))\right] - \frac{1}{2}\mathbb{E}_{X}[q_x(c,d)]\text{Var}\left[\varepsilon  \right] }{h(d)}.
\end{split}
\end{equation}
Crucially, all variables in \eqref{eq:coverage_taylor_bound} are empirically computable so the above lower bound can is known in practice. Additionally, the parameters $c,d$ can be tuned over a validation set to obtain tighter bounds.

\subsubsection{Deriving a Miscoverage Bound from the Density's Smoothness}\label{sec:smoothnes_regression_miscoverage_bound}
We say that the PDF of $Y\mid X=x$ is $K_x$ Lipschitz if
\begin{equation}
|f_{Y\mid X=x}(y+u) - f_{Y\mid X=x}(y)| \leq K_x u,
\end{equation}
where $f_{Y\mid X=x}$ is the PDF of $Y\mid X=x$ and $K_x \in \mathbb{R}$ is a constant that depends only on $x$.
\begin{proposition}
Suppose that $C(x)$ is a prediction interval.
Under the additive noise model $g^\text{add}$ from \eqref{eq:reg_g_sym}, if the PDF of $Y\mid X=x$ is $K_x$ Lipschitz then:
\begin{equation}
\mathbb{P}(Y \in C(X)) \geq \mathbb{P}(\tilde{Y} \in C(X)) - \mathbb{E} \left[|C(X)|K_X\right] \mathbb{E} \left[|Z |\right].
\end{equation}
\end{proposition}
\begin{proof}
First, by the definition of the noise model we get that:
\begin{equation}
\begin{split}
f_{\tilde{Y} \mid X=x}(y) &=  \int_{-\infty}^{\infty} f_{Y\mid X=x}(y-z)f_Z(z) \,dz  \\
& \leq \int_{-\infty}^{\infty} (f_{Y\mid X=x}(y) + K_x |z|) f_Z(z) \,dz
\\ &= f_{Y\mid X=x}(y) \int_{-\infty}^{\infty} f_Z(z) \,dz + K_x \int_{-\infty}^{\infty}  |z| f_Z(z) \,dz \\
&= f_{Y\mid X=x}(y)  + K_x \mathbb{E}[|Z|].
\end{split}
\end{equation}
Therefore:
\begin{equation}
\begin{split}
\mathbb{P}(Y \in C(X) \mid X=x) &= \int_{y \in C(x)} f_{Y\mid X=x}(y) \, dy\\
&\geq \int_{y \in C(x)} ( f_{\tilde{Y}\mid X=x}(y) - K_x \mathbb{E}[|Z|]) \, dy\\
&= \mathbb{P}(\tilde{Y} \in C(X) \mid X=x) - |C(x)|K_x \mathbb{E}[|Z|].
\end{split}
\end{equation}
We marginalize the above to obtain the marginal coverage bound:
\begin{equation}
\begin{split}
\mathbb{P}(Y \in C(X)) = \mathbb{P}(\tilde{Y} \in C(X)) - \mathbb{E}[|C(X)|K_X] \mathbb{E}[|Z|].
\end{split}
\end{equation}
\end{proof}

\subsubsection{The Image Miscoverage Loss}\label{sec:image_miscoverage}

In this section, we analyze the setting where the response variable is a matrix $\mathcal{Y}=\mathbb{R}^{W \times H}$. Here, the uncertainty is represented by a prediction interval $C^{i,j}(X)$ for each pixel $i,j$ in the response image. Here, our goal is to control the \emph{image miscoverage} loss~\eqref{eq:image_miscoverage_loss}, defined as:
\begin{equation}
L^{\text{im}}(Y,C(X)) = 
\frac{1}{WH}\sum_{i=1}^W {\sum_{j=1}^H{1\{Y^{i,j} \notin C^{i,j}(X) \}}}. 
\end{equation}
While this loss can be controlled under an i.i.d assumption by applying the methods proposed by~\cite{angelopoulos2021learn, angelopoulos2022conformal}, these techniques may produce invalid uncertainty sets in the presence of label noise. We now show that conservative image miscoverage risk is obtained under the assumptions of Theorem~\ref{thm:general_regression}. 
\begin{proposition}\label{prop:image_miscoverage}
Suppose that each element of the response matrix is corrupted according to an additive noise model $g^{\rm add}$ with a noise that has mean 0. 
Suppose that for every pixel $i,j$ of the response matrix, the prediction interval $C^{i,j}(X)$ and the conditional distribution of $Y^{i,j} \mid X=x$ satisfy the assumptions of Theorem~\ref{thm:general_regression} for all $x\in\mathcal{X}$.
Then, we obtain valid conditional image miscoverage:
\begin{equation}
    \forall x\in\mathcal{X}: \mathbb{E}[L^{\text{im}}(Y,C(X)) \mid X=x] \leq \mathbb{E}[L^{\text{im}}(\tilde{Y},C(X)) \mid X=x].
\end{equation}
\end{proposition}
\begin{proof}
For ease of notation, we suppose that $Y$ is a random vector of length $k$ and $C^{i}$ is the prediction interval constructed for the $i$-th element in $Y$.
Suppose that $x\in\mathcal{X}$.
Under the assumptions of Theorem~\ref{thm:general_regression}, we get that for all $i\in\{1,..,,k\}$:
\begin{equation}
    \mathbb{P}(Y^i \notin C^i(x) \mid X=x) \leq \mathbb{P}(\Tilde{Y}^i \notin C^i(x) \mid X=x).
\end{equation}
Therefore:
\begin{equation}
\begin{split}
\mathbb{E}[L^{\text{im}}(Y,C(X)) \mid X=x] &= 
\mathbb{E}\left[\frac{1}{k}\sum_{i=1}^k {1\{Y^i \notin C^i(X) }\} \mid X=x\right] \\
&=
\frac{1}{k}\sum_{i=1}^k {\mathbb{E}\left[1\{Y^i \notin C^i(X) \} \mid X=x\right]} \\
&= \frac{1}{k}\sum_{i=1}^k {\mathbb{P}\left[Y^i \notin C^i(X) \mid X=x\right]}\\
&\leq \frac{1}{k}\sum_{i=1}^k {\mathbb{P}\left[\tilde{Y}^i \notin C^i(X) \mid X=x\right]}\\
&= \frac{1}{k}\sum_{i=1}^k {\mathbb{P}\left[\tilde{Y}^i \notin C^i(X) \mid X=x\right]}\\
&= 
\frac{1}{k}\sum_{i=1}^k {\mathbb{E}\left[1\{\tilde{Y}^i \notin C^i(X) \} \mid X=x\right]} \\
&= \mathbb{E}\left[\frac{1}{k}\sum_{i=1}^k {1\{\tilde{Y}^i \notin C^i(X) }\} \mid X=x\right] \\
&= \mathbb{E}[L^{\text{im}}(\tilde{Y},C(X)) \mid X=x].
\end{split}
\end{equation}
\end{proof}

\subsection{Online Learning}

\subsubsection{Label-Noise In Online Learning Settings}\label{sec:online_risk}

Here we provide the proof to Proposition~\ref{prop:online}.
\begin{proof}[Proof of Proposition~\ref{prop:online}]
    
Suppose that for every $t\in\mathbb{N}$:
\begin{equation}
\begin{split} 
\mathbb{E}_{Y_t\mid X_t=x}\left[L_t({Y}_t, \hat{C}_t(X_t)) \mid X_t=x\right] \leq \alpha_t.
\end{split}
\end{equation}

Draw $T$ uniformly from $[0,1,...,\infty]$. Then, from the law of total expectation, it follows that:
\begin{equation}
\begin{split} 
\mathbb{E}_T\left[L_T({Y}_T, \hat{C}_T(X_T))\right] 
&= 
\mathbb{E}_{T}\left[\mathbb{E}_{X_T \mid T=t }\left[\mathbb{E}_{Y_T\mid X_T=x}\left[L_T({Y}_T, \hat{C}_T(X_T)) \mid X_T=x\right] \mid T=t \right]  \right] \\
&\leq 
\mathbb{E}_{T}\left[\mathbb{E}_{X_T \mid T=t }\left[\alpha_T \mid T=t \right] \right] \\
& = \mathbb{E}_{T}\left[\alpha_T \right] \\
&= \alpha,
\end{split}
\end{equation}
where $\alpha := \mathbb{E}_{T}\left[\alpha_T \right]$. Finally, we get:
\begin{equation}
\begin{split} 
\lim_{T \rightarrow \infty} {\frac{1}{T}\sum_{t=0}^T {L_t(Y_t. \hat{C}_t(X_t))}} =
\mathbb{E}_T\left[L_T({Y}_T, \hat{C}_T(X_T))\right] 
\leq \alpha
\end{split}
\end{equation}
\end{proof}

\subsubsection{Label-Noise Robustness with the Miscoverage Counter Loss}\label{sec:miscoverage_counter}

In this section, we suppose an online learning setting, where the data $\{(x_t,y_t)\}_{t=1}^{\infty}$ is given as a stream. 
The miscoverage counter loss \citep{feldman2022achieving} counts the number of consecutive miscoverage events that occurred until the timestamp $t$. Formally, given a series of prediction sets: $\{\hat{C}_t(x_t)\}_{t=1}^T$, and a series of labels: $\{y_t\}_{t=1}^T$, the miscoverage counter at timestamp $t$ is defined as:
\begin{equation}\label{eq:MC}
    L^{\text{MC}}_t(y_t,\hat{C}_t(x_t)) = \begin{cases}
      L^{\text{MC}}_{t-1}(y_{t-1},\hat{C}_{t-1}(x_{t-1})) + 1, & y_t \not \in \hat{C}_t(x_t) \\
      0, & \text{otherwise.}
    \end{cases}
\end{equation}
We now show that conservative miscoverage counter risk is obtained under the presence of label noise. 

\begin{proposition}\label{prop:miscoverage_counter}
Suppose that for all $0<k\in\mathbb{N}$, $1<t\in\mathbb{N}$, and $x_t \in \mathcal{X}$:
\begin{equation}
\begin{split}
\mathbb{P}(Y_t \in \Chatnoisyt(X_t) &\mid X_t=x_t, L^{\text{MC}}_{t-1}(Y_{t-1},\hat{C}_{t-1}(X_{t-1})) = k) \geq \\
&\mathbb{P}(\tilde{Y}_t \in \Chatnoisyt(X_t) \mid X_{t}=x_t, L^{\text{MC}}_{t-1}(\tilde{Y}_{t-1},\hat{C}_{t-1}(X_{t-1})) = k).
\end{split}
\end{equation}
Also assume that:
\begin{equation}
\mathbb{P}(Y_1 \in \Chat_{\rm noisy}^{1}(X_1) \mid X_1=x_1) \geq 
\mathbb{P}(\tilde{Y}_1 \in \Chat_{\rm noisy}^{1}(X_1) \mid X_{1}=x_1).
\end{equation}
If the miscoverage counter risk of the noisy labels is controlled at level $\alpha$, then the miscoverage counter of the clean labels is controlled at level $\alpha$:
\begin{equation}
\lim_{T \rightarrow \infty} \frac{1}{T} \sum_{t=1}^{T} \mathbb{E}[L^{\text{MC}}_{t}({Y}_t,\hat{C}_t(X_t))]\leq \alpha. 
\end{equation}
\end{proposition}
Notice that the conditions of Proposition~\ref{prop:miscoverage_counter} follow from Theorem~\ref{thm:general_class} in classification tasks or from Theorem~\ref{thm:general_regression} in regression tasks. In other words, we are guaranteed to obtain valid miscoverage counter risk when the requirements of these theorems are satisfied. We now demonstrate this result in a regression setting.
\begin{corollary}
Suppose that the noise model and the conditional distributions of the clean $Y_t \mid X_t,X_{t-1},Y_{t-1}$ and noisy $\tilde{Y}_t \mid X_t,X_{t-1},\tilde{Y}_{t-1}$ labels satisfy the assumptions of Theorem~\ref{thm:general_regression} for all $t\in\mathbb{N}$. Then, the miscoverage counter of the clean labels is more conservative than the risk of the noisy labels.
\end{corollary}
We now turn to prove Proposition~\ref{prop:miscoverage_counter}.
\begin{proof}[Proof of Proposition~\ref{prop:miscoverage_counter} ]
Suppose that $x_t \in \mathcal{X}$. Our objective is to prove that any series of intervals $\{\hat{C}_t(X_t)\}_{t=1}^{\infty}$ satisfies:
\begin{equation}
\lim_{T \rightarrow \infty} {\frac{1}{T} \sum_{t=1}^{T} {\mathbb{E}[L^{\text{MC}}_{t}({Y}_t,\hat{C}_t(X_t))]}} 
\leq \lim_{T \rightarrow \infty} {\frac{1}{T} \sum_{t=1}^{T} {\mathbb{E}[L^{\text{MC}}_{t}(\tilde{Y}_t,\hat{C}_t(X_t))]}}
\end{equation}
First, we show by induction over $0<t,k \in \mathbb{N}$ that:
\begin{equation}
\mathbb{P}[L^{\text{MC}}_{t}(Y_t,\hat{C}_t(X_t))= k ] \leq \mathbb{P}[L^{\text{MC}}_{t}(\tilde{Y}_t,\hat{C}_t(X_t))= k ].
\end{equation}
Base: for $k=1$ and $t=1$:
\begin{equation}
\begin{split}
\mathbb{P}[L^{\text{MC}}_{t}(Y_t,\hat{C}_t(X_t))= k \mid X_t =x_t] &= 
\mathbb{P}[Y_t \notin ,\hat{C}_t(X_t) \mid X_t =x_t] \\
&\leq \mathbb{P}[\tilde{Y}_t \notin ,\hat{C}_t(X_t) \mid X_t =x_t]\\
&= \mathbb{P}[L^{\text{MC}}_{t}(\tilde{Y}_t,\hat{C}_t(X_t))= k \mid X_t =x_t].
\end{split}
\end{equation}
Inductive step: suppose that the statement is correct for $t,k$. We now show for $k+1$:
\begin{equation}
\begin{split}
&\mathbb{P}[L^{\text{MC}}_{t}(Y_t,\hat{C}_t(X_t))= k+1 ] \\
&=
\mathbb{P}[L^{\text{MC}}_{t}(Y_t,\hat{C}_t(X_t))= k+1 \mid L^{\text{MC}}_{t-1}(Y_{t-1},\hat{C}_{t-1}(X_{t-1}))= k]\mathbb{P}[L^{\text{MC}}_{t-1}(Y_{t-1},\hat{C}_{t-1}(X_{t-1}))= k]\\
&=
\mathbb{P}[Y_t \notin \hat{C}_t(X_t) \mid L^{\text{MC}}_{t-1}(Y_{t-1},\hat{C}_{t-1}(X_{t-1}))= k]\mathbb{P}[L^{\text{MC}}_{t-1}(Y_{t-1},\hat{C}_{t-1}(X_{t-1}))= k]\\
&\leq
\mathbb{P}[\tilde{Y}_t \notin \hat{C}_t(X_t) \mid L^{\text{MC}}_{t-1}(\tilde{Y}_{t-1},\hat{C}_{t-1}(X_{t-1}))= k]\mathbb{P}[L^{\text{MC}}_{t-1}(\tilde{Y}_{t-1},\hat{C}_{t-1}(X_{t-1}))= k]\\
&=
\mathbb{P}[L^{\text{MC}}_{t}(\tilde{Y}_{t},\hat{C}_{t}(X_{t}))= k+1].
\end{split}
\end{equation}
Inductive step 2: suppose that the statement is correct for $t,k$. We now show for $t+1$:
\begin{equation}
\begin{split}
&\mathbb{P}[L^{\text{MC}}_{t+1}(Y_{t+1},\hat{C}_{t+1}(X_{t+1}))= k ] \\
&= \mathbb{P}[L^{\text{MC}}_{t+1}(Y_{t+1},\hat{C}_{t+1}(X_{t+1}))= k \mid L^{\text{MC}}_{t}(Y_{t},\hat{C}_{t}(X_{t}))= k-1 ] \mathbb{P}[L^{\text{MC}}_{t}(Y_{t},\hat{C}_{t}(X_{t}))= k-1]\\
&= \mathbb{P}[Y_{t+1} \notin \hat{C}_{t+1}(X_{t+1}) \mid L^{\text{MC}}_{t}(Y_{t},\hat{C}_{t}(X_{t}))= k-1 ] \mathbb{P}[L^{\text{MC}}_{t}(Y_{t},\hat{C}_{t}(X_{t}))= k-1]\\
&\leq \mathbb{P}[\tilde{Y}_{t+1} \notin \hat{C}_{t+1}(X_{t+1}) \mid L^{\text{MC}}_{t}(\tilde{Y}_{t},\hat{C}_{t}(X_{t}))= k-1 ] \mathbb{P}[L^{\text{MC}}_{t}(\tilde{Y}_{t},\hat{C}_{t}(X_{t}))= k-1]\\
&=\mathbb{P}[L^{\text{MC}}_{t+1}(\tilde{Y}_{t+1},\hat{C}_{t+1}(X_{t+1}))= k ]
\end{split}
\end{equation}

Finally, we compute the miscoverage counter risk over the time horizon:
\begin{equation}
\begin{split}
\lim_{T \rightarrow \infty} {\frac{1}{T} \sum_{t=1}^{T} {\mathbb{E}[L^{\text{MC}}({Y}_t,\hat{C}_t(X_t))]}}  &= 
\lim_{T \rightarrow \infty} {\frac{1}{T} \sum_{t=1}^{T} {\sum_{k=1}^{\infty} {k \mathbb{P}[L^{\text{MC}}_{t}({Y}_t,\hat{C}_t(X_t))= k ]}}} \\
&\leq \lim_{T \rightarrow \infty} {\frac{1}{T} \sum_{t=1}^{T} {\sum_{k=1}^{\infty} {k \mathbb{P}[L^{\text{MC}}_{t}(\tilde{Y}_t,\hat{C}_t(X_t))= k] }}} \\
&= \lim_{T \rightarrow \infty} {\frac{1}{T} \sum_{t=1}^{T} {\mathbb{E}[L^{\text{MC}}_{t}(\tilde{Y}_t,\hat{C}_t(X_t))]}}\\
& \leq \alpha.
\end{split}
\end{equation}

\end{proof}

\section{Additional Experimental Details and Results}

\subsection{Classification: Object Recognition Experiment} \label{app:cifar-10-supp}

\begin{figure}[t]
\centering
\includegraphics[width=.5\linewidth]{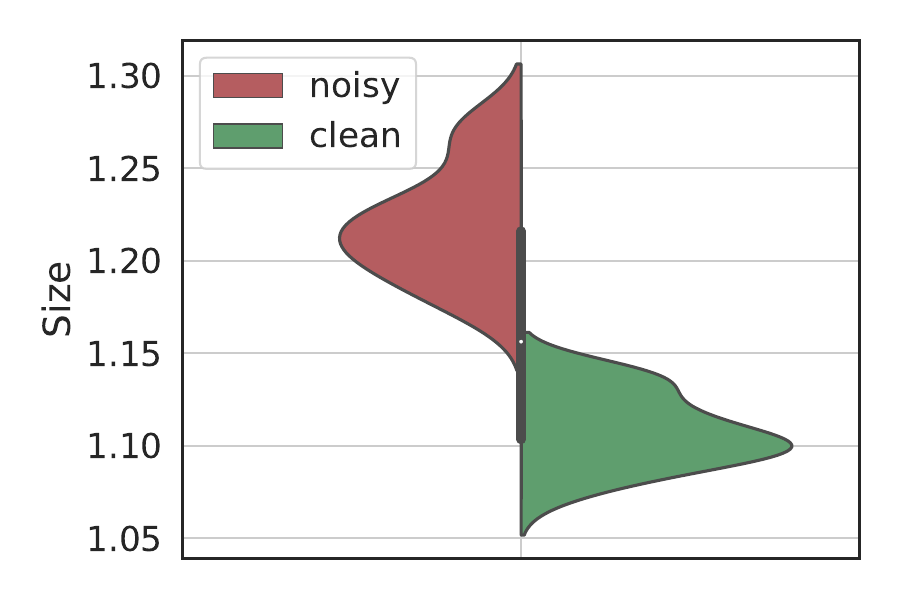}
\caption{\textbf{Effect of label noise on CIFAR-10}. Distribution of average prediction set sizes over 30 independent experiments evaluated on CIFAR-10H test data using noisy and clean labels for calibration. Other details are as in Figure~\ref{fig:cifar10-clean}}
\label{fig:sets_cifar10-clean}
\end{figure}

Here we present additional results of the classification experiment with CIFAR-10H explained in Section~\ref{sec:real_class}, but first provide further details about the data set and training procedure.
The CIFAR-10H data set contains the same 10,000 images as CIFAR-10, but with labels from a single annotator instead of a majority vote of 50 annotators. We fine-tune a ResNet18 model pre-trained on the clean training set of CIFAR-10, which contains 50,000 samples. Then we randomly select 2,000 observations from CIFAR-10H for calibration. The test set contains the remaining 8,000 samples, but with CIFAR-10 labels.
We apply conformal prediction with the APS score. The marginal coverage achieved when using noisy and clean calibration sets are depicted in Figure~\ref{fig:cifar10-clean}. This figure shows that (i) we obtain the exact desired coverage when using the clean calibration set; and (ii) when calibrating on noisy data, the constructed prediction sets over-cover the clean test labels. Figure~\ref{fig:sets_cifar10-clean} illustrates the average prediction set sizes that are larger when using noisy data for calibration and thus lead to higher coverage levels.

\subsection{Regression: Aesthetic Visual Rating} \label{app:AVA-supp}

Herein, we provide additional details regarding the training of the predictive models for the real-world regression task. As explained in Section~\ref{sec:real_reg}, we use a VGG-16 model---pre-trained on the ImageNet data set---whose last (deepest) fully connected layer is removed. Then, we feed the output of the VGG-16 model to a linear fully connected layer to predict the response. We train two different models: a quantile regression model for CQR and a classic regression model for conformal with residual magnitude score. Both models are trained on $34,000$ noisy samples, calibrated on $7,778$ noisy holdout points, and tested on $7,778$ clean samples. We train the quantile regression model for 70 epochs using 'SGD' optimizer with a batch size of 128 and an initial learning rate of 0.001 decayed every 20 epochs exponentially with a rate of 0.95 and a frequency of 10. We apply dropout regularization to avoid overfitting with a rate of 0.2.
We train the classic regression model for 70 epochs using 'Adam' optimizer with a batch size of 128 and an initial learning rate of 0.00005 decayed every 10 epochs exponentially with a rate of 0.95 and a frequency of 10. The dropout rate in this case is 0.5.

\subsection{Synthetic Classification: Adversarial Noise Models}\label{app:adv-supp}

In contrast with the noise distributions presented in Section~\ref{sec:syn_classification}, here we construct adversarial noise models to intentionally reduce the coverage rate.

\begin{figure}[t]
    {\includegraphics[width=.60\linewidth]{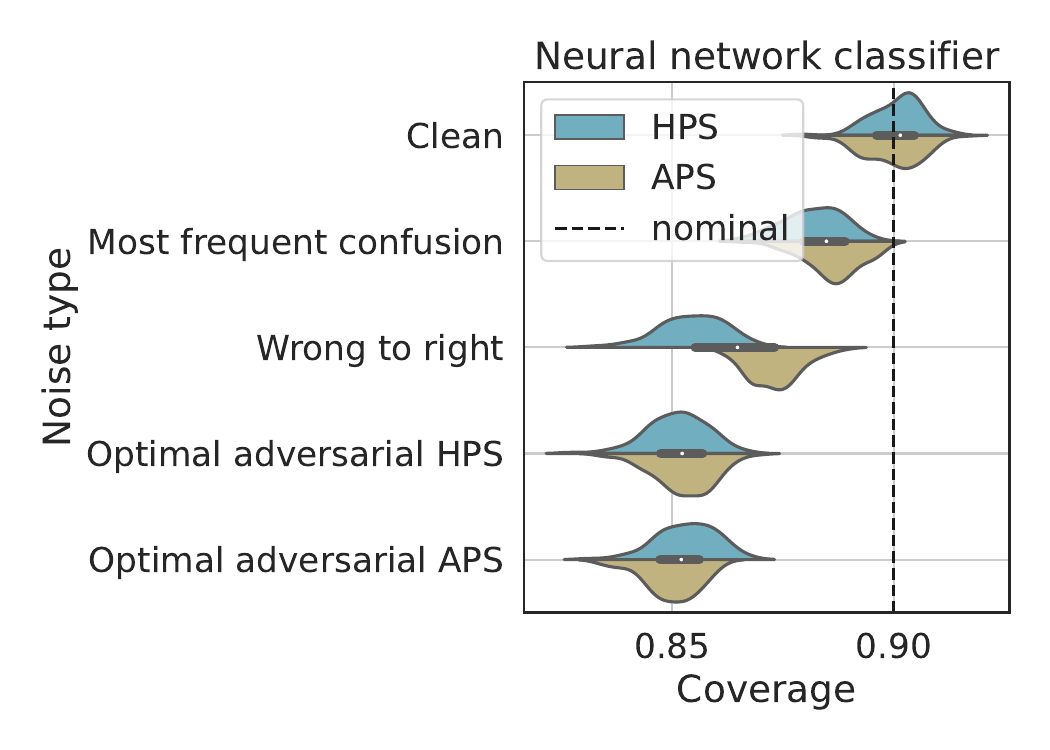}}\hfill
    {\includegraphics[width=.35\linewidth]{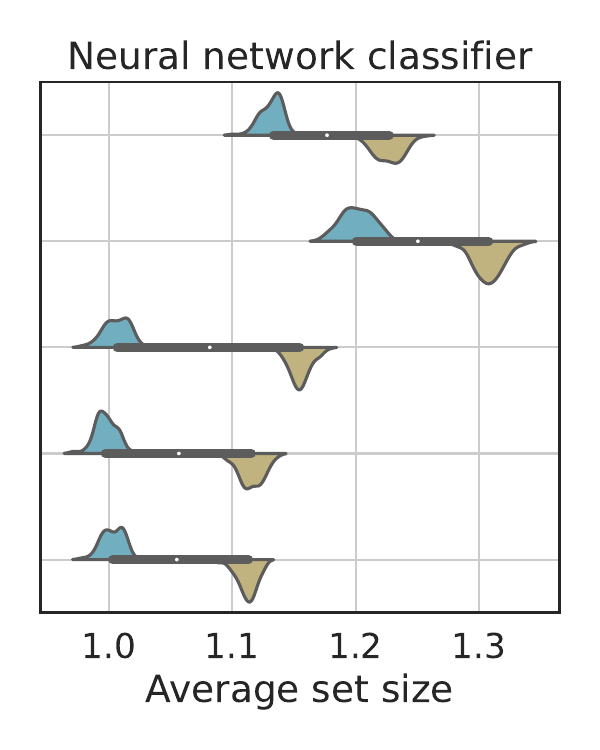}}\par 
    \caption{\textbf{Effect of label noise on synthetic multi-class classification data}. Performance of conformal prediction sets with target coverage $1-\alpha = 90\% $, using a noisy training set and a noisy calibration set with adversarial noise models. Left: Marginal coverage; Right: Average size of predicted sets. The results are evaluated over 100 independent experiments.}
    \label{fig:syntethic_classification_adv}
\end{figure}

\begin{enumerate}
\item \texttt{Most frequent confusion}:
we extract from the confusion matrix the pair of classes with the highest probability to be confused between each other, and switch their labels until reaching a total probability of $\epsilon$. In cases where switching between the most common pair is not enough to reach $\epsilon$, we proceed by flipping the labels of the second most confused pairs of labels, and so on.
\item \texttt{Wrong to right}:
wrong predictions during calibration cause larger prediction sets during test time. Hence making the model think it makes fewer mistakes than it actually does during calibration can lead to under-coverage during test time. Here, we first observe the model predictions over the calibration set, and then switch the labels only of points that were misclassified. We switch the label to the class that is most likely to be the correct class according to the model, hence making the model think it was correct. We switch a suitable amount of labels in order to reach a total switching probability of $\epsilon$ (this noise model assumes there are enough wrong predictions in order to do so).
\item \texttt{Optimal adversarial}:
we describe here an algorithm for building the worst possible label noise for a specific model using a specific non-conformity score. This noise will decrease the calibration threshold at most and as a result, will cause significant under-coverage during test time. To do this, we perform an iterative process. In each iteration, we calculate the non-conformity scores of all of the calibration points with their current labels. We calculate the calibration threshold as in regular conformal prediction and then, from the points that have a score above the threshold, we search for the one that switching its label can reduce its score by most. We switch the label of this point to the label that gives the lowest score and then repeat the iterative process with the new set of labels. Basically, at every step, we make the label swap that will decrease the threshold by most.

\end{enumerate}        
In these experiments, we apply the same settings as described in Section~\ref{sec:syn_classification} and present the results in Figure~\ref{fig:syntethic_classification_adv}. We can see that the \texttt{optimal adversarial} noise causes the largest decrease in coverage as one would expect. 
The \texttt{most-frequent-confusion} noise decreases the neural network coverage to approximately $89\%$. The \texttt{wrong-to-right} noise decreases the coverage to around $85\%$ with the HPS score and to around $87\%$ with the APS score. This gap is expected as this noise directly reduces the HPS score. We can see that the optimal worst-case noise for each score function reduces the coverage to around $85\%$ when using that score. This is in fact the maximal decrease in coverage possible theoretically, hence it strengthens the optimally of our iterative algorithm.

\subsection{Synthetic Regression: Additional Results} \label{app:synthetic-reg-supp}
 
\begin{figure}[ht]
\subfloat[]{\includegraphics[width=.45\linewidth]{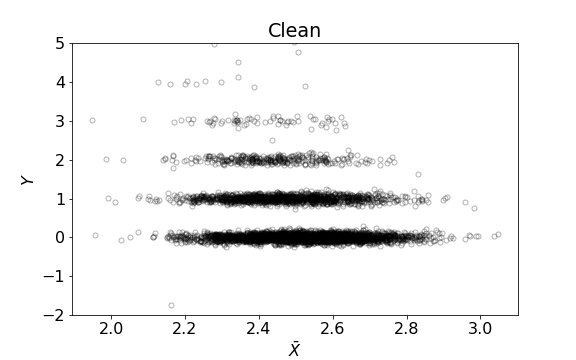}}\hfill
\subfloat[]{\includegraphics[width=.45\linewidth]{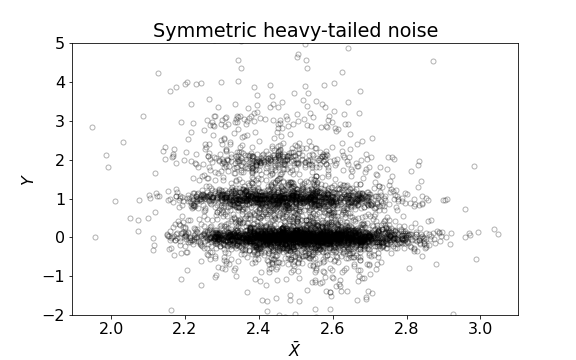}}\par 
\subfloat[]{\includegraphics[width=.45\linewidth]{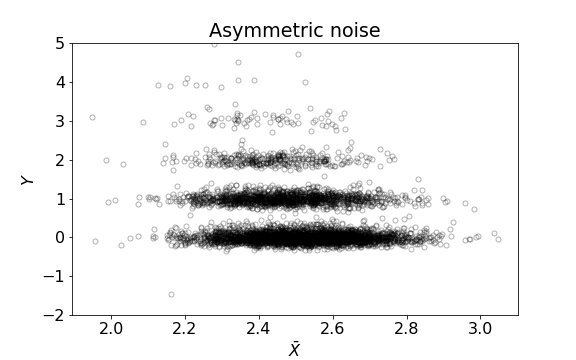}}\hfill
\subfloat[]{\includegraphics[width=.45\linewidth]{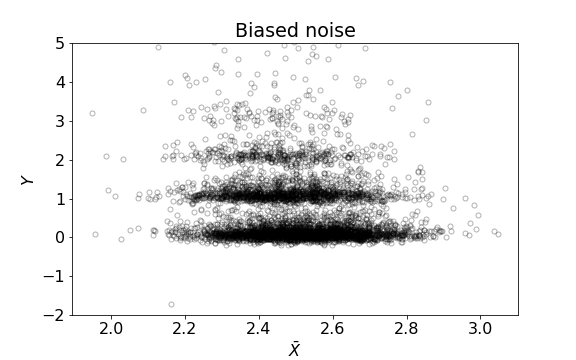}}\par
\subfloat[]{\includegraphics[width=.45\linewidth]{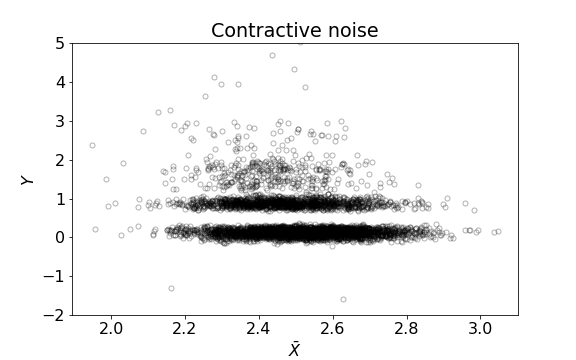}}\hfill
\subfloat[]{\includegraphics[width=.45\linewidth]{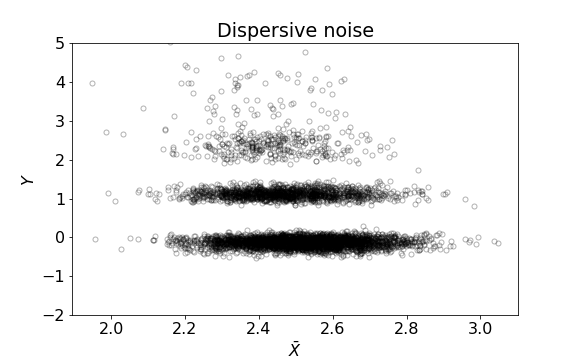}}
\caption{\textbf{Illustration of the generated data with different corruptions.} (a): Clean samples. (b): Samples with symmetric heavy-tailed noise. (c): Samples with asymmetric noise. (d): Samples with biased noise. Noise magnitude is set to 0.1. (e): Samples with contractive noise. (f): Samples with dispersive noise.}
\label{fig:data-example}
\end{figure}

Here we first illustrate in Figure~\ref{fig:data-example} the data we generate in the synthetic regression experiment from Section~\ref{sec:synthetic-reg} and the different corruptions we apply.

In Section~\ref{sec:synthetic-reg} we apply some realistic noise models and examine the performance of conformal prediction using CQR score with noisy training and calibration sets. Here we construct some more experiments using the same settings, however we train the models using clean data instead of noisy data.
Moreover, we apply an additional adversarial noise model that differs from those presented in Section~\ref{sec:synthetic-reg} in the sense that it is designed to intentionally reduce the coverage level.
 
 \texttt{Wrong to right}: an adversarial noise that depends on the underlying trained regression model. In order to construct the noisy calibration set we switch 7\% of the responses as follows: we randomly swap between outputs that are not included in the interval predicted by the model and outputs that are included. 

\begin{figure}[h]
{\includegraphics[width=.45\linewidth]{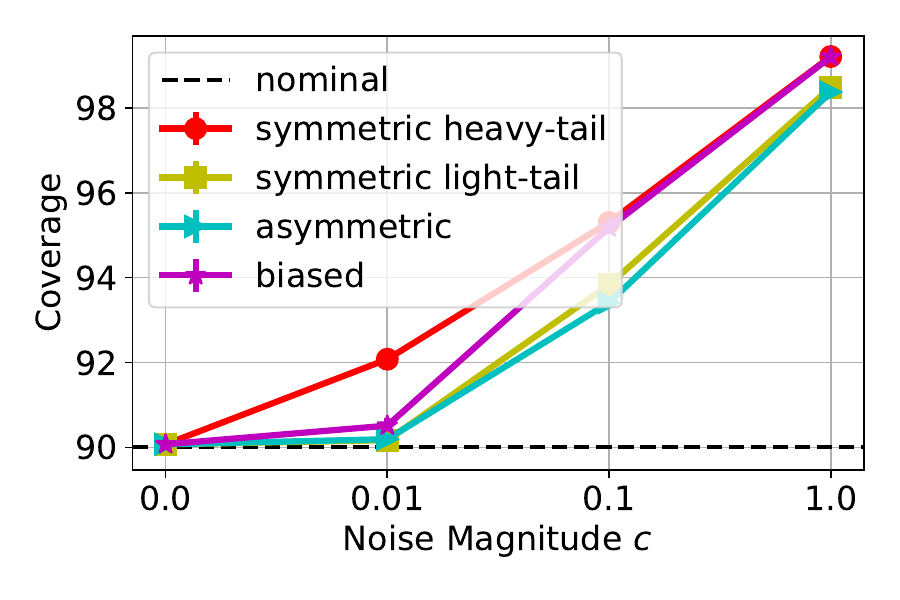}}\hfill
{\includegraphics[width=.45\linewidth]{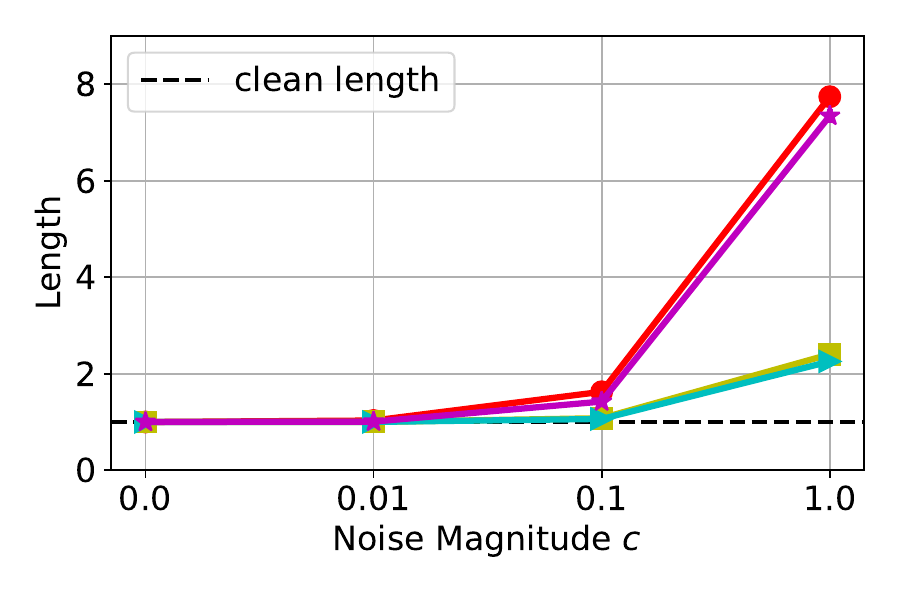}}\par 
\caption{\textbf{Response-independent noise.} Performance of conformal prediction intervals with target coverage $1-\alpha=90\%$, using a clean training set and a noisy calibration set. Left: Marginal coverage; Right: Length of predicted intervals (divided by the average clean length) using symmetric, asymmetric and biased noise with a varying magnitude. Other details are as in Figure~\ref{fig:sym_noisy_train}.}
\label{fig:noise_clean_train}
\end{figure}

\begin{figure}[h]
{\includegraphics[width=.45\linewidth]{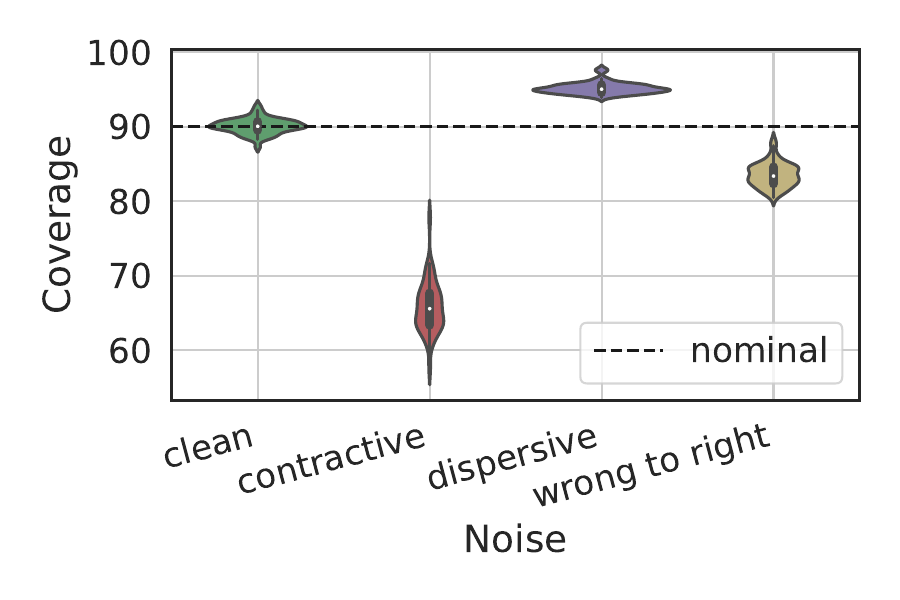}}\hfill
{\includegraphics[width=.45\linewidth]{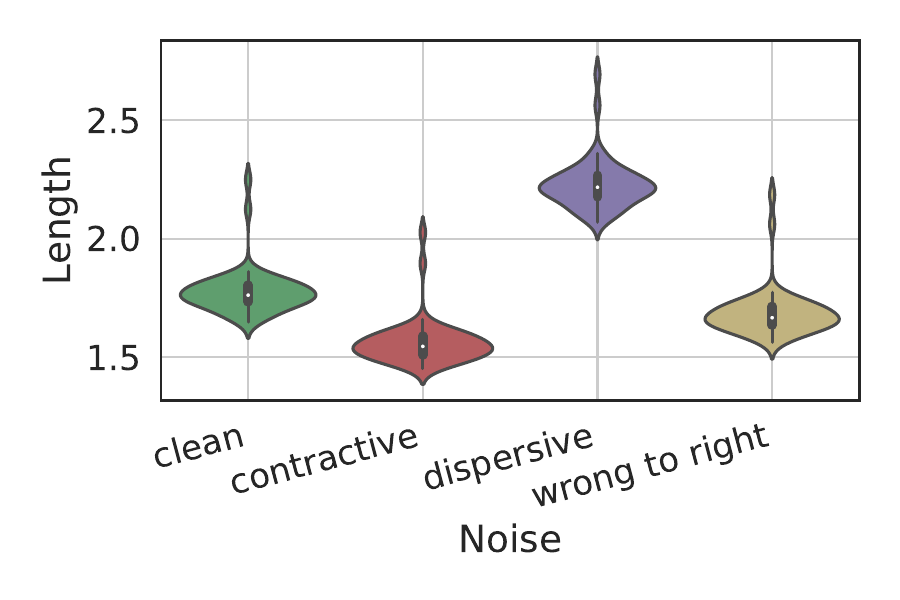}}\par 
\caption{\textbf{Response-dependent noise.} Performance of conformal prediction intervals with target coverage $1-\alpha=90\%$, using a clean training set and a noisy calibration set. Left: Marginal coverage; Right: Length of predicted intervals. The results are evaluated over 50 independent experiments.}
\label{fig:w2r_and_r2c}
\end{figure}

Figures~\ref{fig:noise_clean_train} and~\ref{fig:w2r_and_r2c} depict the marginal coverage and interval length achieved when applying the different noise models. We see that the adversarial \texttt{wrong to right} noise model reduces the coverage rate to approximately 83\%. Moreover, these results are similar to those achieved in Section~\ref{sec:synthetic-reg}, except for the conservative coverage attained using \texttt{biased} noise, which can be explained by the more accurate low and high estimated quantiles.  

\begin{figure}[ht]
\subfloat[]{\includegraphics[width=.45\linewidth]{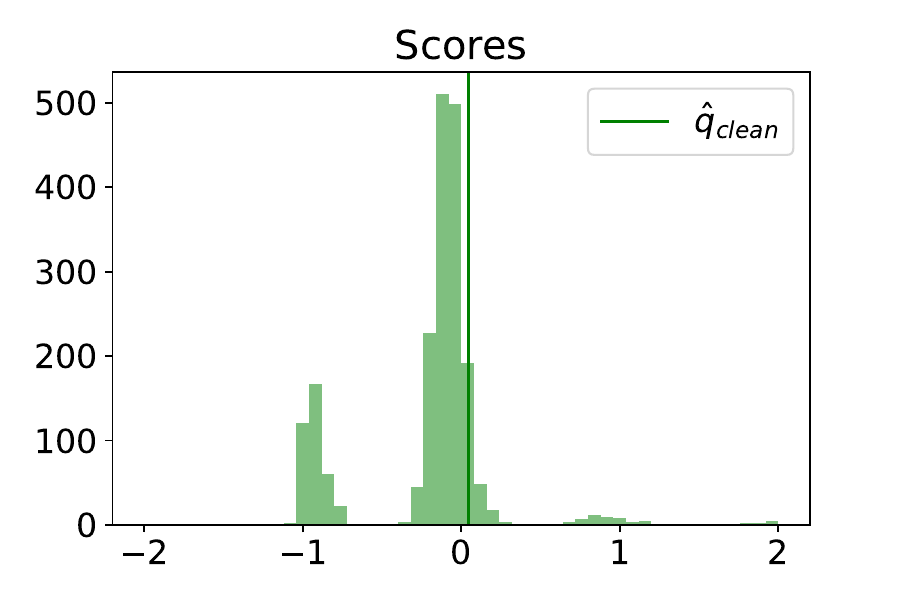}}\hfill
\subfloat[]{\includegraphics[width=.45\linewidth]{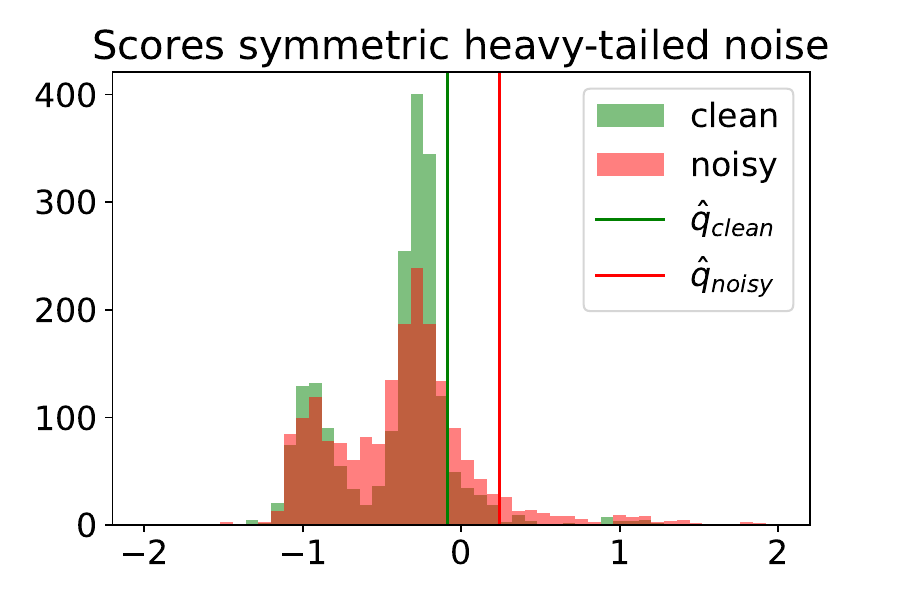}}\par 
\subfloat[]{\includegraphics[width=.45\linewidth]{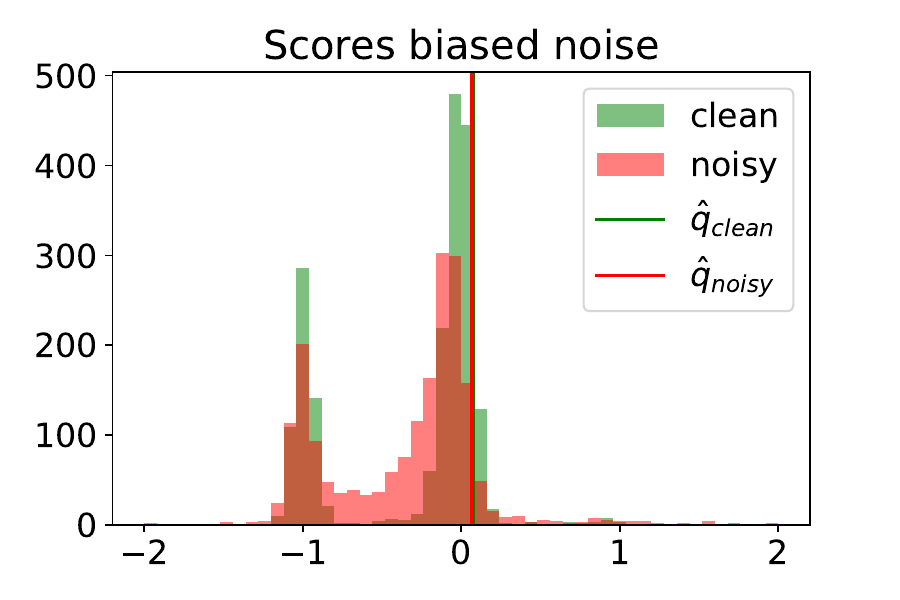}}\hfill
\subfloat[]{\includegraphics[width=.45\linewidth]{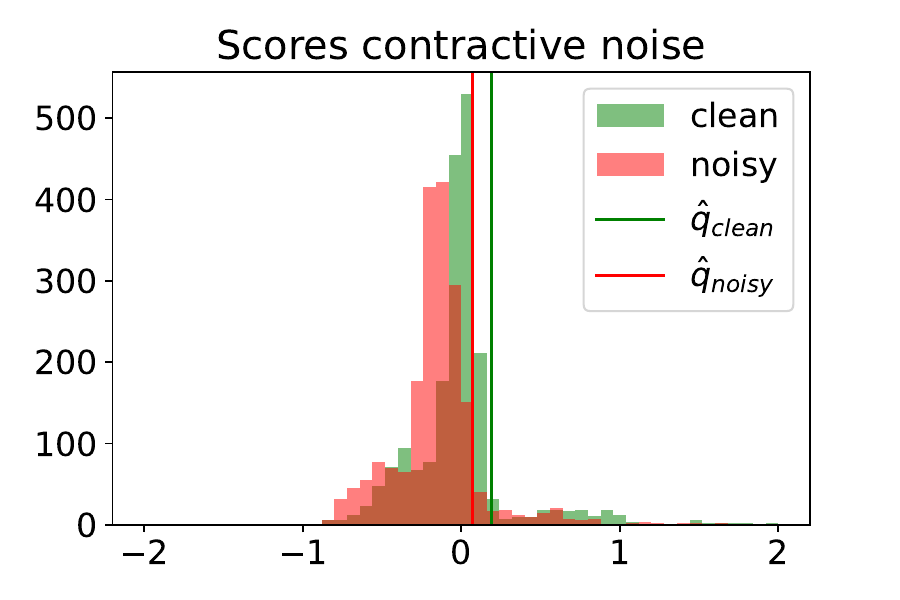}}
\caption{\textbf{Illustration of the CQR scores}. (a): Clean training and calibration sets. (b): Symmetric heavy-tailed noise. (c): Biased noise. Noise magnitude is set to 0.1. (d): Contractive noise. Other details are as in Figure~\ref{fig:sym_noisy_train}.}
\label{fig:scores-noisy_train}
\end{figure}

Lastly, in order to explain the over-coverage or under-coverage achieved for some of the different noise models, as depicted in Figures~\ref{fig:sym_noisy_train} and ~\ref{fig:center_noisy_train}, we present in Figure~\ref{fig:scores-noisy_train} the CQR scores and their 90\%'th empirical quantile. Over-coverage is achieved when the noisy scores are larger than the clean ones, for example, in the \texttt{symmetric heavy tailed} case, and under-coverage is achieved when the noisy scores are smaller.

\subsection{The Multi-label Classification Experiment with the COCO Data Set}\label{sec:real_coco_noise_details}

\begin{figure}[ht]
\centering
\includegraphics[width=.5\linewidth]{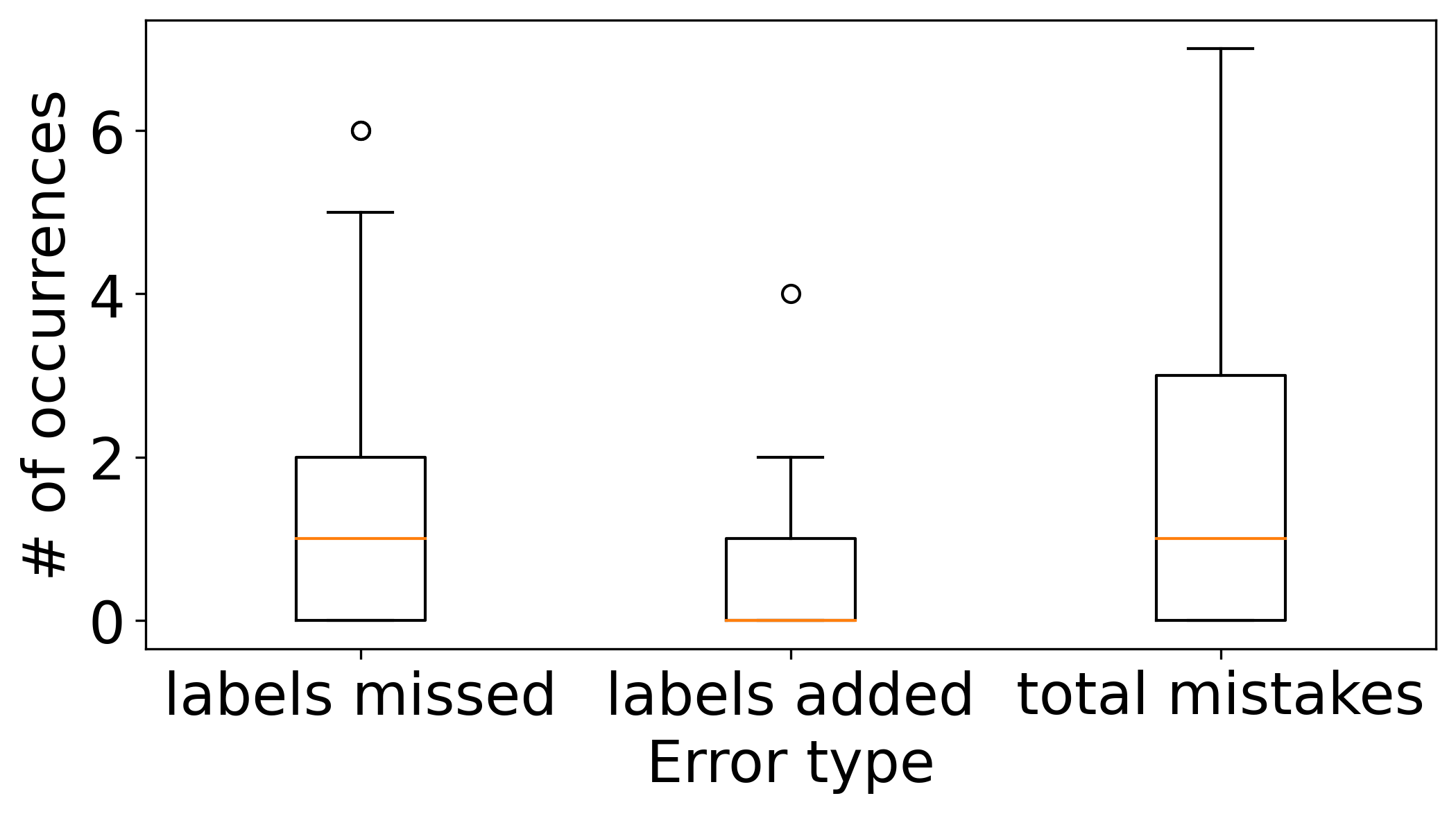}
\caption{Analysis of human-made label corruptions in MS COCO data set. }
\label{fig:coco_real_corruption_analysis}
\end{figure}

Here, we provide the full details about the experimental setup of the real multi-label corruptions from Section~\ref{sec:real_coco_noise_exp}. We asked 9 annotators to annotate 117 images. Each annotator labeled approximately 15 images separately, except for two annotators who labeled 15 images as a couple. The annotators were asked to label each image under 30 seconds, although this request was not enforced. Figure~\ref{fig:coco_real_corruption_analysis} presents the number of labels that are missed or mistakenly added to each image. On average, 1.25 of the labels were missed and 0.5 were mistakenly added, meaning that each image contains a total of 1.75 label mistakes on average.

\subsection{Multi-label Classification with Artificial Corruptions}

In this section, we study three types of corruptions that demonstrate Proposition~\ref{prop:multi_label} in the sense that the number of positive labels, $|\tilde{Y}|$, is a deterministic function of $X$.
The data set we use in the following experiments is the MS COCO data set \citep{lin2014microsoft}, in which the input image may contain up to $K=80$ positive labels. We consider the given annotations as ground-truth labels and artificially corrupt them to generate noisy ones, as conducted in~\cite{zhao2021evaluating, kumar2020robust}. 
The first noise model applies dependent corruptions by taking $\beta$ of the positive labels and turning them into different ones, according to a pre-defined transition map that was randomly chosen. The second one applies independent corruptions: $\beta$ of the positive labels turn into different ones, chosen independently and uniformly. Finally, the partial noise model corrupts the labels as the independent one, except for 30 pre-defined labels that remain unchanged. In all experiments, we set $\beta=0.45$.

\begin{figure}[ht]
\centering
\includegraphics[width=.99\linewidth]{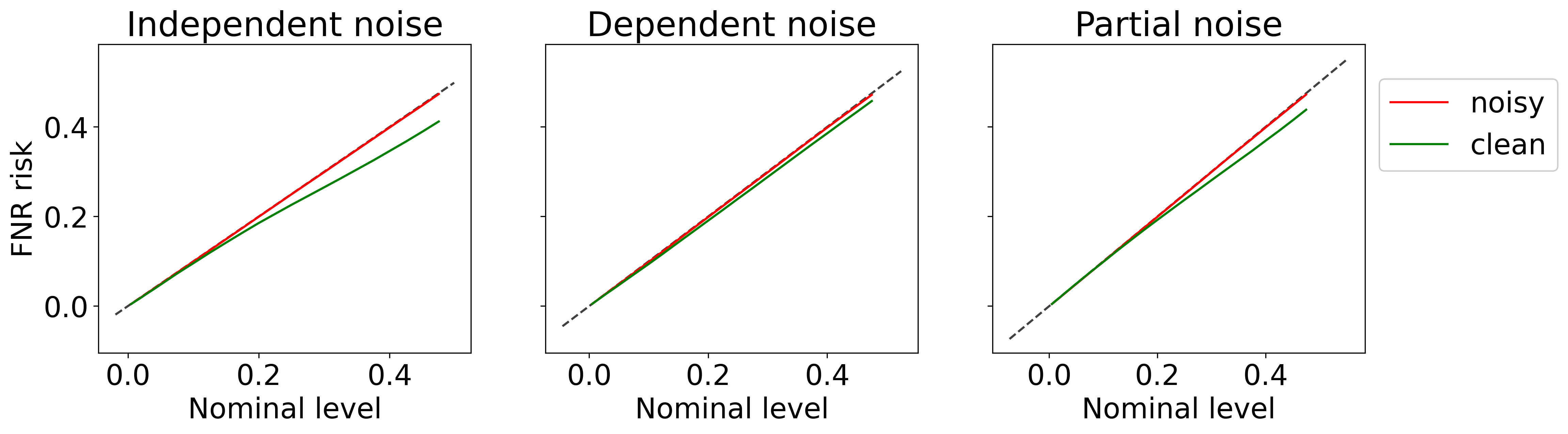}
\caption{{FNR on MS COCO data set, achieved over noisy (red) and clean (green) test sets. The calibration scheme is applied with noisy annotations. Results are averaged over 50 trials.}}
\label{fig:coco_exp}
\end{figure}

We fit a TResNet \citep{ridnik2021tresnet} model on 100k clean samples and calibrate it using 10K noisy samples with conformal risk control, as outlined in \cite[Section 3.2]{angelopoulos2022conformal}. We control the false-negative rate (FNR), defined in~\eqref{eq:fnr}, at different labels and measure the FNR obtained over clean and noisy versions of the test set which contains 30k samples. Figure~\ref{fig:coco_exp} displays the results, showing that the risk obtained over the clean labels is more conservative than the risk of the clean labels. This is not a surprise, as it is supported by our theory, in Proposition~\ref{prop:multi_label}. 

\subsection{Regression Miscoverage Bounds Experiment}\label{sec:miscoverage_bounds_exp}
\begin{figure}[ht]
\begin{center}
{\includegraphics[width=.32\linewidth]{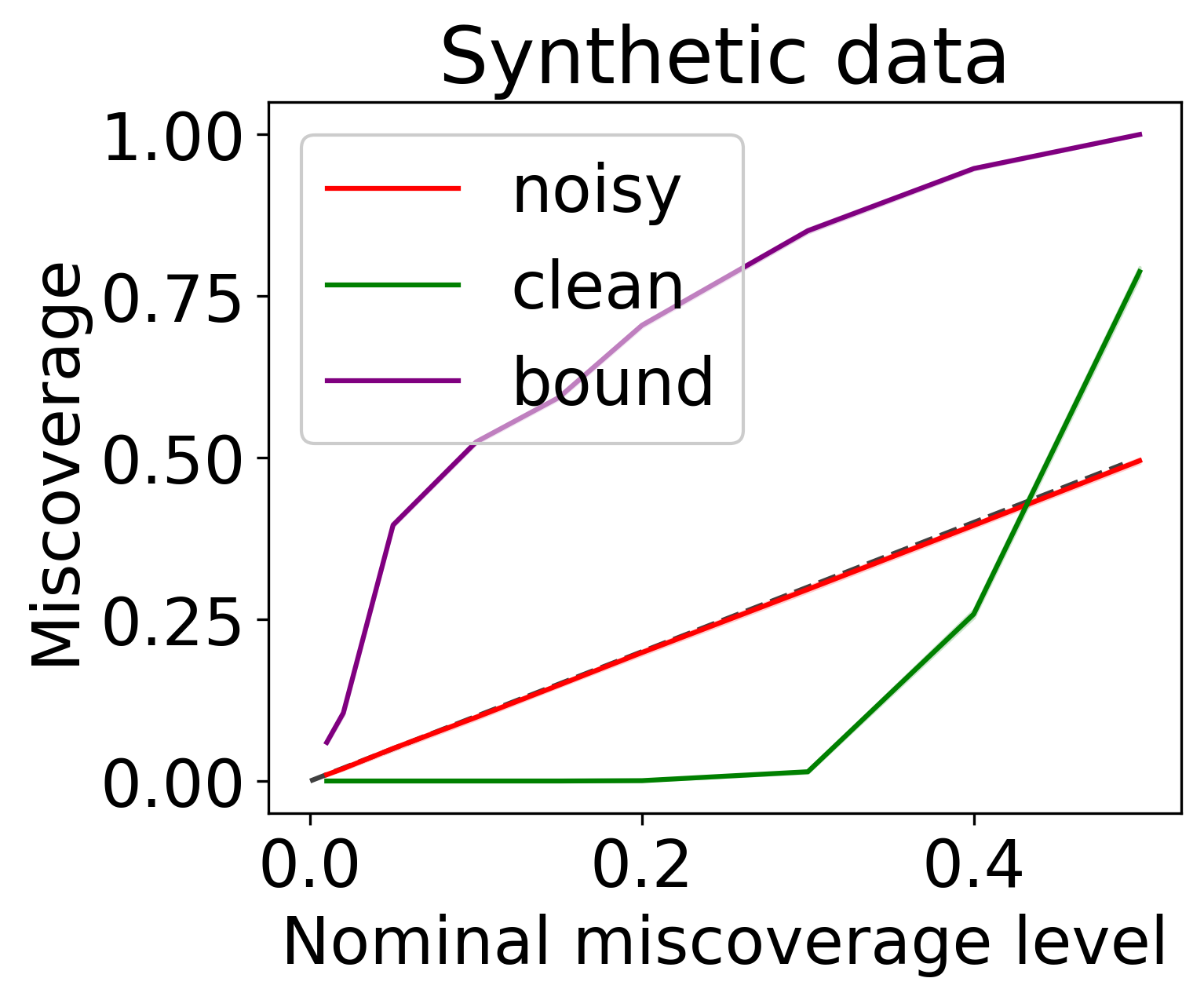}}
{\includegraphics[width=.32\linewidth]{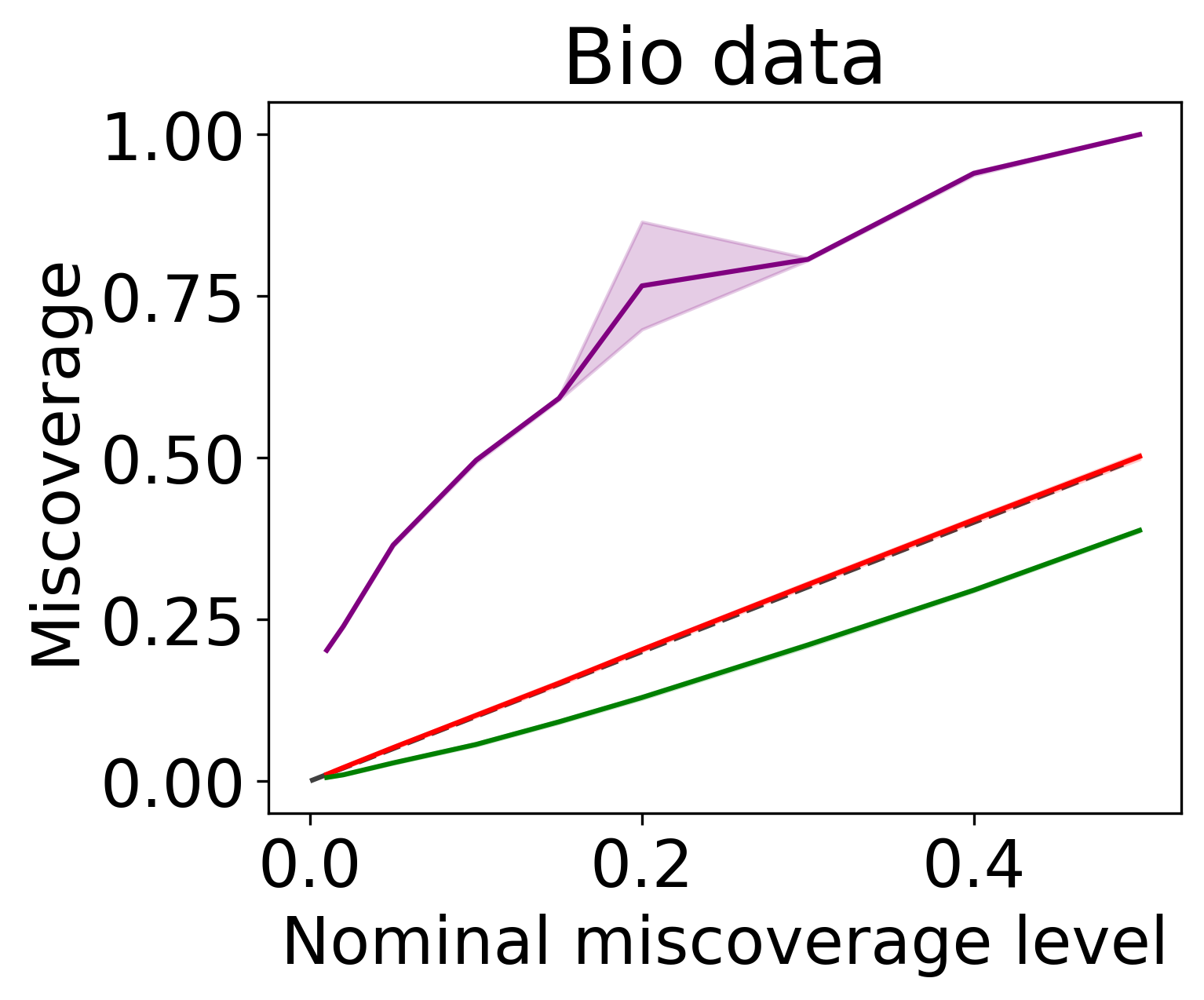}}
{\includegraphics[width=.32\linewidth]{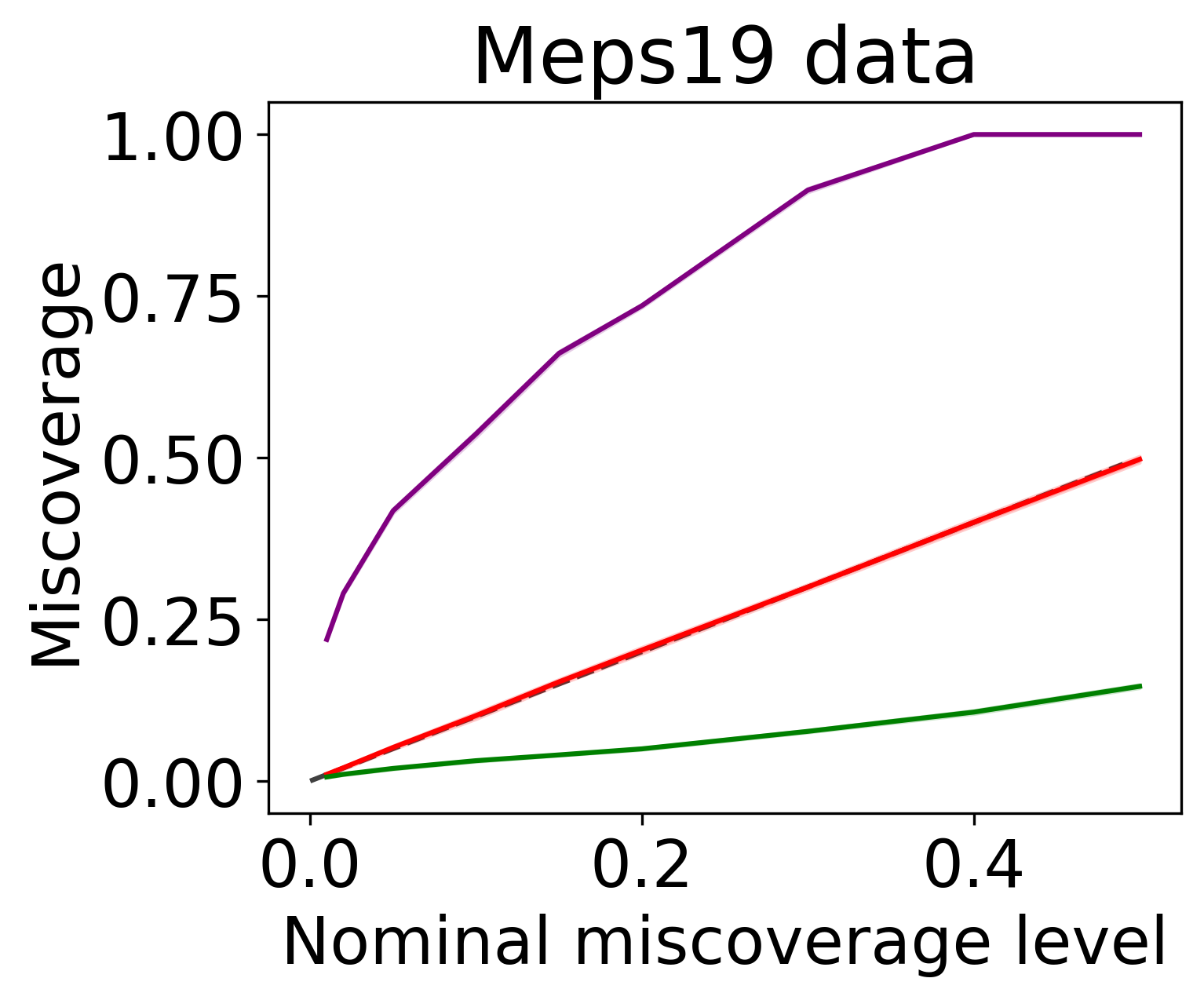}}
\caption{Miscoverage rate achieved over noisy (red) and clean (green) test sets. The calibration scheme is applied using noisy annotations to control the miscoverage level. Results are averaged over 10 random splits of the calibration and test sets.}
\label{fig:miscoverage_bounds}
\end{center}
\end{figure}

We repeat the experimental protocol detailed in Section~\ref{sec:exp_reg_bounds} with the same data sets and analyze the miscoverage bounds formulated in Section~\ref{sec:risk_regression}. Here, we apply conformalized quantile regression to control the miscoverage rate at different levels using a noisy calibration set. Furthermore, we choose the miscoverage bound hyperparameters, $c,d$,  from~\eqref{eq:coverage_taylor_bound} by a grid search over the calibration set with the objective of tightening the miscoverage bound. Figure~\ref{fig:miscoverage_bounds} displays the miscoverage rate achieved on the clean and noisy versions of the test set, along with the miscoverage bound. Importantly, in contrast to Theorem~\ref{thm:general_regression}, this bound requires no assumptions on the distribution of $Y\mid X=x$ or on the distribution of the noise $\varepsilon$. The only requirement is that the noise is independent of the response variable and has mean $0$. This great advantage covers up for the looseness of the bound.

\section{Related Algorithms}\label{sec:related_algorithms}

\subsection{Conformal Risk Control}\label{sec:risk_alg}
Here we provide a pseudo code of the conformal risk control algorithm, following~\citep{angelopoulos2022conformal}.

\begin{algorithm}
\caption{Conformal risk control}\label{alg:risk_algo}
\textbf{Input:} Exchangeable non-increasing losses, $L_i:\Lambda \rightarrow (-\infty,B], i=1,\dots,n$, where $L_i(\lambda)=L(Y_i, \Chat_{\lambda}(X_i))$, desired risk level $\alpha$.\\

\textbf{Process:}\\
     $\hat{R}_n(\lambda)=\frac{1}{n}\cdot(L_1(\lambda)+\dots+L_n(\lambda))$\\
     $\hat{\lambda}=\text{inf}\left\{\lambda:\frac{n}{n+1}\hat{R}_n(\lambda)+\frac{B}{n+1}\leq\alpha \right\}$\\
\textbf{Output:} Uncertainty set for a new test point, $\Chat_{\hat{\lambda}}(\Xtest)$, applied with $\hat{\lambda}$. 
\end{algorithm}

This prediction set $\Chat_{\hat{\lambda}}(\Xtest)$ produced by Algorithm~\ref{alg:risk_algo} satisfies $\mathbb{E}[L(\Ytest, \Chat_{\hat{\lambda}}(\Xtest))] \leq \alpha$, for proof see~\citep{angelopoulos2022conformal}.

\subsection{Adaptive Conformal Inference}\label{sec:aci_alg}
Below, we provide a pseudo-code of \texttt{ACI}, following~\cite[Algorithm~1]{aci}.
\begin{algorithm}
  \caption{Adaptive Conformal Inference}
  \label{alg:aci}
  
  	\textbf{Input:} Data $\{(X_t, Y_t)\}_{t=1}^T \subseteq \mathcal{X} \times \mathcal{Y}$, given as a stream, miscoverage level $\alpha\in(0,1)$, a score function $S$, a calibration set size $n_2$, a step size $\gamma >0$, and a learning model $\mathcal{M}$.
	
  	\textbf{Process:}
         Initialize $\alpha_0=\alpha$.\\
  	    \For{$t=n_2,...,T$}{
  	          Split the observed data $\{(X_i,Y_i) \}_{i=1}^{t-1}$ into a training set of size $t-1-n_2$, indexed by $\mathcal{I}_1$ and a calibration set of size $n_2$, indexed by $\mathcal{I}_2$.\\
  	         Obtain a predictive model $\mathcal{M}_{t}$ by fitting the initial one on the training set $\{(X_i,Y_i) \}_{i\in\mathcal{I}_1}$.\\
  	         Construct a prediction set for the new point $X_{t}$:
  	        \begin{equation}
  	            \Chat_t(X_t)= \{y\in\mathcal{Y} : S(\mathcal{M}_t(X_t),y) \leq Q_{1-\alpha_t}(\{S(\mathcal{M}_{t}(X_i),Y_i) \}_{i\in\mathcal{I}_2})\}
  	        \end{equation}\\
  	         Obtain $Y_t$.\\
  	         Compute $\text{err}_t=\mathbbm{1}{\{Y_t \notin \Chat_t(X_t)\}}$.\\
	         Update $\alpha_{t+1}= \alpha_t + \gamma(\alpha-\text{err}_t)$.\\
}  
  	\textbf{Output:}
		 Uncertainty sets $\Chat_t(X_t)$ for each time step $t=n_2,...,T$.
\end{algorithm}

\subsection{Rolling Risk Control}\label{sec:rrc_alg}

Here we provide a pseudo code of \texttt{Rolling RC}, first developed by~\citep{feldman2022achieving}.

\begin{algorithm}
\caption{Rolling Risk Control}\label{alg:rrc}
\textbf{Input:} Data $\{(X_t, Y_t)\}_{t=1}^T \subseteq \mathcal{X} \times \mathcal{Y}$, given as a stream, desired risk level $r\in\mathbb{R}$, a step size $\gamma >0$, a set constructing function $f:(\mathcal{X}, \mathbb{R}, \mathbb{M})\rightarrow 2^{\mathcal{Y}}$ and an online learning model $\mathcal{M}$.

\textbf{Process:}\\
          Initialize $\theta_0=0$.\\
          \For{$t = 1$ \KwTo $T$}{
                  Construct a prediction set for the new point $X_{t}$: $\Chat_t(X_t)=f(X_t,\theta_t,\mathcal{M}_t)$.\\
                Obtain $Y_t$.\\
                 Compute $l_{t} =  L(Y_t, \Chat_t(X_t))$.\\
                 Update $\theta_{t+1}= \theta_t + \gamma(l_t-r)$.\\
                 Fit the model $\mathcal{M}_t$ on $(X_t,Y_t)$ and obtain the updated model $\mathcal{M}_{t+1}$.
    }

\textbf{Output:} Uncertainty sets $\Chat_t(X_t)$ for each time step $t\in\{1,...T\}$. 
\end{algorithm}
\bibliography{bibliography}

\begin{thebibliography}{55}
\providecommand{\natexlab}[1]{#1}
\providecommand{\url}[1]{\texttt{#1}}
\expandafter\ifx\csname urlstyle\endcsname\relax
  \providecommand{\doi}[1]{doi: #1}\else
  \providecommand{\doi}{doi: \begingroup \urlstyle{rm}\Url}\fi

\bibitem[Abdalla and Fine(2023)]{abdalla2023hurdles}
Mohamed Abdalla and Benjamin Fine.
\newblock Hurdles to artificial intelligence deployment: Noise in schemas and
  “gold” labels.
\newblock \emph{Radiology: Artificial Intelligence}, 5\penalty0 (2):\penalty0
  e220056, 2023.

\bibitem[Algan and Ulusoy(2020)]{algan2020label}
G{\"o}rkem Algan and Ilkay Ulusoy.
\newblock Label noise types and their effects on deep learning.
\newblock \emph{arXiv preprint arXiv:2003.10471}, 2020.

\bibitem[Angelopoulos and Bates(2023)]{angelopoulos2021gentle}
Anastasios~N. Angelopoulos and Stephen Bates.
\newblock Conformal prediction: A gentle introduction.
\newblock \emph{Foundations and Trends® in Machine Learning}, 16\penalty0
  (4):\penalty0 494--591, 2023.
\newblock ISSN 1935-8237.

\bibitem[Angelopoulos et~al.(2021)Angelopoulos, Bates, Cand{\`e}s, Jordan, and
  Lei]{angelopoulos2021learn}
Anastasios~N. Angelopoulos, Stephen Bates, Emmanuel~J. Cand{\`e}s, Michael~I.
  Jordan, and Lihua Lei.
\newblock Learn then test: Calibrating predictive algorithms to achieve risk
  control.
\newblock \emph{arXiv preprint}, 2021.
\newblock arXiv:2110.01052.

\bibitem[Angelopoulos et~al.(2024)Angelopoulos, Bates, Fisch, Lei, and
  Schuster]{angelopoulos2022conformal}
Anastasios~Nikolas Angelopoulos, Stephen Bates, Adam Fisch, Lihua Lei, and Tal
  Schuster.
\newblock Conformal risk control.
\newblock In \emph{The Twelfth International Conference on Learning
  Representations}, 2024.

\bibitem[Angluin and Laird(1988)]{angluin1988learning}
Dana Angluin and Philip Laird.
\newblock Learning from noisy examples.
\newblock \emph{Machine Learning}, 2\penalty0 (4):\penalty0 343--370, 1988.

\bibitem[Aslam and Decatur(1996)]{aslam1996sample}
Javed~A Aslam and Scott~E Decatur.
\newblock On the sample complexity of noise-tolerant learning.
\newblock \emph{Information Processing Letters}, 57\penalty0 (4):\penalty0
  189--195, 1996.

\bibitem[Barber(2020)]{barber2020distributionfree}
Rina~Foygel Barber.
\newblock {Is distribution-free inference possible for binary regression?}
\newblock \emph{Electronic Journal of Statistics}, 14\penalty0 (2):\penalty0
  3487 -- 3524, 2020.

\bibitem[Barber et~al.(2021)Barber, Cand{\`e}s, Ramdas, and
  Tibshirani]{barber2019jackknife}
Rina~Foygel Barber, Emmanuel~J Cand{\`e}s, Aaditya Ramdas, and Ryan~J
  Tibshirani.
\newblock Predictive inference with the jackknife+.
\newblock \emph{The Annals of Statistics}, 49\penalty0 (1):\penalty0 486 --
  507, 2021.

\bibitem[Barber et~al.(2023)Barber, Cand{\`e}s, Ramdas, and
  Tibshirani]{barber2022conformal}
Rina~Foygel Barber, Emmanuel~J. Cand{\`e}s, Aaditya Ramdas, and Ryan~J.
  Tibshirani.
\newblock {Conformal prediction beyond exchangeability}.
\newblock \emph{The Annals of Statistics}, 51\penalty0 (2):\penalty0 816 --
  845, 2023.

\bibitem[Bates et~al.(2021)Bates, Angelopoulos, Lei, Malik, and
  Jordan]{bates2021distributionfree}
Stephen Bates, Anastasios Angelopoulos, Lihua Lei, Jitendra Malik, and
  Michael~I. Jordan.
\newblock Distribution-free, risk-controlling prediction sets.
\newblock \emph{Journal of the ACM}, 68\penalty0 (6), September 2021.
\newblock ISSN 0004-5411.

\bibitem[Battleday et~al.(2020)Battleday, Peterson, and
  Griffiths]{battleday2020capturing}
Ruairidh~M Battleday, Joshua~C Peterson, and Thomas~L Griffiths.
\newblock Capturing human categorization of natural images by combining deep
  networks and cognitive models.
\newblock \emph{Nature Communications}, 11\penalty0 (1):\penalty0 1--14, 2020.

\bibitem[bio()]{bio_data}
bio.
\newblock Physicochemical properties of protein tertiary structure data set.
\newblock
  \url{https://archive.ics.uci.edu/ml/datasets/Physicochemical+Properties+of+Protein+Tertiary+Structure}.
\newblock Accessed: January, 2019.

\bibitem[Cauchois et~al.(2022)Cauchois, Gupta, Ali, and
  Duchi]{cauchois2022weak}
Maxime Cauchois, Suyash Gupta, Alnur Ali, and John Duchi.
\newblock Predictive inference with weak supervision.
\newblock \emph{arXiv preprint arXiv:2201.08315}, 2022.

\bibitem[Cheng et~al.(2022)Cheng, Asi, and Duchi]{cheng2022many}
Chen Cheng, Hilal Asi, and John Duchi.
\newblock How many labelers do you have? a closer look at gold-standard labels.
\newblock \emph{arXiv preprint arXiv:2206.12041}, 2022.

\bibitem[Fan et~al.(2020)Fan, Ji, Zhou, Chen, Fu, Shen, and
  Shao]{fan2020pranet}
Deng-Ping Fan, Ge-Peng Ji, Tao Zhou, Geng Chen, Huazhu Fu, Jianbing Shen, and
  Ling Shao.
\newblock Pranet: Parallel reverse attention network for polyp segmentation.
\newblock In \emph{International conference on medical image computing and
  computer-assisted intervention}, pages 263--273. Springer, 2020.

\bibitem[Farinhas et~al.(2024)Farinhas, Zerva, Ulmer, and
  Martins]{farinhas2024nonexchangeable}
Ant{\'o}nio Farinhas, Chrysoula Zerva, Dennis~Thomas Ulmer, and Andre Martins.
\newblock Non-exchangeable conformal risk control.
\newblock In \emph{The Twelfth International Conference on Learning
  Representations}, 2024.
\newblock URL \url{https://openreview.net/forum?id=j511LaqEeP}.

\bibitem[Feldman et~al.(2023{\natexlab{a}})Feldman, Einbinder, Bates,
  Angelopoulos, Gendler, and Romano]{feldman2023conformal}
Shai Feldman, Bat-Sheva Einbinder, Stephen Bates, Anastasios~N Angelopoulos,
  Asaf Gendler, and Yaniv Romano.
\newblock Conformal prediction is robust to dispersive label noise.
\newblock In \emph{Conformal and Probabilistic Prediction with Applications},
  pages 624--626. PMLR, 2023{\natexlab{a}}.

\bibitem[Feldman et~al.(2023{\natexlab{b}})Feldman, Ringel, Bates, and
  Romano]{feldman2022achieving}
Shai Feldman, Liran Ringel, Stephen Bates, and Yaniv Romano.
\newblock Achieving risk control in online learning settings.
\newblock \emph{Transactions on Machine Learning Research}, 2023{\natexlab{b}}.
\newblock ISSN 2835-8856.

\bibitem[Fr{\'e}nay and Verleysen(2013)]{frenay2013classification}
Beno{\^\i}t Fr{\'e}nay and Michel Verleysen.
\newblock Classification in the presence of label noise: a survey.
\newblock \emph{IEEE Transactions on Neural Networks and Learning Systems},
  25\penalty0 (5):\penalty0 845--869, 2013.

\bibitem[Geiger et~al.(2013)Geiger, Lenz, Stiller, and Urtasun]{kitti}
Andreas Geiger, Philip Lenz, Christoph Stiller, and Raquel Urtasun.
\newblock Vision meets robotics: The kitti dataset.
\newblock \emph{International Journal of Robotics Research}, 2013.

\bibitem[Gendler et~al.(2021)Gendler, Weng, Daniel, and
  Romano]{gendler2021adversarially}
Asaf Gendler, Tsui-Wei Weng, Luca Daniel, and Yaniv Romano.
\newblock Adversarially robust conformal prediction.
\newblock In \emph{International Conference on Learning Representations}, 2021.

\bibitem[Gibbs and Candes(2021)]{aci}
Isaac Gibbs and Emmanuel Candes.
\newblock Adaptive conformal inference under distribution shift.
\newblock In A.~Beygelzimer, Y.~Dauphin, P.~Liang, and J.~Wortman Vaughan,
  editors, \emph{Advances in Neural Information Processing Systems}, 2021.

\bibitem[Jenni and Favaro(2018)]{jenni2018deep}
Simon Jenni and Paolo Favaro.
\newblock Deep bilevel learning.
\newblock In \emph{Proceedings of the European Conference on Computer Vision},
  pages 618--633, 2018.

\bibitem[Jindal et~al.(2016)Jindal, Nokleby, and Chen]{jindal2016learning}
Ishan Jindal, Matthew Nokleby, and Xuewen Chen.
\newblock Learning deep networks from noisy labels with dropout regularization.
\newblock In \emph{2016 IEEE 16th International Conference on Data Mining},
  pages 967--972. IEEE, 2016.

\bibitem[Kao et~al.(2015)Kao, Wang, and Huang]{kao2015visual}
Yueying Kao, Chong Wang, and Kaiqi Huang.
\newblock Visual aesthetic quality assessment with a regression model.
\newblock In \emph{2015 IEEE International Conference on Image Processing},
  pages 1583--1587. IEEE, 2015.

\bibitem[Koenker and Bassett(1978)]{koenker1978regression}
Roger Koenker and Gilbert Bassett.
\newblock Regression quantiles.
\newblock \emph{Econometrica: Journal of the Econometric Society}, pages
  33--50, 1978.

\bibitem[Kumar et~al.(2020)Kumar, Manwani, and Sastry]{kumar2020robust}
Himanshu Kumar, Naresh Manwani, and PS~Sastry.
\newblock Robust learning of multi-label classifiers under label noise.
\newblock In \emph{Proceedings of the ACM IKDD CoDS and COMAD}, pages 90--97.
  2020.

\bibitem[Lee and Barber(2022)]{lee2022binary}
Yonghoon Lee and Rina~Foygel Barber.
\newblock {Binary classification with corrupted labels}.
\newblock \emph{Electronic Journal of Statistics}, 16\penalty0 (1):\penalty0
  1367 -- 1392, 2022.

\bibitem[Lei et~al.(2013)Lei, Robins, and Wasserman]{lei2013distribution}
Jing Lei, James Robins, and Larry Wasserman.
\newblock Distribution-free prediction sets.
\newblock \emph{{Journal of the American Statistical Association}},
  108\penalty0 (501):\penalty0 278--287, 2013.

\bibitem[Lei et~al.(2018)Lei, G’Sell, Rinaldo, Tibshirani, and
  Wasserman]{lei2018distribution}
Jing Lei, Max G’Sell, Alessandro Rinaldo, Ryan~J. Tibshirani, and Larry
  Wasserman.
\newblock Distribution-free predictive inference for regression.
\newblock \emph{{Journal of the American Statistical Association}},
  113\penalty0 (523):\penalty0 1094--1111, 2018.

\bibitem[Lin et~al.(2014)Lin, Maire, Belongie, Hays, Perona, Ramanan,
  Doll{\'a}r, and Zitnick]{lin2014microsoft}
Tsung-Yi Lin, Michael Maire, Serge Belongie, James Hays, Pietro Perona, Deva
  Ramanan, Piotr Doll{\'a}r, and C~Lawrence Zitnick.
\newblock Microsoft {COCO}: Common objects in context.
\newblock In \emph{European Conference on Computer Vision}, pages 740--755.
  Springer, 2014.

\bibitem[Ma et~al.(2018)Ma, Wang, Houle, Zhou, Erfani, Xia, Wijewickrema, and
  Bailey]{ma2018dimensionality}
Xingjun Ma, Yisen Wang, Michael~E Houle, Shuo Zhou, Sarah Erfani, Shutao Xia,
  Sudanthi Wijewickrema, and James Bailey.
\newblock Dimensionality-driven learning with noisy labels.
\newblock In \emph{International Conference on Machine Learning}, pages
  3355--3364. PMLR, 2018.

\bibitem[meps\_19()]{meps19_data}
meps\_19.
\newblock Medical expenditure panel survey, panel 19.
\newblock
  \url{https://meps.ahrq.gov/mepsweb/data_stats/download_data_files_detail.jsp?cboPufNumber=HC-181}.
\newblock Accessed: January, 2019.

\bibitem[Murray et~al.(2012)Murray, Marchesotti, and Perronnin]{murray2012ava}
Naila Murray, Luca Marchesotti, and Florent Perronnin.
\newblock Ava: A large-scale database for aesthetic visual analysis.
\newblock In \emph{2012 IEEE Conference on Computer Vision and Pattern
  Recognition}, pages 2408--2415. IEEE, 2012.

\bibitem[Papadopoulos et~al.(2002)Papadopoulos, Proedrou, Vovk, and
  Gammerman]{papadopoulos2002inductive}
Harris Papadopoulos, Kostas Proedrou, Vladimir Vovk, and Alex Gammerman.
\newblock Inductive confidence machines for regression.
\newblock In \emph{{Machine Learning: European Conference on Machine
  Learning}}, pages 345--356, 2002.

\bibitem[Peterson et~al.(2019)Peterson, Battleday, Griffiths, and
  Russakovsky]{peterson2019human}
Joshua~C Peterson, Ruairidh~M Battleday, Thomas~L Griffiths, and Olga
  Russakovsky.
\newblock Human uncertainty makes classification more robust.
\newblock In \emph{Proceedings of the IEEE/CVF International Conference on
  Computer Vision}, pages 9617--9626, 2019.

\bibitem[Podkopaev and Ramdas(2021)]{podkopaev2021distribution}
Aleksandr Podkopaev and Aaditya Ramdas.
\newblock Distribution-free uncertainty quantification for classification under
  label shift.
\newblock In \emph{Uncertainty in Artificial Intelligence}, pages 844--853.
  PMLR, 2021.

\bibitem[Ridnik et~al.(2021)Ridnik, Lawen, Noy, Ben~Baruch, Sharir, and
  Friedman]{ridnik2021tresnet}
Tal Ridnik, Hussam Lawen, Asaf Noy, Emanuel Ben~Baruch, Gilad Sharir, and
  Itamar Friedman.
\newblock Tresnet: High performance gpu-dedicated architecture.
\newblock In \emph{proceedings of the IEEE/CVF Winter Conference on
  Applications of Computer Vision}, pages 1400--1409, 2021.

\bibitem[Romano et~al.(2019)Romano, Patterson, and
  Cand{\`e}s]{romano2019conformalized}
Yaniv Romano, Evan Patterson, and Emmanuel Cand{\`e}s.
\newblock Conformalized quantile regression.
\newblock In \emph{Advances in Neural Information Processing Systems},
  volume~32, pages 3543--3553. 2019.

\bibitem[Romano et~al.(2020)Romano, Sesia, and
  Cand{\`e}s]{romano2020classification}
Yaniv Romano, Matteo Sesia, and Emmanuel Cand{\`e}s.
\newblock Classification with valid and adaptive coverage.
\newblock In \emph{Advances in Neural Information Processing Systems},
  volume~33, pages 3581--3591, 2020.

\bibitem[Sesia et~al.(2023)Sesia, Wang, and Tong]{sesia2023adaptive}
Matteo Sesia, YX~Wang, and Xin Tong.
\newblock Adaptive conformal classification with noisy labels.
\newblock \emph{arXiv preprint arXiv:2309.05092}, 2023.

\bibitem[Singh et~al.(2020)Singh, Peterson, Battleday, and
  Griffiths]{singh2020end}
Pulkit Singh, Joshua~C Peterson, Ruairidh~M Battleday, and Thomas~L Griffiths.
\newblock End-to-end deep prototype and exemplar models for predicting human
  behavior.
\newblock \emph{arXiv preprint arXiv:2007.08723}, 2020.

\bibitem[Stutz et~al.(2023)Stutz, Roy, Matejovicova, Strachan, Cemgil, and
  Doucet]{stutz2023conformal}
David Stutz, Abhijit~Guha Roy, Tatiana Matejovicova, Patricia Strachan,
  Ali~Taylan Cemgil, and Arnaud Doucet.
\newblock Conformal prediction under ambiguous ground truth.
\newblock \emph{arXiv preprint arXiv:2307.09302}, 2023.

\bibitem[Talebi and Milanfar(2018)]{talebi2018nima}
Hossein Talebi and Peyman Milanfar.
\newblock {NIMA}: Neural image assessment.
\newblock \emph{IEEE Transactions on Image Processing}, 27\penalty0
  (8):\penalty0 3998--4011, 2018.

\bibitem[Tanno et~al.(2019)Tanno, Saeedi, Sankaranarayanan, Alexander, and
  Silberman]{tanno2019learning}
Ryutaro Tanno, Ardavan Saeedi, Swami Sankaranarayanan, Daniel~C Alexander, and
  Nathan Silberman.
\newblock Learning from noisy labels by regularized estimation of annotator
  confusion.
\newblock In \emph{Proceedings of the IEEE/CVF Conference on Computer Vision
  and Pattern Recognition}, pages 11244--11253, 2019.

\bibitem[Tibshirani et~al.(2019)Tibshirani, Foygel~Barber, Candes, and
  Ramdas]{tibshirani2019conformal}
Ryan~J Tibshirani, Rina Foygel~Barber, Emmanuel Candes, and Aaditya Ramdas.
\newblock Conformal prediction under covariate shift.
\newblock In \emph{{Advances in Neural Information Processing Systems}},
  volume~32, pages 2530--2540. 2019.

\bibitem[Vovk(2015)]{vovk2015cross}
Vladimir Vovk.
\newblock Cross-conformal predictors.
\newblock \emph{{Annals of Mathematics and Artificial Intelligence}},
  74\penalty0 (1-2):\penalty0 9--28, 2015.

\bibitem[Vovk et~al.(1999)Vovk, Gammerman, and Saunders]{vovk1999machine}
Vladimir Vovk, Alexander Gammerman, and Craig Saunders.
\newblock Machine-learning applications of algorithmic randomness.
\newblock In \emph{{International Conference on Machine Learning}}, pages
  444--453, 1999.

\bibitem[Vovk et~al.(2005)Vovk, Gammerman, and Shafer]{vovk2005algorithmic}
Vladimir Vovk, Alex Gammerman, and Glenn Shafer.
\newblock \emph{{Algorithmic Learning in a Random World}}.
\newblock Springer, New York, NY, USA, 2005.

\bibitem[Wei et~al.(2022)Wei, Zhu, Cheng, Liu, Niu, and Liu]{wei2022learning}
Jiaheng Wei, Zhaowei Zhu, Hao Cheng, Tongliang Liu, Gang Niu, and Yang Liu.
\newblock Learning with noisy labels revisited: A study using real-world human
  annotations.
\newblock In \emph{International Conference on Learning Representations}, 2022.

\bibitem[Xu et~al.(2019)Xu, Cao, Kong, and Wang]{xu2019l_dmi}
Yilun Xu, Peng Cao, Yuqing Kong, and Yizhou Wang.
\newblock L\_dmi: A novel information-theoretic loss function for training deep
  nets robust to label noise.
\newblock \emph{Advances in Neural Information Processing Systems}, 32, 2019.

\bibitem[Yin et~al.(2021)Yin, Zhang, Wang, Niklaus, Mai, Chen, and Shen]{leres}
Wei Yin, Jianming Zhang, Oliver Wang, Simon Niklaus, Long Mai, Simon Chen, and
  Chunhua Shen.
\newblock Learning to recover 3d scene shape from a single image.
\newblock In \emph{Proceedings of the IEEE/CVF Conference on Computer Vision
  and Pattern Recognition}, pages 204--213, 2021.

\bibitem[Yuan et~al.(2018)Yuan, Chen, Zhang, Tai, and
  McMains]{yuan2018iterative}
Bodi Yuan, Jianyu Chen, Weidong Zhang, Hung-Shuo Tai, and Sara McMains.
\newblock Iterative cross learning on noisy labels.
\newblock In \emph{IEEE Winter Conference on Applications of Computer Vision},
  pages 757--765. IEEE, 2018.

\bibitem[Zhao and Gomes(2021)]{zhao2021evaluating}
Wenting Zhao and Carla Gomes.
\newblock Evaluating multi-label classifiers with noisy labels.
\newblock \emph{arXiv preprint arXiv:2102.08427}, 2021.

\end{thebibliography}

\end{document}